%% file: main.tex
\documentclass{article}

\usepackage[final]{neurips_2019}

\usepackage[utf8]{inputenc} 
\usepackage[T1]{fontenc}    
\usepackage{hyperref}       
\usepackage{url}            
\usepackage{booktabs}       
\usepackage{amsfonts}       
\usepackage{nicefrac}       
\usepackage{microtype}      
\usepackage{mathrsfs}
\usepackage{qiangstyle}

\usepackage{wrapfig,lipsum,booktabs}

\usepackage{tabularx}

\title{Stein Variational Gradient Descent with Matrix-Valued Kernels}

\author{%
{\rm Dilin Wang\textsuperscript{*} ~~~Ziyang Tang\thanks{Equal contribution} ~~~Chandrajit Bajaj ~~~Qiang Liu}\\
{\rm Department of Computer Science, UT Austin}\\
{\rm \texttt{\{dilin, ztang, bajaj, lqiang\}@cs.utexas.edu}} 
}

\begin{document}

\maketitle

\begin{abstract}
    Stein variational gradient descent (SVGD)
    is a particle-based inference algorithm that  leverages gradient information for efficient approximate inference.
    In this work, we enhance SVGD by leveraging preconditioning matrices,  
    such as the Hessian and Fisher information matrix, to incorporate geometric information into SVGD updates. 
    We achieve this by presenting a generalization of SVGD that 
    replaces the \emph{scalar-valued} kernels in vanilla SVGD with more general 
     \emph{matrix-valued} kernels. 
    This yields a significant extension of  SVGD, and more importantly, 
     allows us to flexibly incorporate various preconditioning matrices
    to accelerate the 
    exploration in the probability landscape. 
    Empirical results show that our method outperforms vanilla SVGD and a variety of baseline approaches over a range of real-world Bayesian inference tasks.  
\end{abstract}

 \input{tex/introduction.tex}

\input{tex/background.tex}
\input{tex/method_new.tex}
\input{tex/experiments.tex}
\section{Conclusion}
We present a generalization of SVGD by leveraging general matrix-valued positive definite kernels,
which allows us to flexibly incorporate various preconditioning matrices, including Hessian and Fisher information matrices, to improve exploration in the probability landscape. 
We test our practical algorithms on various practical tasks and demonstrate its efficiency compared to various existing methods. 

\section*{Acknowledgement}
This work is supported in part by NSF CRII 1830161 and NSF CAREER
1846421. We would like to acknowledge Google Cloud and Amazon Web Services (AWS) for their support.

 \bibliographystyle{icml2019}
\bibliography{ref}

\newpage\clearpage
 \input{tex/proof.tex}

\clearpage
 \input{tex/appendix_2d_figures.tex}

\end{document}

%% file: tex/introduction.tex
\section{Introduction}

Approximate inference  of intractable distributions is a central  task in probabilistic  learning and statistics.  
An efficient approximation
inference algorithm 
must perform both 
efficient \emph{optimization} to explore the high probability regions of the distributions of interest, 
and reliable \emph{uncertainty quantification} for evaluating the variation of the given  distributions.    
%
Stein variational gradient descent (SVGD) \citep{liu2016stein} 
is a deterministic sampling algorithm that 
achieves both desiderata 
by optimizing the samples using a procedure similar to gradient-based optimization,  
while achieving reliable uncertainty estimation using an interacting repulsive mechanism.  
SVGD has been shown to provide a fast and flexible alternative to traditional methods such  as  Markov chain Monte Carlo (MCMC) \citep[e.g.,][]{neal2011mcmc, hoffman2014no} and  parametric variational  inference (VI) \citep[e.g.,][]{wainwright2008graphical, blei2017variational} in various challenging applications \citep[e.g.,][]{pu2017vae, wang2016learning, kim2018bayesian, haarnoja2017reinforcement}.

On the other hand, standard SVGD only uses the first order  gradient information, and can not leverage the advantage of the second order methods, such as Newton's method and natural gradient, to achieve better performance on 
challenging problems with complex loss landscapes or domains. 
Unfortunately,  
due to the special form of SVGD, 
it is not straightforward to derive second order extensions of SVGD by simply extending similar ideas from optimization. 
While this problem has been recently considered   \citep[e.g.,][]{detommaso2018stein, liu2018riemannian, chen2019stein}, the presented solutions either require heuristic approximations \citep{detommaso2018stein}, 
or lead to complex algorithmic procedures that are difficult to implement in practice \citep{liu2018riemannian}. 

Our solution to this problem is through a key generalization of SVGD that
replaces the original scalar-valued positive definite kernels in SVGD 
with a class of more general \emph{matrix-valued} positive definite kernels. 
Our generalization includes all previous variants of SVGD  \citep[e.g.,][]{wang2018stein, han2018stein} as special cases. More significantly, it   
allows us to easily incorporate various structured preconditioning matrices into SVGD updates,   
including both Hessian and Fisher information matrices, as part of the generalized \emph{matrix-valued} positive definite kernels. 
We develop theoretical results that shed insight on optimal design of the matrix-valued kernels,  
and also propose simple and fast practical  procedures.  
We empirically evaluate both Newton and Fisher based extensions of SVGD  on various practical  benchmarks, including Bayesian neural regression  and sentence classification,   
on which our methods show significant improvement over vanilla SVGD and other baseline approaches.


\paragraph{Notation and Preliminary}
For notation, we use 
bold lower-case letters (e.g., $\vx$) for vectors in $\RR^d$, 
and bold upper-case letters (e.g., $\vv Q$) for matrices.  
A symmetric function $k\colon \X\times \X \to \R$ is called a 
positive definite kernel if $\sum_{ij} c_i k(\vx_i, \vx_j) c_j \geq 0$ for any $\{c_i\}\subset \R$ and $\{\vx_i\}\subset \X$.  
Every positive definite kernel $k(\vx,\vx')$ is associated with a \emph{reproducing kernel Hilbert space} (RKHS)  $\H_{k}$, which consists of the closure of functions of form
\begin{align} \label{equ:fck} 
f(\vx) = \sum_i c_i k(\vx, \vx_i), ~~~~
\forall \{c_i\}\subset \R, ~~\{\vx_i\}\subset \X, 
\end{align}
for which the inner product and norm are defined by   
$
\langle f, ~g \rangle_{\H_k} 
\!=\!\! \sum_{ij} c_i s_j k(\vx_i,\vx_j)$, 
$\norm{f}^2_{\H_k}\! = \!\!\sum_{ij} c_i c_j k(\vx_i,\vx_j),$
where we assume $g(\vx) = \sum_i s_i k(\vx,\vx_i)$. 
Denote by $\H_k^d: = \H_k \times \ldots \times \H_k$ the vector-valued RKHS consisting of $\RR^d$ vector-valued functions $\ff=[\phi^1, \ldots, \phi^d]^\top$ with each $\phi^\ell \in \H_k$. 
See e.g., \citet[][]{berlinet2011reproducing} for more rigorous treatment. 
For notation convenience, 
we do not distinguish distributions on $\RR^d$ and and their density functions. 

%% file: tex/background.tex

\section{Stein Variational Gradient Descent (SVGD)}
We introduce the basic derivation of Stein variational gradient descent (SVGD), 
which provides a foundation for our new generalization. See \citet{liu2016stein,liu2018stein,  liu2017stein} for more details. 

Let $p(\vx)$ be  a positive and continuously differentiable probability density function on $\RR^d$. 
Our goal is to find a set of points (a.k.a. particles)  $\{\vx_i\}_{i=1}^n \subset \RR^d$ to approximate $p$, such that
the empirical distribution $q(\vx) = \sum_i \delta(\vx- \vx_i)/n$ 
of the particles weakly converges to $p$ when $n$ is large. Here $\delta(\cdot)$ denotes the Dirac delta function. 

SVGD achieves this by 
starting from a set of initial  particles, 
and iteratively updating them 
with a deterministic transformation  of form  
\begin{align}
\vx_i \leftarrow \vx_i + \epsilon \ff_k^*(\vx_i), ~~~\forall i = 1, \cdots, n, &&& 
	\ff_k^* = 
	\argmax_{\ff \in \mathcal B_k} \bigg\{  \!\!-\frac{\dno}{\dno \epsilon} \KL(q_{[\epsilon \ff]} \parallel p) \bigg |_{\epsilon=0} 
	\bigg \},
	\label{equ:max-kl}
\end{align}
where $\epsilon$ is a small step size, $\ff_k^* \colon \X \to \R^d$ 
is an 
optimal transform function 
chosen to maximize the decreasing rate of the KL divergence between the distribution of particles and the target $p$,
and 
$q_{[\epsilon\ff]}$ denotes the distribution of the updated particles 
$\vx' = \vx + \epsilon \ff(\vx)$ as $\vx\sim q$,
and $\mathcal B_k$ is the unit ball of RKHS $\H_k^d: = \H_k \times \ldots \times \H_k$ associated with a positive definite kernel $k(\vx,\vx')$, that is, 
\begin{align} \label{equ:fset} 
\mathcal B_k = \{\ff\in \H_k^d \colon ~~ \norm{\ff}_{\H_k^d} \leq 1\}. 
\end{align}
%
\citet{liu2016stein} showed that 
the objective in \eqref{equ:max-kl} can be expressed as a linear functional of $\ff$,
\begin{align}
-
\frac{\dno}{\dno \epsilon}   \KL(q_{[\epsilon\ff]}~||~p)  \bigg |_{\epsilon =0} = \E_{\vx\sim q}[\steinp^\top \ff(\vx)], 
&& 
 \steinp^\top \ff(\vx) =  \nabla_\vx \log p(\vx)^\top \ff(\vx) + \nabla_\vx^\top \ff(\vx),  
\label{def:stein_obj}
\end{align}
where $\steinp$ is a differential operator  called \emph{Stein operator}; here we formally view $\steinp$ and the derivative operator $\nabla_\vx$ as $\R^{d}$ column vectors, hence $\steinp^\top \ff$ and $\nabla_\vx^\top \ff$ are  viewed as inner products, e.g., $\nabla_{\vx}^\top \ff = \sum_{\ell=1}^d \nabla_{x^\ell} \phi^\ell$, with $x^\ell$ and $\phi^\ell$ being the $\ell$-th coordinate of vector $\vx$ and $\ff$, respectively. 

With \eqref{def:stein_obj}, 
it is shown in \citet{liu2016stein} that
the solution of \eqref{equ:max-kl} is 
\begin{align}
\label{equ:phistar}
\begin{split} 
\ff_k^*(\cdot) & \propto \E_{\vv x\sim q} [\steinp k(\vx,\cdot)]  
               = \E_{\vx\sim q}[ \nabla_\vx \log p(\vx) k(\vx,\cdot) + \nabla_\vx k(\vx,\cdot)].
\end{split}               
\end{align}
Such $\ff_k^*$ provides the best update direction for the particles within RKHS $\H_k^d$.    
By taking $q$ to be the empirical measure of the particles, i.e., $q(\vx)= \sum_{i=1}^n \delta(\vx-\vx_i)/n$ 
and repeatedly applying this update on the particles, 
we obtain the SVGD algorithm using equations \eqref{equ:max-kl} and \eqref{equ:phistar}. 

\section{SVGD with Matrix-valued Kernels} 


Our goal is to extend SVGD to allow efficient incorporation of precondition information for better optimization. We achieve this by providing a generalization of SVGD that leverages more general \emph{matrix-valued} kernels, to flexibly incorporate preconditioning information.  

The key idea is to observe that the standard SVGD searches for the optimal $\ff$ in RKHS $\H_k^d = \H_k\times\cdots\times \H_k$, 
a product of $d$ copies of RKHS of scalar-valued functions, which does not allow us to encode potential correlations between different coordinates of $\ff$. 
This limitation can be addressed by replacing $\H_k^d$ with a more general RKHS of vector-valued functions (called \emph{vector-valued RKHS}),  
which uses more flexible \emph{matrix-valued}  positive definite kernels to  
specify rich correlation structures between  different coordinates. 
In this section,  
we first introduce the background of 
vector-valued RKHS with matrix-valued kernels in Section~\ref{sec:vrkhs},  
and then propose and discuss our generalization of SVGD using matrix-valued kernels  in  Section~\ref{sec:main}-\ref{sec:discuss}.  

\subsection{Vector-Valued RKHS with Matrix-Valued Kernels} \label{sec:vrkhs}
We now introduce the background of matrix-valued positive definite kernels, which provides a most general framework for specifying vector-valued RKHS. 
We focus on the intuition and key ideas in our introduction, and refer the readers to  \citet{alvarez2012kernels, carmeli2006vector}
for mathematical treatment.  

Recall that a standard real-valued RKHS $\H_k$ consists of the closure of the linear span of its kernel function $k(\cdot,\vx)$, as shown in \eqref{equ:fck}.  
Vector-valued RKHS can be defined in a similar way, 
but consist of the linear span of a \emph{matrix-valued} kernel function: 
\begin{align}\label{equ:fvxkmat}  
\vv f(\vx) = \sum_i \vv K(\vx, \vx_i)\vv c_i,
\end{align} 
for any  
 $\{\vv c_i\} \subset \X$ and 
  $\{\vx_i \} \subset \X$, 
where  $\vv K\colon \X\times \X \to \RR^{d\times d}$ is now a  matrix-valued kernel function, and $\vv c_i$ are vector-valued weights. 
Similar to the scalar case, we can define an inner product structure $\langle \vv f, \vv g \rangle_{\H_{\vv K}} =\sum_{ij} \vv c_i^\top\vv K(\vx_i, \vx_j) \vv s_j$, where we assume $\vv g = \sum_{i} \vv K(\vx, \vx_i) \vv s_i$,  and hence a norm $\norm{\vv f}_{\H_k}^2= \sum_{ij} \vv c_i^\top \vv K(\vx_i, \vx_j) \vv c_j$.
In order to make the inner product and norm well defined, 
the matrix-value kernel $\vv K$ is  required to be 
symmetric in that $\vv K(\vx, \vx') = \vv K(\vx', \vx)^\top$, and positive definite in that  
$
\sum_{ij} \vv c_i^\top \vv K(\vx_i, \vx_j) \vv c_j \geq 0,
$ 
for any 
$\{\vv x_i \} \subset \RR^d$ and $\{\vv c_i
\}\subset \RR^d$.  

Mathematically, one can show that the closure of the set of functions in \eqref{equ:fvxkmat},  equipped with the inner product defined above, defines a RKHS that we denote by  $\H_{\kk}$. It is ``{reproducing}'' because it has the following reproducing property that generalizes the version for scalar-valued RKHS: 
 for any $\vv f \in \H_{\vv K}$ and any $\vv c \in \RR^d$, we have 
\begin{equation}
    \label{equ:matrxi_reproduce}
    \vv f(\vx)^\top \vv c  =  \langle \vv f(\cdot ), ~ \vv K(\cdot, ~ \vx) \vv c\rangle_{\H_{\vv K}},
\end{equation}
where it is necessary to introduce 
$\vv c$ 
because the result of the inner product of two functions must be a scalar.  
A simple example of matrix kernel is $\vv K(\vx, \vx') = k(\vx, \vx') \vv I$, where $\vv I$ is the $d\times d$ identity matrix. It is related RKHS is $\H_{\vv K} = \H_k\times \cdots \times \H_k = \H_k^d$, as used in the original SVGD. 

%% file: tex/method_new.tex
\subsection{SVGD with Matrix-Valued Kernels}
\label{sec:main}  

It is now natural to leverage matrix-valued kernels to obtain a generalization of SVGD (see Algorithm~\ref{alg:main}).  
The idea is simple: we now optimize $\ff$ in the unit ball of a general vector-valued RKHS $\H_{\vv K}$ with a matrix valued kernel $\vv K(\vx, \vx')$:
\begin{align}\label{equ:matrixphi}
\ff^*_{\kk}=\argmax_{\ff\in \H_{\vv K}}  
\left \{ \E_{\vv x\sim q}\left  [ \steinp^\top \ff (\vv x) \right ], ~~ s.t. ~~ 
\norm{\ff}_{\H_{\vv K}} \leq 1
\right \}.
\end{align} 

This yields a simple closed form solution
similar to \eqref{equ:phistar}. 
\begin{thm}
Let $\kk(\vx,\vx')$ be a 
matrix-valued positive definite kernel that is continuously differentiable on $\vv x$ and $\vv x'$, the  
optimal $\ff^*$ in \eqref{equ:matrixphi} is 
\begin{align}
\begin{split} 
    \ff_{\kk}^*(\cdot) 
     \propto \E_{\vv x\sim q} \left [ \vv K(\cdot, \vx)\steinp \right ] 
    = \E_{\vv x\sim q}
    \left [\kk(\cdot, \vv x)\nabla_{\vv x} \log p(\vv x) + 
    \kk(\cdot, \vv x) \nabla_{\vv x} \right ], 
\end{split}
\label{equ:matrix_svgd}
\end{align}
where the Stein operator $\steinp$ and derivative operator $\nabla_{\vx}$ are again formally viewed as $\R^d$-valued column vectors, and $\vv K(\cdot, \vx)\steinp$ and $\vv K(\cdot, \vx)\nabla_{\vx}$ are interpreted by the matrix multiplication rule.  
Therefore, 
$\kk(\cdot, \vv x)\steinp$ 
is a $\RR^d$-valued column vector, 
whose $\ell$-th element is defined by 
\begin{align} \label{equ:kdd}
 (\kk(\cdot, \vv x) \steinp )_{\ell} 
& = 
\sum_{m=1}^d \left (K_{\ell,m}(\cdot, \vv x) \nabla_{x^m} \log p(\vv x) +  \nabla_{x^{m}} K_{\ell, m}(\cdot, \vv x)  \right),
\end{align}
where $K_{\ell, m}(\vx,\vx')$ denotes the $(\ell,m)$- element of matrix $\kk(\vx,\vx')$ 
and $x^m$ the $m$-th element of $\vx$. 
\label{thm:matrix_svgd}
\end{thm}

\begin{algorithm*}[t] 
\caption{Stein Variational Gradient Descent with Matrix-valued Kernels (Matrix SVGD)}
\begin{algorithmic} 
    \STATE \textbf{Input}: A (possibly unnormalized) differentiable density function $p(\vx)$ in $\RR^d$.  A matrix-valued positive definite kernel $\kk(\vx,\vx')$. Step size $\epsilon$. 
    \STATE \textbf{Goal}: Find a set of particles $\{\vv x_i\}_{i=1}^n$ to represent the distribution $p$. 
    \STATE \textbf{Initialize} a set of particles $\{\vx_{i}\}_{i=1}^n$, e.g., by drawing from some simple distribution. 
    \REPEAT 
    \STATE \vspace{-2\baselineskip}
    \begin{align*}
    \hspace{-2em}
\begin{split} 
\vx_i \gets  \vx_i +
    \frac{\epsilon}{n}
    \sum_{j=1}^n\left [\kk(\vx_i, \vv x_j)\nabla_{\vv x_j} \log p(\vv x_j) + 
    \kk(\vx_i, \vv x_j) \nabla_{\vv x_j} \right ], 
\end{split}
    \end{align*} 
where $\kk(\cdot, \vv x) \nabla_{\vv x}$ 
is formally defined as the product of matrix $\kk(\cdot, \vv x)$ and vector $\nabla_{\vv x}$. 
The 
 $\ell$-th element of  $\kk(\cdot, \vv x) \nabla_{\vv x}$ is 
$(\kk(\cdot, \vv x) \nabla_{\vv x} )_{\ell} 
= 
\sum_{m=1}^d \nabla_{x^{m}} K_{\ell, m}(\cdot, \vv x)$; see also 
\eqref{equ:kdd}.  
    \UNTIL{Convergence}
\end{algorithmic}
\label{alg:main}  
\end{algorithm*}

Similar to the case of standard SVGD, 
recursively applying the optimal transform $\ff_{\kk}^*$ on the particles yields a general SVGD algorithm shown in 
Algorithm~\ref{alg:main}, 
which we call \emph{matrix SVGD}.

Parallel to vanilla SVGD,    
the gradient of matrix SVGD in \eqref{equ:matrix_svgd}
consists of two parts that
account for optimization and diversity, respectively: 
the first part is a weighted average of gradient  $\nabla_{\vv x} \log p(\vx)$ multiplied by a matrix-value kernel $\kk(\cdot, \vv x)$;  
the other part consists of the gradient of the matrix-valued kernel $\kk$, which, like standard SVGD, serves as a repulsive force to keep the particles away from each other to reflect the uncertainty captured in  distribution $p$. 

Matrix SVGD includes various previous variants of SVGD as special cases.  
The {vanilla SVGD} corresponds to the case when $\kk(\vx,\vx') = k(\vx,\vx')\vv{I}$, with $\vv{I}$ as the $d\times d$ identity matrix; 
the {gradient-free SVGD} of \citet{han2018stein} can be treated as the case when  $\kk(\vx,\vx') = k(\vx,\vx') w(\vx) w(\vx') \vv{I}$,
where $w(\vx)$ is an importance weight function; 
the {graphical SVGD} of \citet{wang2018stein, zhuo2018message}   
corresponds to a diagonal matrix-valued kernel: 
$
\kk(\vx,\vx') =\mathrm{diag}[ \{k_\ell(\vx,\vx')\}_{\ell=1}^d ],
$
where each $k_\ell(\vx,\vx')$ is a ``local'' scalar-valued kernel function related to the $\ell$-th coordinate $x^\ell$ of vector $\vx$. 


\subsection{Matrix-Valued Kernels and  
Change of Variables}
\label{sec:discuss}
It is well known that 
preconditioned gradient descent
can be interpreted as applying standard gradient descent on 
a reparameterization of the variables.  
For example, let $\vv y = \vv Q^{1/2}\vx$, 
where $\vv Q$ is a positive definite matrix, then $\log p(\vx) = \log p(\vv Q^{-1/2}\vv y )$. 
Applying gradient descent on $\vv y$ and transform it back to the updates on $\vx$  yields
a preconditioned gradient update
$
\vv x  \gets \vx + \epsilon 
\vv Q^{-1} \nabla_{\vx}  \log p(\vx ). 
$

We now extend this idea to SVGD, for which matrix-valued kernels show up naturally as a consequence of change of variables. 
This justifies the use of matrix-valued kernels and provides guidance on the practical choice of matrix-valued kernels. 
We start with a basic result 
of 
how matrix-valued kernels change 
under 
change of variables (see  \citet{paulsen2016introduction}). 

\begin{lem}
Assume $\H_0$ is an RKHS with a matrix kernel $\vv K_0 \colon 
\RR^d\times \RR^d \to \RR^{d\times d}$.    
Let $\H$ be the set of functions formed by 
$$
\ff(\vx) = \vv M(\vx) \ff_0(\vv t(\vx)),
~~~~~~\forall \ff_0\in \H_0, 
$$ 
where $\vv M\colon \RR^{d}\to \RR^{d\times d}$ is 
a fixed matrix-valued function and we assume $\vv M(\vx)$ is an invertible matrix for all $\vx$, and $\vv t\colon \X \to \X$ is a fixed continuously  differentiable one-to-one transform on $\X$. 

For $ \forall \ff ,  \ff' \in \H$, we can identity an unique $\ff_0, \ff_0' \in \H_0$ such that  
$\ff(\vx ) = \vv M(\vx) \ff_0(\vv t(\vx))$ and $\ff'(\vx ) = \vv M(\vx) \ff_0'(\vv t(\vx))$. Define the inner product on $\H$ via $\langle \ff, \ff'\rangle_{\H} = \langle \ff_0, \ff_0'\rangle_{\H_0}$, 
then $\H$ is also a  vector-valued RKHS, whose matrix-valued kernel is 
$$
{\vv K}(\vx,\vx') = \vv M(\vx) {\vv K}_0(\vv t(\vx),\vv t(\vx'))\vv M(\vx')^\top. 
$$
\label{lem:transform_RKHS}
\end{lem}
%

We now present a key result, which characterizes the change of kernels when we apply invertible variable transforms on the SVGD trajectory.
\newcommand{\bbd}{0}
\begin{thm}\label{thm:main}
i) 
Let $p$ and $q$ be two distributions and $p_{\bbd}$, $q_{\bbd}$ the distribution of $ \vx_{\bbd}  = \vv t(\vx)$ when $\vx$ is drawn from $p$, $q$, respectively, where $\vv t$ is a continuous differentiable one-to-one map on $\X$.
Assume $p$ is a continuous differentiable density with Stein operator $\steinp$, and $\mathcal P_{\bbd}$ the Stein operator of $p_{\bbd}$. 
We have 
\begin{align}\label{equ:svgdchange}
\E_{ \vx \sim q_{\bbd}} [\steinp_{{\bbd}}^\top \ff_\bbd (\vx)] = 
\E_{\vx\sim q} [\steinp^\top   \ff (\vx)], && 
\text{with} 
&&
 \ff(\vx) := \nabla \vv  t(\vx)^{-1}\ff_\bbd(\vv t (\vx)),
\end{align}
where $\nabla \vv t$ is the Jacobian matrix of $\vv t$. 

ii) 
Therefore, 
in the asymptotics of infinitesimal step size ($\epsilon\to 0^+$), running SVGD with kernel $\kk_0$ on $p_{\bbd}$  
is equivalent to running SVGD on $p$ with kernel 
$$
\kk(\vx, \vx') = 
\nabla \vv t (\vx)^{-1} \kk_0(\vv t(\vx),\vv t(\vx')) \nabla \vv t (\vx' )^{-\top}, 
$$
in the sense that the trajectory of these two SVGD can be mapped to each other by the one-to-one map $\vv t$ (and its inverse).
\end{thm} 
%

\subsection{Practical Choice of Matrix-Valued Kernels} 
Theorem~\ref{thm:main}    
suggests a conceptual procedure 
for constructing proper matrix kernels to  incorporate desirable preconditioning  information:  
one can construct a one-to-one map $\vv t$ so  that the distribution $p_{\bbd}$ of $\vx_{\bbd} = \vv t(\vx)$ is an easy-to-sample distribution with a simpler kernel $\kk_0 (\vx, \vx')$, which can be a standard scalar-valued kernel or a simple diagonal kernel. 
Practical choices of $\vv t$ often involve rotating $\vx$ with either Hessian matrix or Fisher information, allowing us to incorporating these information into SVGD.
In the sequel, we first illustrate this idea for simple Gaussian cases and then discuss practical approaches for non-Gaussian cases.

\paragraph{Constant Preconditioning Matrices} 
Consider the simple case when 
$p$ is  multivariate Gaussian, e.g., 
$\log p(\vx) = - \frac{1}{2} \vx^\top \vv Q\vx+const$, 
where $\vv Q$ is a positive definite matrix. 
In this case, 
the distribution $p_{\bbd}$ of 
the transformed variable 
$\vv t(\vx) = \vv Q^{1/2} \vx$ 
is the standard Gaussian distribution 
that can be better approximated with a simpler kernel $\kk_0(\vx,\vx')$, which can be chosen to be  the standard RBF kernel suggested in \citet{liu2016stein},   the graphical kernel suggested in \citet{wang2018stein}, or the linear kernels suggested in \citet{liu2018stein}. 
Theorem~\ref{thm:main} then suggests 
to use a kernel of form 
\begin{align} \label{equ:kkq11}
\kk_{\vv Q}(\vx, \vx') := 
\vv Q^{-1/2} \kk_0\left (\vv Q^{1/2}\vx,~~ \vv Q^{1/2}\vx' \right   )\vv Q^{-1/2},  
\end{align}
in which $\vv Q$ is applied on both the input $\vx$ and the output side.  
As an example, taking $\kk_0$ to be the scalar-valued Gaussian RBF kernel gives 
\begin{align}\label{equ:kqrbf}
\kk_{\vv Q}(\vx, \vx') = 
\vv Q^{-1} \exp\left(- \frac{1}{2h}
||\vx-\vx'||_{\vv Q}^2 \right),
\end{align}
where  $
||\vx-\vx'||_{\vv Q}^2 := (\vx - \vx')^{\top} \vv Q (\vx - \vx')$ and  $h$ is a bandwidth parameter. 
Define $ \kk_{0,\vv Q}(\vx,\vx') := \kk_0\big (\vv Q^{1/2}\vx,~~ \vv Q^{1/2}\vx' \big )$. 
One can show that the SVGD direction 
of the kernel in \eqref{equ:kkq11} equals 
\begin{align}
    \ff_{\kk_{\vv Q}}^*(\cdot) =\vv 
    Q^{-1} \E_{\vx\sim q}[\nabla \log p(\vx) \kk_{0,\vv Q}(\cdot,\vx)  +  \kk_{0,\vv Q}(\cdot,\vx) \nabla_{\vx} ]  
    =\vv Q^{-1} \ff_{\kk_{0,\vv Q}}^*(\cdot), 
\label{equ:seq_svgd}
\end{align}
which is a linear transform of the SVGD direction 
of kernel $ \kk_{0,\vv Q}(\vx,\vx')$ 
with 
matrix $\vv Q^{-1}.$  

In practice, when $p$ is non-Gaussian, we can construct $\vv Q$ by taking averaging over the particles. For example, denote by $\vv H(\vx) = -\nabla_\vx^2 \log p(\vx)$ the negative Hessian matrix at $\vx$, we can construct $\vv Q$ by 
\begin{align}\label{equ:avgQ}
\vv Q = \sum_{i=1}^n \vv H(\vx_i)/n,
\end{align}
where $\{\vx_i\}_{i=1}^n$ are the particles from the previous iteration. We may replace $\vv H$ with the Fisher information matrix to obtain a natural gradient like variant of SVGD.  

\paragraph{Point-wise Preconditioning}  
A constant preconditioning matrix 
can not reflect different curvature or geometric information at different points. 
A simple heuristic to address this limitation is to replace the constant matrix $\vv Q$ with a point-wise matrix function $\vv Q(\vx)$; this motivates a kernel of  form 
\begin{align*}
\kk(\vx,\vx') = 
\vv Q^{-1/2}(\vx) 
\kk_0\big (\vv Q^{1/2}(\vx)\vx,~~ 
\vv Q^{1/2}(\vx')\vx'\big )
\vv Q^{-1/2}(\vx').
\end{align*} 
Unfortunately, this choice may yield 
expensive 
computation and difficult implementation 
in practice,  
because the SVGD update involves taking the gradient of the kernel $\kk(\vx,\vx')$, which would need to differentiate through matrix valued function $\vv Q(\vx)$. When $\vv Q(\vx)$ equals the Hessian matrix, for example, it involves taking the third order derivative of $\log p(\vx)$, yielding an unnatural algorithm. 

\paragraph{Mixture Preconditioning} 
We instead propose a more practical approach 
to achieve point-wise preconditioning with a much simpler algorithm. 
The idea is to use a weighted combination of several constant preconditioning matrices. 
This involves leveraging a set of \emph{anchor points} $\{\vv z_\ell\}_{\ell=1}^m\subset \RR^d$,
each of which is associated with a preconditioning matrix $\vv Q_\ell = \vv Q(\vv z_\ell)$ (e.g., their Hessian or Fisher information matrices). 
 In practice, the anchor points $\{\vv z_\ell\}_{\ell=1}^m$ can be conveniently set to be  the same as the particles $\{\vv x_i\}_{i=1}^n$. 
We then construct a kernel by 
\begin{align} \label{equ:kmixture}
\kk(\vx,\vx') = \sum_{\ell=1}^m  \kk_{\vv Q_\ell}(\vx,\vx') w_\ell(\vx) w_\ell(\vx'), 
\end{align}
where 
$\kk_{\vv Q_\ell}(\vx, \vx')$ 
is defined in \eqref{equ:kkq11} or \eqref{equ:kqrbf},    
and $w_\ell(
\vv x)$ is a positive scalar-valued function that decides the contribution of kernel $\kk_{\vv Q_\ell}$ 
at point $\vx$. 
Here $w_\ell(\vx)$ should be viewed as a mixture probability, and hence  should satisfy $\sum_{\ell=1}^m w_\ell(\vx) = 1$ for all $\vv x$. 
In our empirical studies, we take $w_{\ell}(\vx)$ 
as the Gaussian mixture probability from the  anchor points: 
\begin{align} \label{equ:well}
w_\ell(\vv x) = \frac{\N(\vv x; \vv z_\ell, \vv Q_\ell^{-1})}{\sum_{\ell'=1}^m\N(\vv x; \vv z_{\ell'}, \vv Q_{\ell'}^{-1})}, 
~~~~~~~~\N(\vv x; \vv z_\ell, \vv Q_\ell^{-1}):= 
\frac{1}{Z_\ell}
\exp\left (-\frac{1}{2}\norm{\vv x - \vv z_\ell}_{\vv Q_\ell}^2 \right ), 
\end{align}
where $Z_\ell = (2\pi)^{d/2}\det(Q_\ell)^{-1/2}$. 
In this way, each point $\vx $ is mostly influenced by the anchor point closest to it, 
allowing us to apply different preconditioning for different points. 
Importantly, 
the SVGD update direction related to the  kernel in  \eqref{equ:kmixture} has a simple and easy-to-implement form:  
\begin{align}
\ff_{\kk}^*(\cdot ) 
 = \sum_{\ell=1}^m w_\ell(\cdot ) \E_{\vx \sim q}  
\left [ (w_\ell(\vx) \kk_{\vv Q_\ell} (\cdot,\vx)) \steinp \right ] 
= \sum_{\ell=1}^m w_\ell(\cdot)  \ff_{w_\ell \kk_{\vv Q_\ell}}^*(\cdot), \label{equ:mixsvgd}
\end{align}
which is a weighted sum of a number of SVGD directions with constant preconditioning matrices  
(but with an asymmetric kernel $w_\ell(\vx)\kk_{\vv Q_\ell}(\cdot,\vx))$. 

\paragraph{A Remark on Stein Variational Newton (SVN)}
\citet{detommaso2018stein} provided a Newton-like variation of SVGD. It is motivated by an intractable functional Newton framework, 
and arrives a practical algorithm using a series of approximation steps. 
The update of SVN is 
\begin{align} \label{equ:svn}
\vx_i \gets \vx_i +\epsilon \tilde{\vv H}_i^{-1} \ff_k^*(\vx_i), ~~~\forall  i = 1,\ldots, n,
\end{align}
where $\ff_k^*(\cdot)$ is the standard SVGD gradient, and $\tilde {\vv H}_i$ is a Hessian like matrix associated with particle $\vx_i$,  defined by 
$$
\tilde{\vv H}_i = \E_{\vx \sim q}\left [ \vv H(\vx) k(\vx, \vx_i)^2 + (\nabla_{\vx_i} k(\vx, \vx_i))^{\otimes 2} 
\right ],
$$
where $\vv H(\vx) = -\nabla_{\vx}^2\log p(\vx)$, and 
$\vv w^{\otimes 2} := \vv w \vv w^\top$.  
Due to the approximation introduced in the derivation of SVN, it does not correspond to a standard functional gradient flow like SVGD (unless $\tilde{\vv H}_i = \vv Q$ for all $i$,
in which case it reduces to using  a constant preconditioning matrix on SVGD like \eqref{equ:seq_svgd}).  
SVN can be heuristically viewed as a ``hard'' variant of \eqref{equ:mixsvgd}, which assigns each particle with its own preconditioning matrix with probability one,  
but the mathematical form do not match precisely.  
%
%
%
On the other hand, 
it is useful to note that the set of fixed points of SVN update \eqref{equ:svn} is the \emph{identical to} that of 
the standard SVGD update with $\ff_{k}^*(\cdot)$, once all $\tilde{\vv H}_i$ are positive definite matrices. 
This is because at the fixed points of \eqref{equ:svn}, 
we have $ \tilde{\vv H}_i^{-1} \ff_k^*(\vx_i) =0$ for $\forall i=1,\ldots, n$, which is equivalent to $ \ff_k^*(\vx_i) =0, \forall i$ when all the $\tilde{\vv H}_i$, $\forall i$ are positive definite. 
Therefore, SVN can be justified as an alternative 
fixed point iteration method  
to achieve 
the same set of fixed points as the standard SVGD.  

%% file: tex/experiments.tex
\section{Experiments}

\newcolumntype{L}[1]{>{\raggedright\arraybackslash}p{#1}}
\newcolumntype{C}[1]{>{\centering\arraybackslash}p{#1}}
\newcolumntype{R}[1]{>{\raggedleft\arraybackslash}p{#1}}
\newcommand{\wsgv}{.13}
\begin{figure*}[ht]
    \centering
    \setlength\tabcolsep{1pt} 
    \begin{tabular}{C{21mm}C{18mm}C{18mm}C{18mm}C{18mm}C{35mm}C{28mm}}
    \raisebox{0.5em}{\rotatebox{90}{\scriptsize Iteration=30}} 
    \includegraphics[height = \wsgv\textwidth]{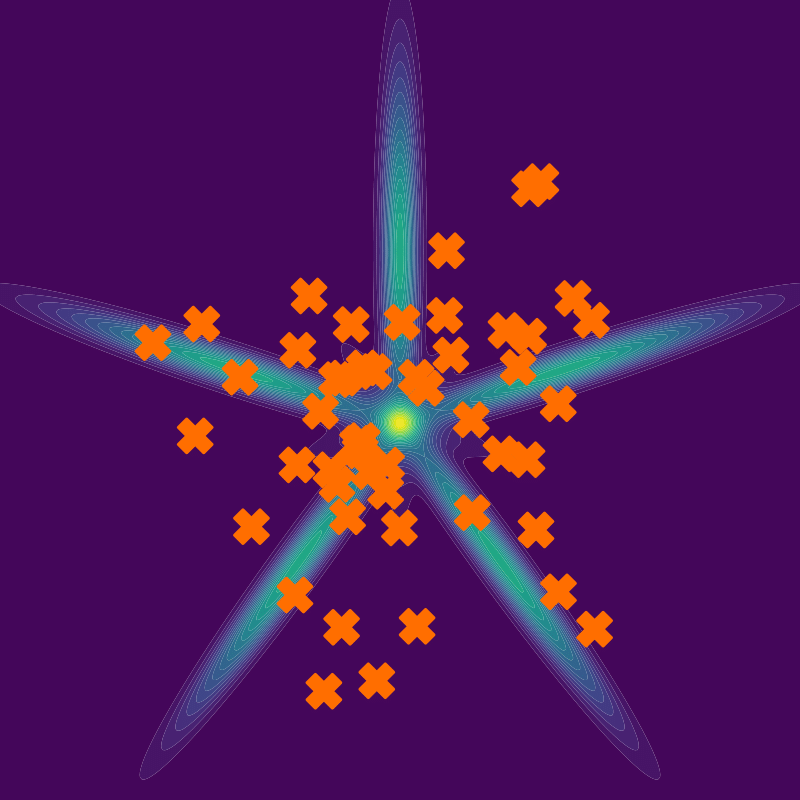} &
    \includegraphics[height = \wsgv\textwidth]{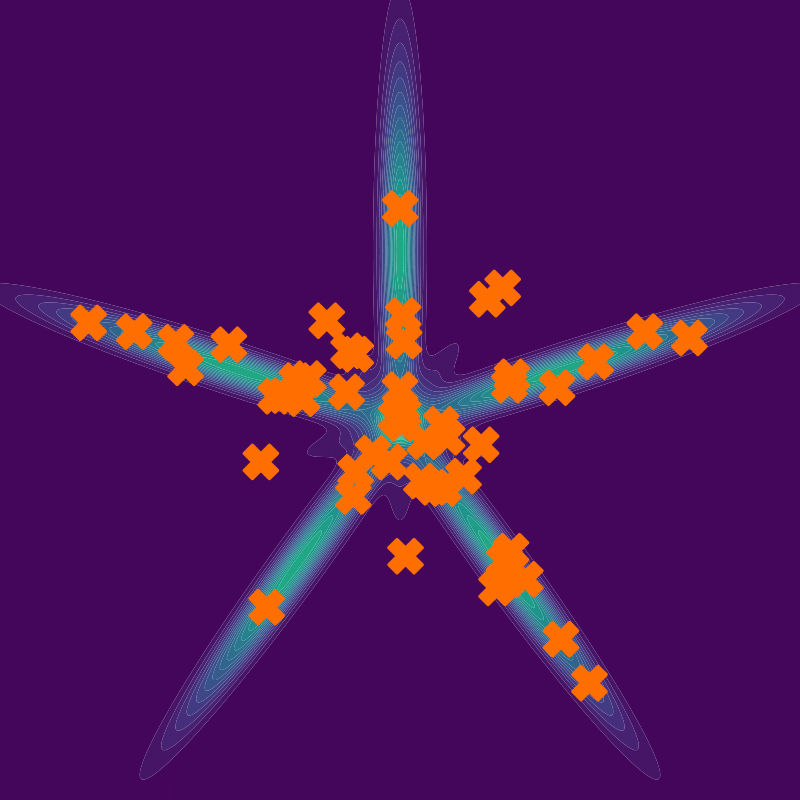} &
    \includegraphics[height = \wsgv\textwidth]{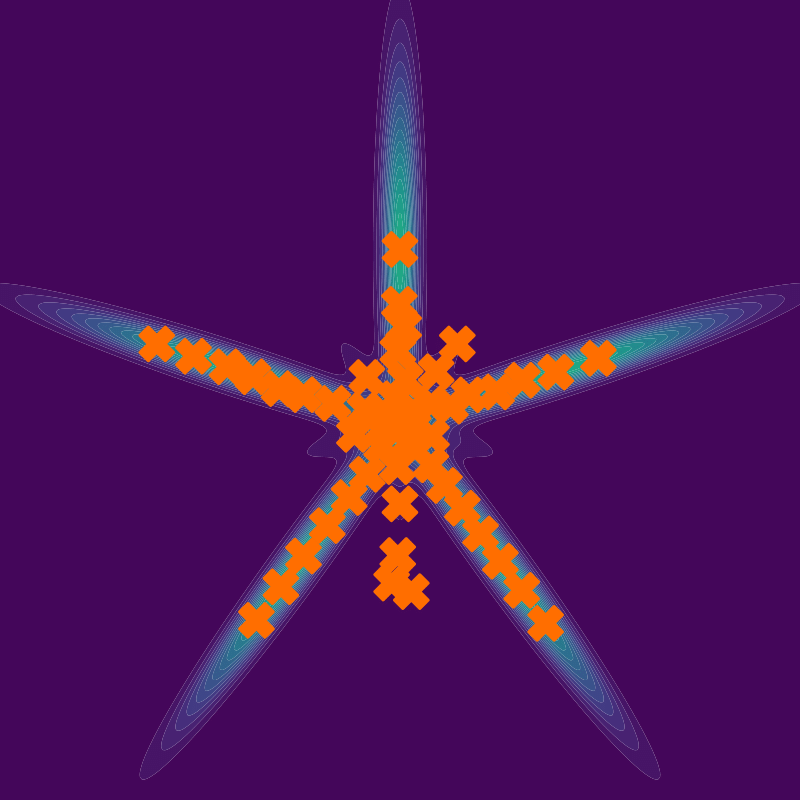} &
    \includegraphics[height = \wsgv\textwidth]{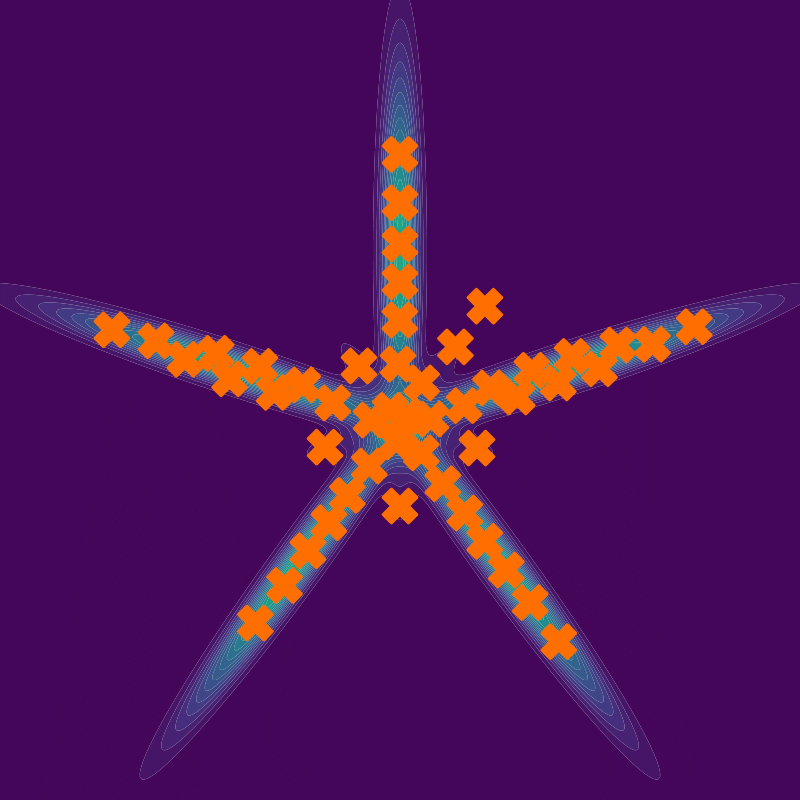} &
    \includegraphics[height = \wsgv\textwidth]{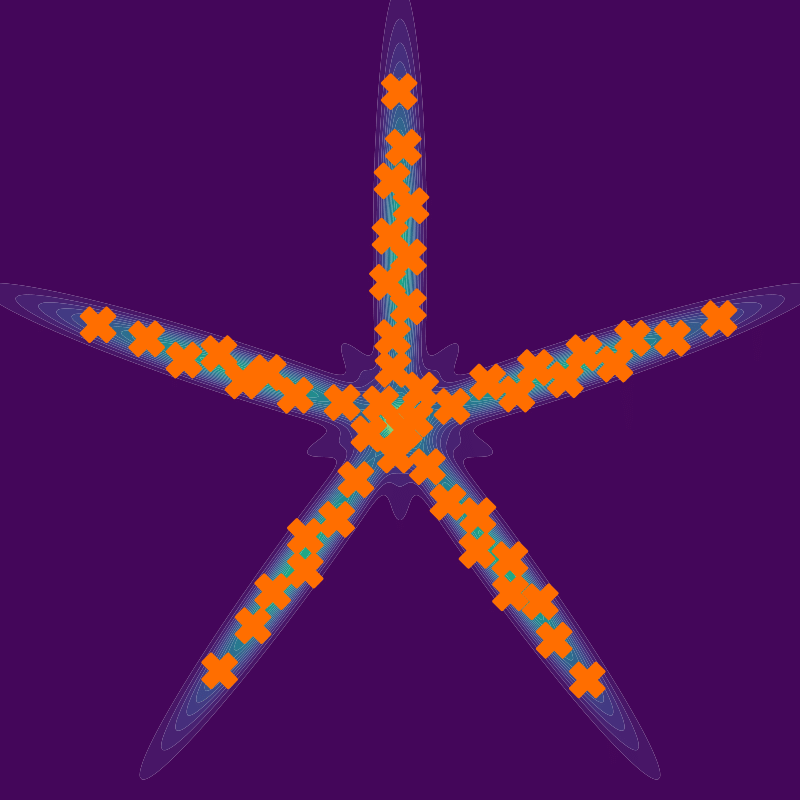} &
    \raisebox{1.0em}{\rotatebox{90}{\scriptsize Log MMD}} 
    \raisebox{-0.3em}{
    \includegraphics[height= \wsgv\textwidth]{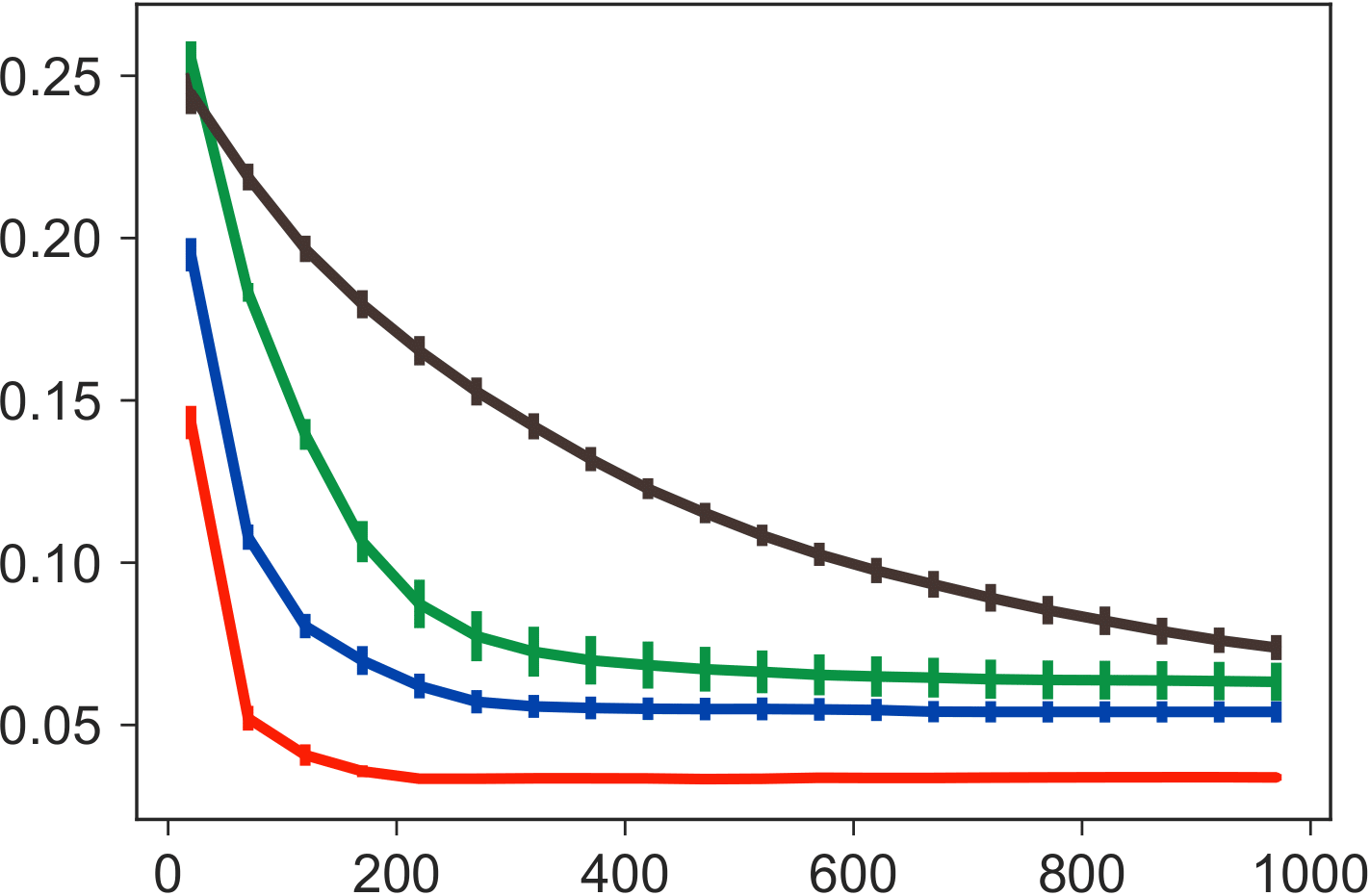}} &
    \hspace{-11em}\raisebox{2.1em}{\includegraphics[width = 0.17\textwidth]{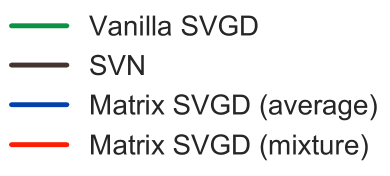}}\\
     {\scriptsize (a) Initializations} &  {\scriptsize (b) Vanilla SVGD} &    { \scriptsize (c) SVN } &   {\scriptsize (d) Matrix-SVGD (average)}  &   {\scriptsize    (e) Matrix-SVGD (mixture)} & {\scriptsize (f) Iteration} \\
    \end{tabular}
    \caption{Figure (a)-(e) show the particles obtained by various methods at the 30-th iteration. Figure (f) plots the log MMD \citep{gretton2012kernel} vs. training iteration starting from the 10-th iteration. 
    We use the standard RBF kernel for evaluating MMD.
  } 
    \label{fig:2d_toy}
\end{figure*}





We demonstrate the effectiveness of our matrix SVGD on various practical tasks. We start with a toy example and then proceed to more challenging tasks that involve logistic regression, neural networks and recurrent neural networks.  
For our method,
we take the preconditioning matrices to be 
either Hessian or Fisher information matrices, depending on the application. 
For large scale Fisher matrices in (recurrent) neural networks, we leverage the 
Kronecker-factored (KFAC) approximation
by \citet{martens2015optimizing, martens2018kronecker} to enable efficient computation.   
We use RBF kernel for vanilla SVGD.
The 
kernel $\kk_0(\vv x,\vv x')$ in our matrix SVGD (see \eqref{equ:kkq11} and \eqref{equ:kqrbf}) is also taken to be Gaussian RBF. 
Following \citet{liu2016stein}, 
we choose the bandwidth of the Gaussian RBF kernels using the  standard median trick and use Adagrad \citep{duchi2011adaptive} for stepsize. 
{Our code is available at \url{https://github.com/dilinwang820/matrix_svgd}.}


The algorithms we test are summarized here: 

\noindent {\tt Vanilla SVGD}, using the code by \citet{liu2016stein}; 

\noindent {\tt Matrix-SVGD (average)}, 
using the constant preconditioning matrix kernel in \eqref{equ:kqrbf}, with $\vv Q$ to be either the average of the Hessian matrices or Fisher matrices of the particles (e.g., \eqref{equ:avgQ});

\noindent {\tt Matrix-SVGD (mixture)}, 
using the mixture preconditioning matrix kernel in \eqref{equ:kmixture}, where we pick the anchor points to be particles themselves, 
that is, $\{\vv z_\ell\}_{\ell=1}^m = \{\vv x_i\}_{i=1}^n$; 

\noindent {\tt Stein variational Newton (SVN)},  based on the implementation of \citet{detommaso2018stein};

\noindent {\tt Preconditioned Stochastic Langevin} {\tt Dynamics} {\tt (pSGLD)}, which is a variant of SGLD \citep{li2016preconditioned}, using a diagonal approximation of Fisher information as the preconditioned matrix. 


\subsection{Two-Dimensional Toy Examples} 
\paragraph{Settings} 
We start with illustrating our method using 
a Gaussian mixture toy model (Figure~\ref{fig:2d_toy}), 
with exact Hessian matrices for preconditioning.  
For fair comparison, we search the best learning rate for all algorithms exhaustively. 
We use 50 particles for all the cases.
We use the same initialization for all methods with the same random seeds.

\paragraph{Results} 
Figure \ref{fig:2d_toy} show the results for 2D toy examples.
Appendix~\ref{sec:moretoy} shows more visualization and results on more examples. 
We can see that methods with Hessian information generally converge faster than vanilla SVGD,
and {\tt Matrix-SVGD (mixture)} yields the best performance.

\begin{figure*}[t]
\centering
\setlength\tabcolsep{1pt} 
\begin{tabular}{cccc}
{\small (a) Covtype} & {\small (b) Covtype} & {\small (c) Protein} & {\small (d) Protein} \\
\raisebox{1.4em}{\rotatebox{90}{\scriptsize Test Accuracy}}
\includegraphics[height=0.155\textwidth]{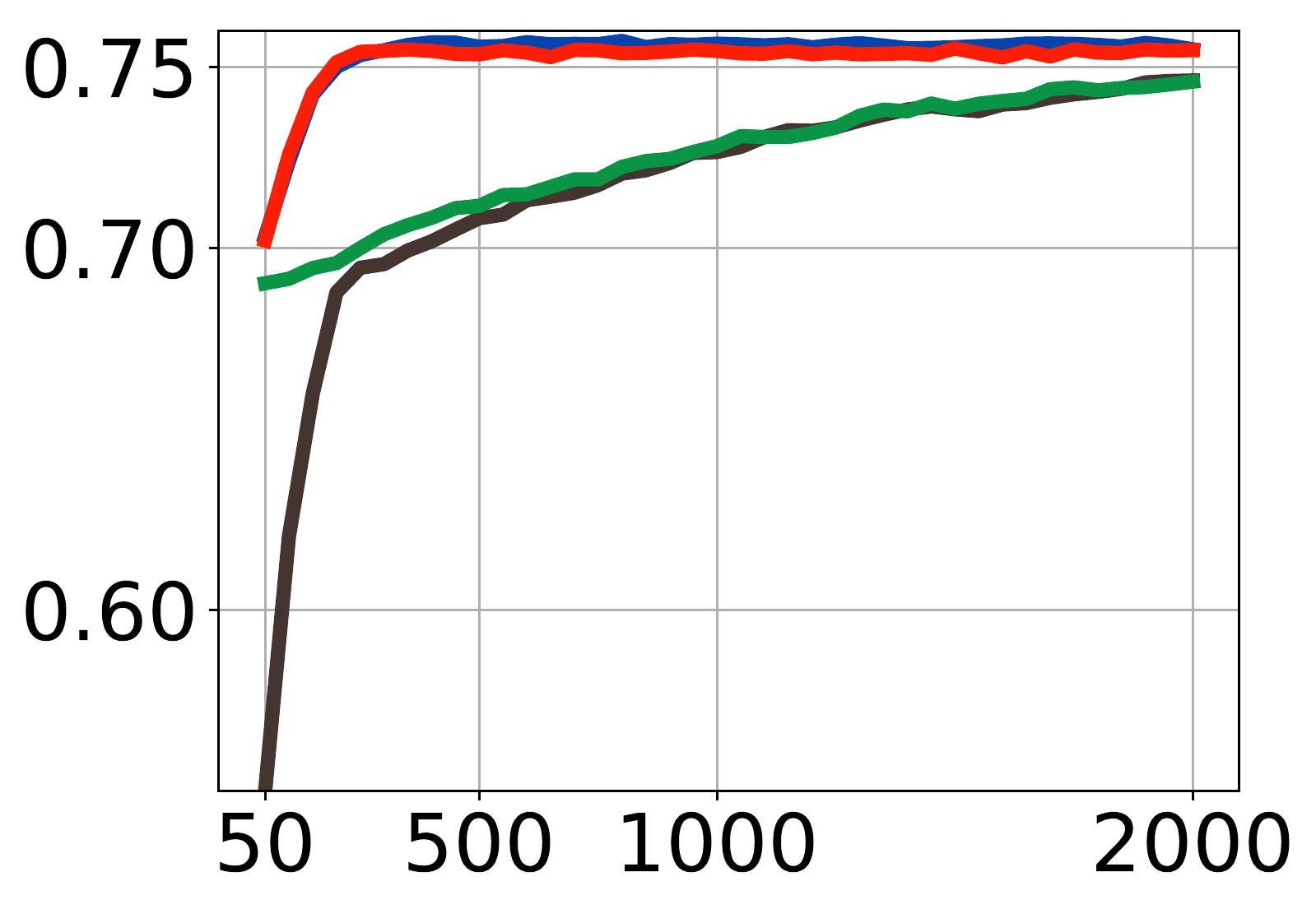}  &
\raisebox{0.6em}{\rotatebox{90}{\scriptsize Test Log-Likelihood}}
\includegraphics[height=0.155\textwidth]{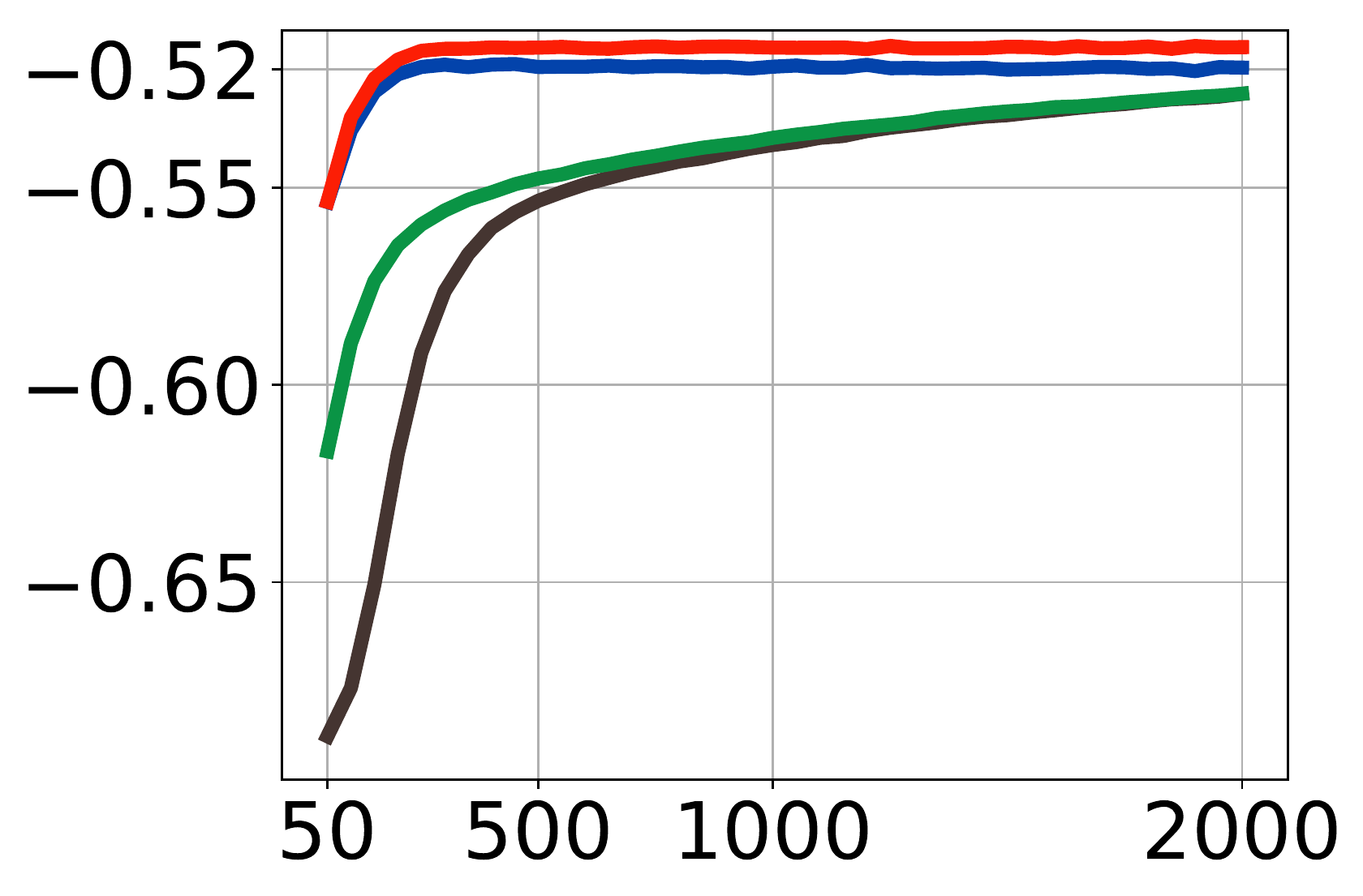}  
&  
\raisebox{2.1em}{\rotatebox{90}{\scriptsize Test RMSE}}
\includegraphics[height=0.155\textwidth]{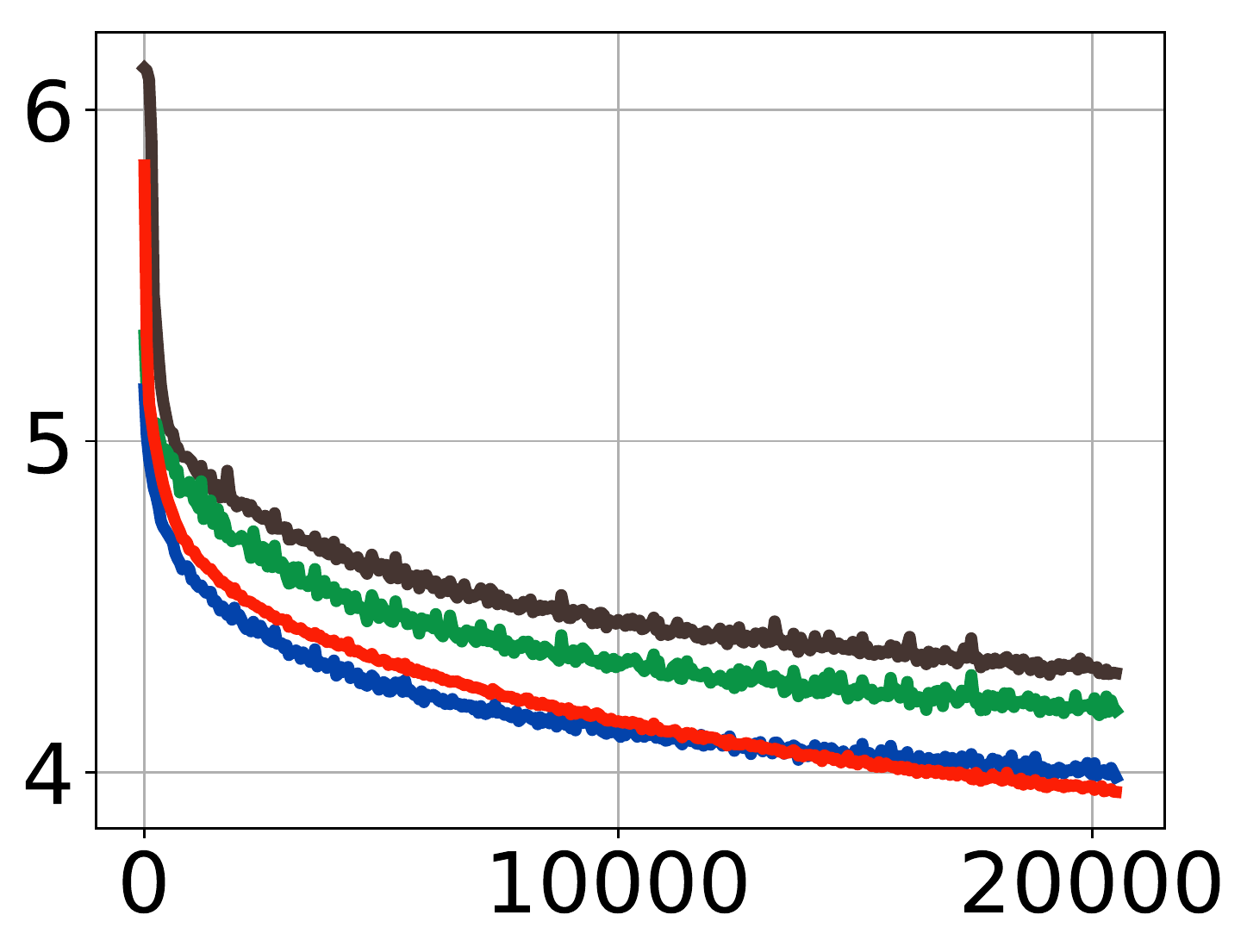}  &
\raisebox{0.6em}{\rotatebox{90}{\scriptsize Test Log-Likelihood}}
\includegraphics[height=0.155\textwidth]{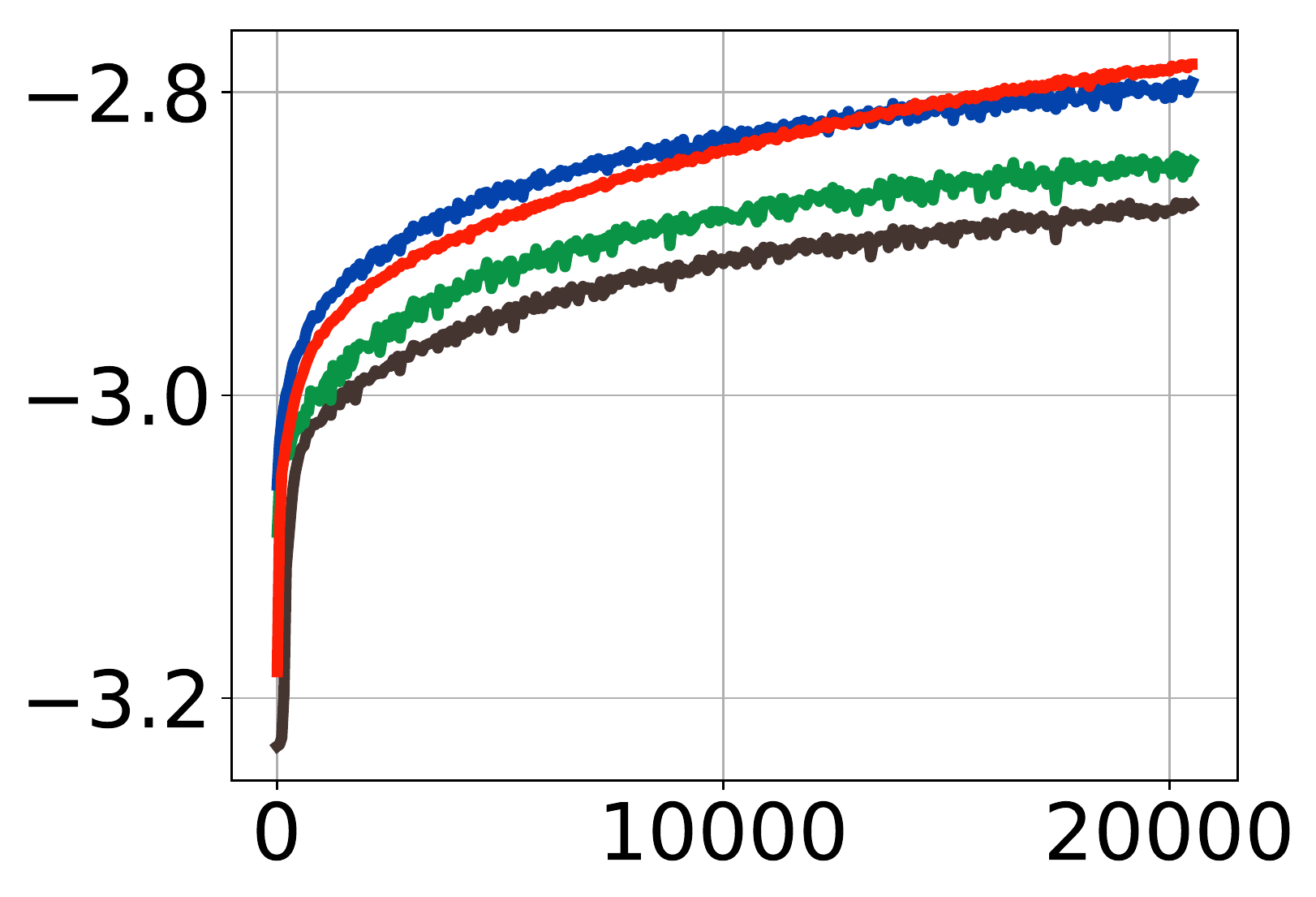} 
\hspace{-6.7em}
\raisebox{0.9em}{\includegraphics[width=0.17\textwidth]{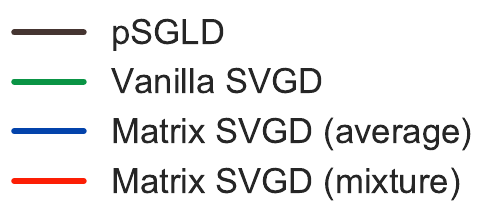}} 
\\ 
{\small ~~~~~~\# of Iterations } &  ~~~~~~ {\small \# of Iterations} & {\small \# of Iterations} & {\small \# of Iterations}  \\
\end{tabular}
\vspace{-1em}
\caption{(a)-(b) Results of Bayesian Logistic regression on the Covtype dataset. 
(c)-(d) Average test RMSE and log-likelihood vs. training batches  on the Protein dataset
for Bayesian neural regression.}
\label{fig:lr_nn_covergence}
\end{figure*}

\subsection{Bayesian Logistic Regression}

\paragraph{Settings} 
We consider the Bayesian logistic regression model for binary classification.
Let $\vv D = \{(\vv x_j, y_j)\}_{j=1}^N$ be a dataset with feature vector $\vv x_j$
and binary label $y_j\in\{0, 1\}$. 
The distribution of interest is 
\begin{align*}
p(\vv \theta ~|~ \vv D) \propto p(\vv D~|~\vv \theta)p(\vv \theta)
&&
\text{with}
&&
p(\vv D~|~\vv \theta) = \prod_{j=1}^N\left [y_j \sigma(\vv \theta^\top \vv x_j) + (1 - y_j)\sigma(-\vv \theta^\top \vv x_j)\right],
\end{align*}
where $\sigma(z) := 1/ (1+\exp(-z))$, and  $p_0(\vv \theta)$ is the prior distribution,
which we set to be standard normal $\normal(\vv \theta; \vv 0, \vv I)$. 
The goal is to approximate the posterior distribution $p(\vv \theta ~|~\vv D)$ with a set of particles $
\{\vv \theta_i\}_{i=1}^n$, and then use it to predict the class labels for testing data points.  
We compare our methods with preconditioned stochastic gradient Langevin dynamics (pSGLD)
\citep{li2016preconditioned}.
Because pSGLD is a sequential algorithm, 
for fair comparison, 
we obtain the samples of pSGLD 
by running $n$ parallel chains of pSGLD for estimation. 
{The preconditioning matrix in both pSGLD and matrix SVGD 
is taken to be the Fisher information matrix.
}

We consider the binary \textit{Covtype}\footnote{\url{https://www.csie.ntu.edu.tw/~cjlin/libsvmtools/datasets/binary.html}} 
dataset with $581,012$ data points and $54$ features. 
We partition the data into 70\% for training, 10\% for validation and 20\% for testing. 
{We use Adagrad optimizer with a mini-batch size of 256.} We choose the best learning rate from $[0.001, 0.005, 0.01, 0.05, 0.1, 0.5, 1.0]$ for each method on the validation set. 
For all the experiments and algorithms,  we use $n=20$ particles. Results are average over 20 random trials. 

\paragraph{Results} 
Figure~\ref{fig:lr_nn_covergence} (a) and (b) show the test accuracy and test log-likelihood of different algorithms. We can see that 
both {\tt Matrix-SVGD (average)} and {\tt Matrix-SVGD (mixture)} converge much faster than both vanilla SVGD and pSGLD, reaching an accuracy of 0.75 in less than 500 iterations.


\newcolumntype{L}[1]{>{\raggedright\arraybackslash}p{#1}}
\newcolumntype{C}[1]{>{\centering\arraybackslash}p{#1}}
\newcolumntype{R}[1]{>{\raggedleft\arraybackslash}p{#1}}
\begin{table*}[t]
\centering
\scalebox{0.825}{
\begin{small}
\setlength{\tabcolsep}{1pt}
\begin{tabular}{l|C{18mm}C{18mm}C{19mm}C{19mm}|C{18mm}C{18mm}C{19mm}C{19mm}}
\hline
 & \multicolumn{4}{|c|}{Test RMSE} & \multicolumn{4}{|c}{Test Log-Likelihood}\\
\hline
Dataset & pSGLD & Vanilla SVGD & Matrix-SVGD\newline (average) & Matrix-SVGD \newline (mixture) & pSGLD & Vanilla SVGD & Matrix-SVGD \newline(average) & Matrix-SVGD \newline (mixture) \\ 
\hline
Boston & $\pmb{2.699{\scriptstyle \pm 0.155}}$ &$2.785{\scriptstyle \pm 0.169}$ &$2.898{\scriptstyle \pm 0.184}$ &$2.717{\scriptstyle \pm 0.166}$  & $-2.847{\scriptstyle \pm 0.182}$ &$-2.706{\scriptstyle \pm 0.158}$ &$\pmb{-2.669{\scriptstyle \pm 0.141}}$ &$-2.861{\scriptstyle \pm 0.207}$  \\
Concrete & $5.053{\scriptstyle \pm 0.124}$ &$5.027{\scriptstyle \pm 0.116}$ &$4.869{\scriptstyle \pm 0.124}$ &$\pmb{4.721{\scriptstyle \pm 0.111}}$  & $-3.206{\scriptstyle \pm 0.056}$     &$\pmb{-3.064{\scriptstyle \pm 0.034}}$ &$-3.150{\scriptstyle \pm 0.054}$ &$-3.207{\scriptstyle \pm 0.071}$  \\
Energy & $0.985{\scriptstyle \pm 0.024}$ &$0.889{\scriptstyle \pm 0.024}$ &$\pmb{0.795{\scriptstyle \pm 0.025}}$ &$0.868{\scriptstyle \pm 0.025}$  & $-1.395{\scriptstyle \pm 0.029}$ &    $-1.315{\scriptstyle \pm 0.020}$ &$\pmb{-1.135{\scriptstyle \pm 0.026}}$ &$-1.249{\scriptstyle \pm 0.036}$   \\
Kin8nm & $0.091{\scriptstyle \pm 0.001}$ &$0.093{\scriptstyle \pm 0.001}$ &$0.092{\scriptstyle \pm 0.001}$ &$\pmb{0.090{\scriptstyle \pm 0.001}}$  & $~~~0.973{\scriptstyle \pm 0.010}$ &$    ~~~0.964{\scriptstyle \pm 0.012}$ &$~~~0.956{\scriptstyle \pm 0.011}$ &$~~~\pmb{0.975{\scriptstyle \pm 0.011}}$ \\
Naval & $0.002{\scriptstyle \pm 0.000}$ &$0.004{\scriptstyle \pm 0.000}$ &$0.001{\scriptstyle \pm 0.000}$ &$\pmb{0.000{\scriptstyle \pm 0.000}}$  & $~~~4.535{\scriptstyle \pm 0.093}$ &$~~~4.312{\scriptstyle \pm 0.087}$ &$~~~5.383{\scriptstyle \pm 0.081}$ &$~~~\pmb{5.639{\scriptstyle \pm 0.048}}$ \\
Combined & $4.042{\scriptstyle \pm 0.034}$ &$4.088{\scriptstyle \pm 0.033}$ &$4.056{\scriptstyle \pm 0.033}$ &$\pmb{4.029{\scriptstyle \pm 0.033}}$  & $-2.821{\scriptstyle \pm 0.009}$     &$-2.832{\scriptstyle \pm 0.009}$ &$-2.824{\scriptstyle \pm 0.009}$ &$\pmb{-2.817{\scriptstyle \pm 0.009}}$ \\
Wine & $0.641{\scriptstyle \pm 0.009}$ &$0.645{\scriptstyle \pm 0.009}$ &$\pmb{0.637{\scriptstyle \pm 0.008}}$ &$\pmb{0.637{\scriptstyle \pm 0.009}}$  & $-0.984{\scriptstyle \pm 0.016}$ &$-0.997{\scriptstyle \pm 0.019}$ &$\pmb{-0.980{\scriptstyle \pm 0.016}}$ &$-0.988{\scriptstyle \pm 0.018}$   \\
Protein & $4.300{\scriptstyle \pm 0.018}$ &$4.186{\scriptstyle \pm 0.017}$ &$3.997{\scriptstyle \pm 0.018}$ &$\pmb{3.852{\scriptstyle \pm 0.014}}$  & $-2.874{\scriptstyle \pm 0.004}$     &$-2.846{\scriptstyle \pm 0.003}$ &$-2.796{\scriptstyle \pm 0.004}$ &$\pmb{-2.755{\scriptstyle \pm 0.003}}$    \\
Year & $8.630{\scriptstyle \pm 0.007}$ &$8.686{\scriptstyle \pm 0.010}$ &$8.637{\scriptstyle \pm 0.005}$ &$\pmb{8.594{\scriptstyle \pm 0.009}}$  & $-3.568{\scriptstyle \pm 0.002}$ &$-    3.577{\scriptstyle \pm 0.002}$ &$-3.569{\scriptstyle \pm 0.001}$ &$\pmb{-3.561{\scriptstyle \pm 0.002}}$  \\
\hline
\end{tabular}
\end{small}
}
\caption{ Average test RMSE and log-likelihood in test data for UCI regression benchmarks. }
\vspace{-1em}
\label{tbl:bayesnn}
\end{table*}

\subsection{Neural Network Regression}
\paragraph{Settings}  
We apply our matrix SVGD on Bayesian neural network regression on UCI datasets. 
For all experiments, we use a two-layer neural network with 50 hidden units with ReLU activation functions. 
We assign isotropic Gaussian priors to the neural network weights. 
All datasets\footnote{\url{https://archive.ics.uci.edu/ml/datasets.php}}
are randomly partitioned into $90\%$ for training and $10\%$ for testing. 
All results are averaged over 20 random trials, except for Protein and
Year, on which 5 random trials are performed.
We use $n=10$ particles for all methods. 
{We use Adam optimizer with a mini-batch size of 100; for large dataset such as \textit{Year}, we set the mini-batch size to be 1000.} 
{We use the Fisher information matrix   with Kronecker-factored (KFAC)
approximation for preconditioning. 
}

\paragraph{Results} 
Table \ref{tbl:bayesnn} shows the performance in terms of the test RMSE and the test log-likelihood.
We can see that both {\tt Matrix-SVGD (average)} and {\tt Matrix-SVGD (mixture)},  
which use second-order information, achieve  better performance than
vanilla SVGD.
{\tt Matrix-SVGD (mixture)} yields the best performance 
for both test RMSE and test log-likelihood in most cases.
Figure~\ref{fig:lr_nn_covergence} (c)-(d) show that both variants of Matrix-SVGD converge much faster than vanilla SVGD and pSGLD on the Protein dataset.

\subsection{Sentence Classification With Recurrent Neural Networks (RNN)}

\paragraph{Settings} We consider the sentence classification task on four datasets: MR \citep{pang2005seeing}, CR \citep{hu2004mining}, SUBJ \citep{pang2004sentimental}, and MPQA \citep{wiebe2005annotating}. 
\begin{wraptable}{r}{0.5\textwidth}
\centering
\scalebox{0.825}{
\setlength{\tabcolsep}{2pt}
  \begin{tabular}{l|cccc}
    \hline
        Method  & MR & CR & SUBJ & MPQA  \\
        \hline
        SGLD  & 20.52 & 18.65  &~~7.66 & 11.24 \\
        pSGLD  & 19.75  & 17.50 & ~~6.99  & 10.80\\ 
        \hline
        Vanilla SVGD & 19.73 & 18.07 & ~~6.67 & \textbf{10.58} \\
        Matrix-SVGD (average) & 19.22   &  17.29 &~~6.76 & 10.79  \\
        Matrix-SVGD (mixture) & \textbf{19.09} & \textbf{17.13} & ~~\textbf{6.59} & 10.71 \\
        \hline
    \end{tabular}
    }
    \caption{Sentence classification errors measured with four benchmarks.}
    \label{tab:rnn}
\end{wraptable}
We use a recurrent neural network (RNN) based model, $p(y~|~\vv x)=\mathrm{softmax}(\vv w_y^\top \vv h_{RNN}(\vv x,\vv v)),$
where $\vv x$ is the input sentence, 
$y$ is a discrete-valued label of the sentence,  and $\vv w_{y}$ is a weight coefficient related to label class $y$. 
And $\vv h_{RNN}(\vv x,\vv v)$ is an RNN function
with parameter $\vv v$
using a one-layer bidirectional GRU model \citep{cho2014learning} with 50 hidden units. 
We apply matrix SVGD to infer the posterior of $\vv w=\{\vv w_y \colon \forall y\}$, while updating the RNN weights $\vv v$ using typical {stochastic} gradient descent. 
In all experiments, 
we use the pre-processed text data provided in \citet{gan2016scalable}.
For all the datasets, we conduct 10-fold cross-validation for evaluation.
We use $n= 10$ particles for all the methods. 
For training, we use a mini-batch size of 50 and run all the algorithms for 20 epochs with early stop. 
{We use the Fisher information matrix for preconditioning. 
}

\paragraph{Results}
Table~\ref{tab:rnn} shows the results of testing classification errors. 
We can see that {\tt Matrix-SVGD (mixture)} generally performs the best among all algorithms.

%% file: tex/proof.tex
\appendix

\onecolumn
\section{Proof}
\paragraph{Proof of Theorem~\ref{thm:matrix_svgd}}
 Let $\vv e_{\ell}$ be the column vector with $1$ in $\ell^{th}$ coordinate and $0$ elsewhere. 
By the RKHS reproducing property \eqref{equ:matrxi_reproduce}
we have 
\begin{align*} 
\E_{\vx\sim q}\left [ \steinp^\top \ff(\vx) \right] 
& = \E_{\vx\sim q}\left [ \nabla_\vx \log p(\vx)^\top \ff(\vx) + \nabla_\vx^\top \ff(\vx) \right]\\
& = \E_{\vx\sim q}\left [  \ff(\vx)^\top \nabla_\vx\log p(\vx) + \sum_{\ell=1}^d \nabla_{x^\ell} \ff(\vx)^\top \vv e_{\ell} \right]\\
& = \E_{\vx\sim q}\left [  \left \langle\ff(\cdot),~~ \kk(\cdot, \vx)\nabla_\vx\log p(\vx)\right \rangle_{\H_{\kk}} + \sum_{\ell=1}^d \nabla_{x^\ell} \left\langle\ff(\cdot),~~ \kk(\cdot, \vx)\vv e_{\ell}\right \rangle_{\H_{\kk}} \right]\\
& = \left \langle\ff(\cdot),  ~~\E_{\vx\sim q}\left[ \kk(\cdot, \vx)\nabla_\vx\log p(\vx) +  \sum_{\ell=1}^d \nabla_{x^\ell}\kk(\cdot, \vx) \vv e_{\ell} \right]\right \rangle_{\H_{\kk}}\\
& = \left \langle\ff(\cdot),  ~~\E_{\vx\sim q}\left[ \kk(\cdot, \vx)\nabla_\vx\log p(\vx) +  \kk(\cdot, \vx)\nabla_\vx \right]\right \rangle_{\H_{\kk}}\\
& = \left \langle \ff(\cdot), ~ \E_{\vx\sim q}\left [\kk(\cdot, \vx) \steinp \right ] \right \rangle_{\H_{\kk}}, 
\end{align*}
The optimization in \eqref{equ:matrixphi} is hence 
$$
\max_{\ff\in \H_{\kk}}  \big \langle \ff(\cdot), ~ \E_{\vx\sim q}\left [\kk(\cdot, \vx) \steinp \right ]\big \rangle_{\H_{\kk}}, ~~~s.t.~~~ \norm{\ff}_{\H_\kk}\leq 1, 
$$
whose solution is $\ff^*(\cdot) \propto \E_{\vx\sim q}\left [\kk(\cdot, \vx) \steinp \right ]$. 

\paragraph{Proof of Lemma~\ref{lem:transform_RKHS}}
This is a basic result of RKHS, which can be found in classical textbooks such as \citet{paulsen2016introduction}. 
The key idea is to show that $\vv K(\vx,\vx')$ satisfies the reproducing property for $\H.$
Recall the reproducing property of $\H_0$:
$$
\ff_0(\vx)^\top  \vv c  = \langle \ff_0, ~ \vv K_0(\cdot, \vx) \vv c\rangle_{\H_0}, ~~~~\forall \vv c \in \R^d. 
$$
Taking $\ff(\vx) = \vv M(\vx) \phi_0(\vv t(\vx))$, we have 
\begin{align*}
\ff(\vx)^\top  \vv c  
& = \langle \ff_0, ~ \vv K_0(\cdot, \vv t(\vx)) \vv M(\vx)^\top\vv c\rangle_{\H_0} \\
& = \langle \ff, ~ \vv M(\cdot) \vv K_0(\vv t(\cdot), \vv t(\vx)) \vv M(\vx)^\top \vv c\rangle_{\H} \\
& = \langle \ff, ~ \vv K(\cdot, \vx) \vv c\rangle_{\H}, 
\end{align*}
where the second step follows  $\langle \ff, \ff'\rangle_{\H} =\langle \ff_0, \ff_0'\rangle_{\H_0} $
with  $\ff_0'(\cdot) =\vv K_0(\cdot, \vv t(\vx)) \vv M(\vx)^T\vv  c$. 

\paragraph{Proof of Theorem~\ref{thm:main}}    ~~~ 
\begin{proof}
Note that KL divergence is invariant under invertible variable transforms, that is,
\begin{align} \label{equ:klinva}
\KL(q_{[\epsilon \ff]} ~||~ p)  = \KL(q_{[\epsilon\ff]\bbd} ~||~ p_{\bbd}). 
\end{align}
where $p_{\bbd}$ denotes the distribution of $\vx_{\bbd} = \vv t(\vx) $ when $\vx \sim p$, 
and 
$q_{[\epsilon\ff]\bbd}$ denotes the 
distribution of $\vx_0' = \vv t(\vv x')$ 
when $\vv x' \sim q_{[\epsilon\ff]}$. 
Recall that $q_{[\epsilon\ff]}$ is defined as the distribution of $\vv x' = \vv x + \epsilon \ff(\vv x)$ when $\vv x\sim q$. 

Denote by ${\vv t}^{-1}$ the inverse map of $\vv t$, that is, ${\vv t}^{-1}(\vv t(\vx)) = \vx$. 
We can see that $\vv x_0' \sim q_{[\epsilon\ff]\bbd}$ can be obtained by  \begin{align} 
\vx_0' & = \vv t(\vv x')  \ant{$\vv x'\sim q_{[\epsilon\ff]}$} \notag \\
& = \vv t(\vv x + \epsilon \ff(\vv x)) 
\ant{$\vv x\sim q$}  \notag
\\
& = \vv t(\vv t^{-1}(\vv x_0) + \epsilon \ff(\vv t^{-1}(\vv x_0))) 
\ant{$\vv x_0\sim q_0$} \notag
\\
& = \vv x_0 + \epsilon 
\nabla \vv t(
\vv t^{-1}(\vv x_0)) \ff(\vv t^{-1}(\vv x_0))
~+~ \bigO{\epsilon^2} \notag\\
& = \vv x_0 + \epsilon \vv \phi_0(\vv x_0) 
~+~ \bigO{\epsilon^2}, \label{equ:hahahdef}
\end{align}
where we used the definition that $\ff(\vv x) = \nabla\vv t(\vv x)^{-1} \ff_0(\vv t(\vv x))$ in \eqref{equ:svgdchange}, and 
 $\bigO{\cdot}$ is the big-O notation. 

From 
Theorem~3.1 of \citet{liu2016stein}, we have 
$$
\frac{d}{d\epsilon} \KL(q_{[\epsilon \ff]} ~||~ p) \bigg|_{\epsilon=0} = - \E_{q}[\steinp^\top \ff]. 
$$
Using Equation~\eqref{equ:hahahdef} and derivation similar to Theorem~3.1 of \citet{liu2016stein}, we can show 
$$
\frac{d}{d\epsilon}\KL(q_{[\epsilon\ff]\bbd} ~||~ p_{\bbd}) \bigg | _{\epsilon=0} 
= - \E_{q_{\bbd}} [\steinp_{0}^\top \ff_0 ].
$$
Combining these with \eqref{equ:klinva} proves \eqref{equ:svgdchange}.  

Following Lemma~\ref{lem:transform_RKHS}, 
when $\ff_0$ is in $\H_0$ with kernel $\kk_0(\vx,\vx')$, 
$\ff$ is in $\H$ with kernel $\kk(\vx,\vx')$. Therefore, 
maximizing $\E_{q}[\steinp^\top \ff]$ in $\H$ is equivalent to $\E_{q_{0}} [\steinp_{0}^\top \ff_0 ]$  in $\H_0$. This suggests the trajectory of SVGD on $p_0$ with $\kk_0$ and that on $p$ with $\kk$  are equivalent. 

\end{proof}

%% file: tex/appendix_2d_figures.tex
\onecolumn


\section{Toy Examples}\label{sec:moretoy}
Figure~\ref{fig:app_2d_toy} and
Figure~\ref{fig:app_MMD_2d_toy} show results of different algorithms on three 2D toy distributions: Star, Double banana and Sine.  
Detailed information of these distributions 
and more results 
are shown in Section~\ref{sec:sine}-\ref{sec:star}. 


We can see from Figure~\ref{fig:app_2d_toy}-\ref{fig:app_MMD_2d_toy} 
that both variants of matrix SVGD consistently
outperform SVN and vanilla SVGD. 
We also find that 
 {\tt Matrix SVGD(mixture)} tends to outperform 
{\tt Matrix SVGD (average)}, 
which is expected since {\tt Matrix SVGD (average)} uses a constant preconditioning matrix for all the particles, and can not capture different curvatures at different locations. 
{\tt Matrix SVGD (mixture)} yields the best performance in general.

\newcommand{\wsgva}{.16}
\begin{figure*}[ht]
    \centering
    \setlength\tabcolsep{3pt} 
    \begin{tabular}{ccccc}
     \raisebox{0.9em}{\rotatebox{90}{\small (a) Sine}}\hspace{10pt}
    \raisebox{0.em}{\rotatebox{90}{Iteration=30}}\hspace{10pt}
    \includegraphics[width = \wsgva\textwidth]{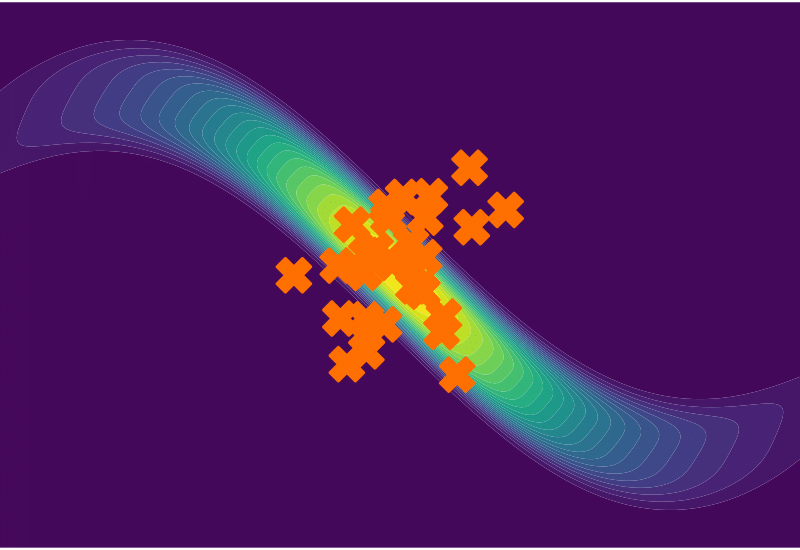} &
    \includegraphics[width = \wsgva\textwidth]{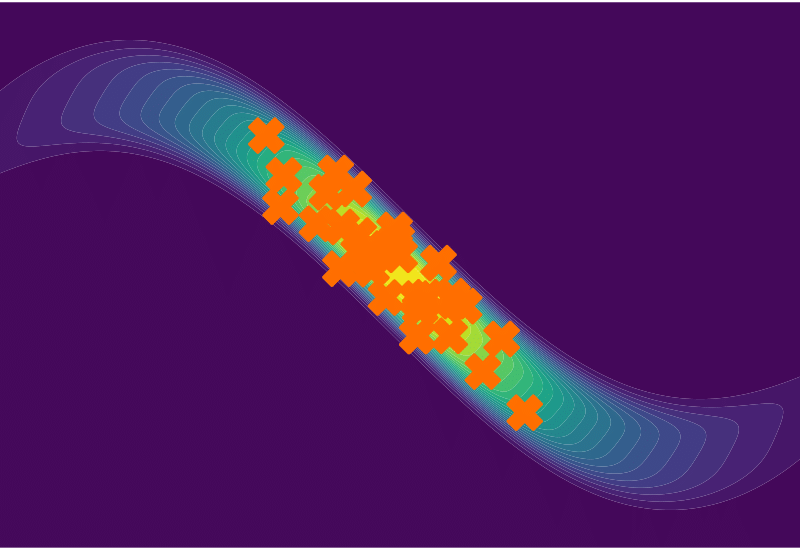} &
    \includegraphics[width = \wsgva\textwidth]{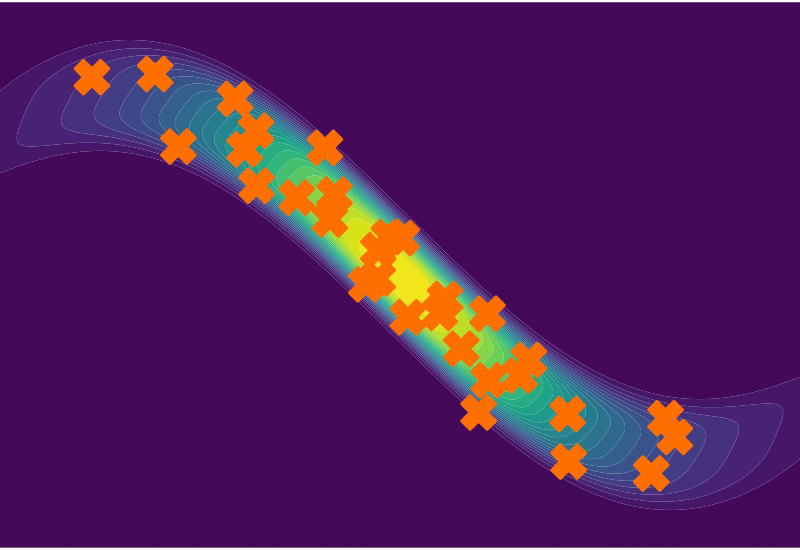} &
    \includegraphics[width = \wsgva\textwidth]{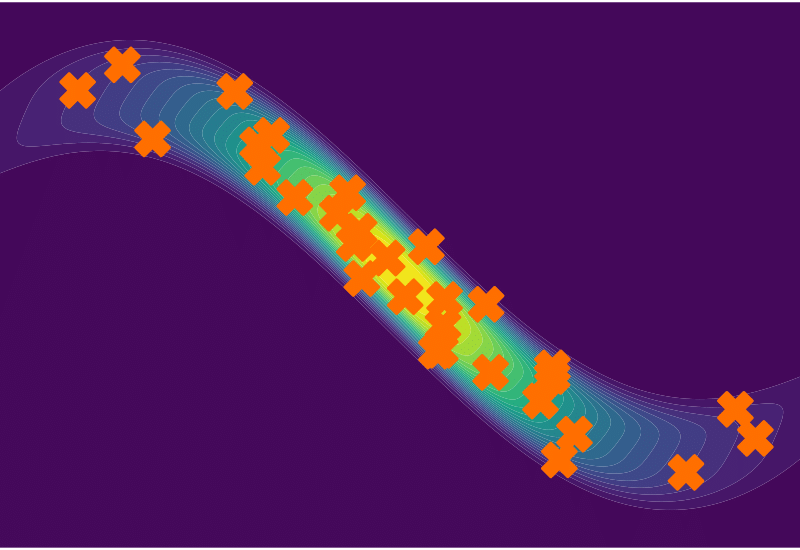} &
    \includegraphics[width = \wsgva\textwidth]{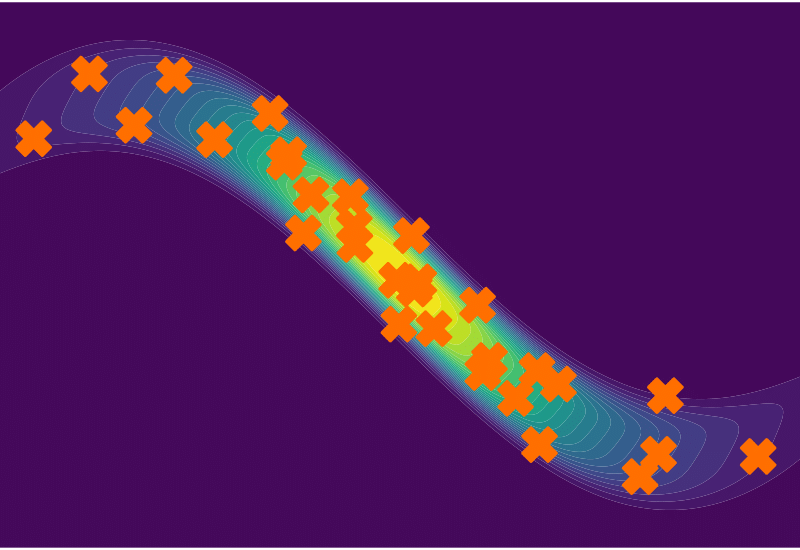} \\
    \raisebox{0.0em}{\rotatebox{90}{\small (b) Double banana}}\hspace{10pt}
    \raisebox{0.4em}{\rotatebox{90}{Iteration=100}}\hspace{10pt}
    \includegraphics[width = \wsgva\textwidth]{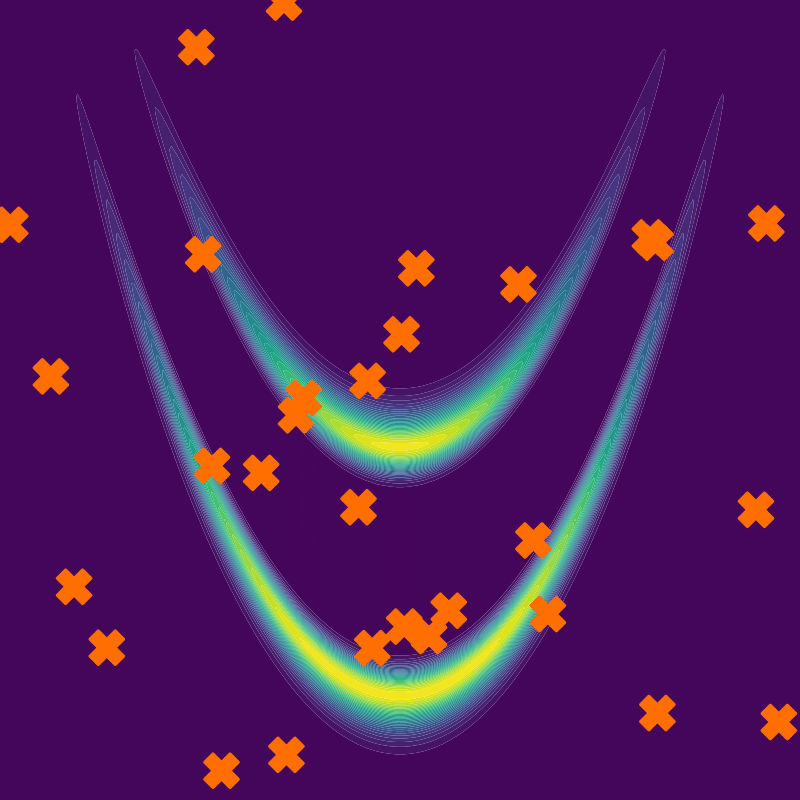} &
    \includegraphics[width = \wsgva\textwidth]{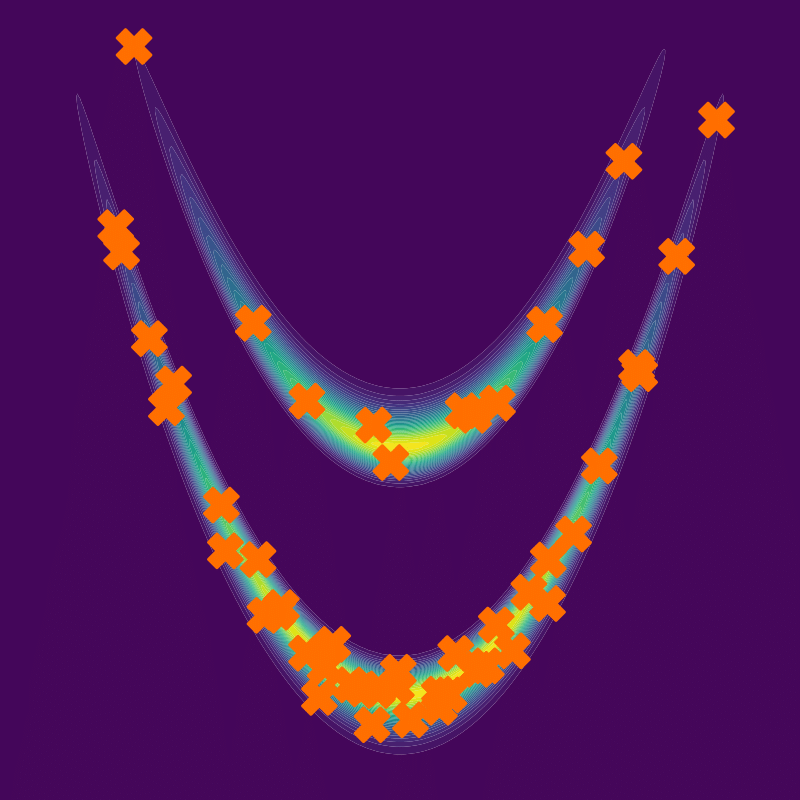} &
    \includegraphics[width = \wsgva\textwidth]{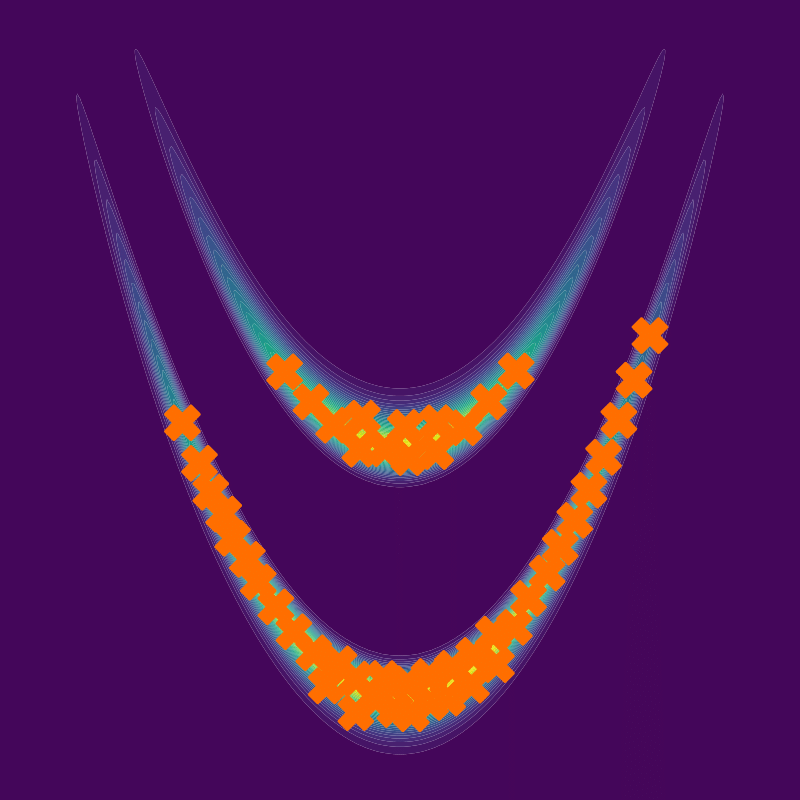} &
    \includegraphics[width = \wsgva\textwidth]{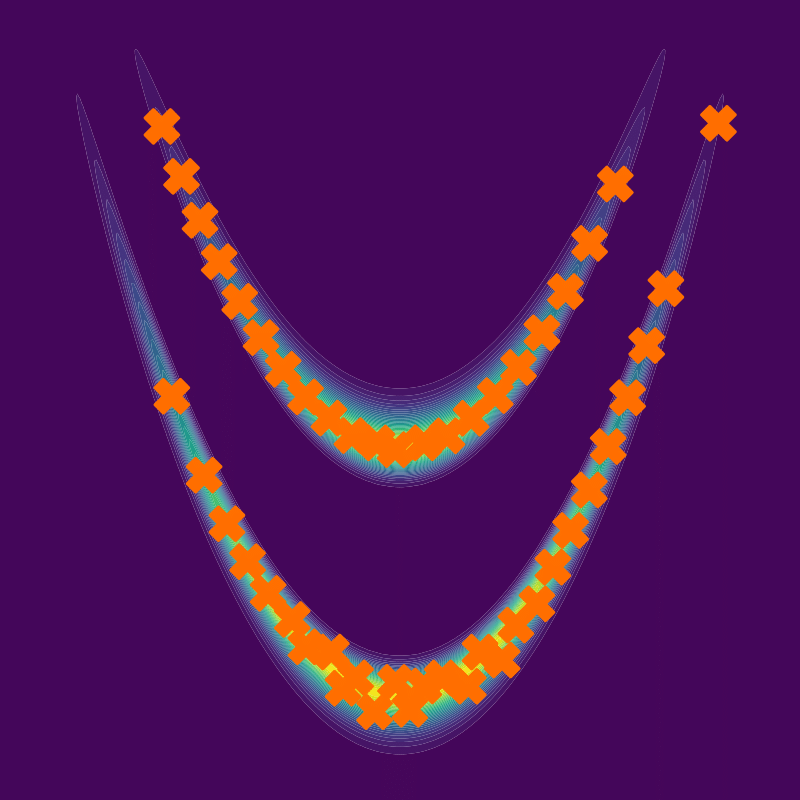} &
    \includegraphics[width = \wsgva\textwidth]{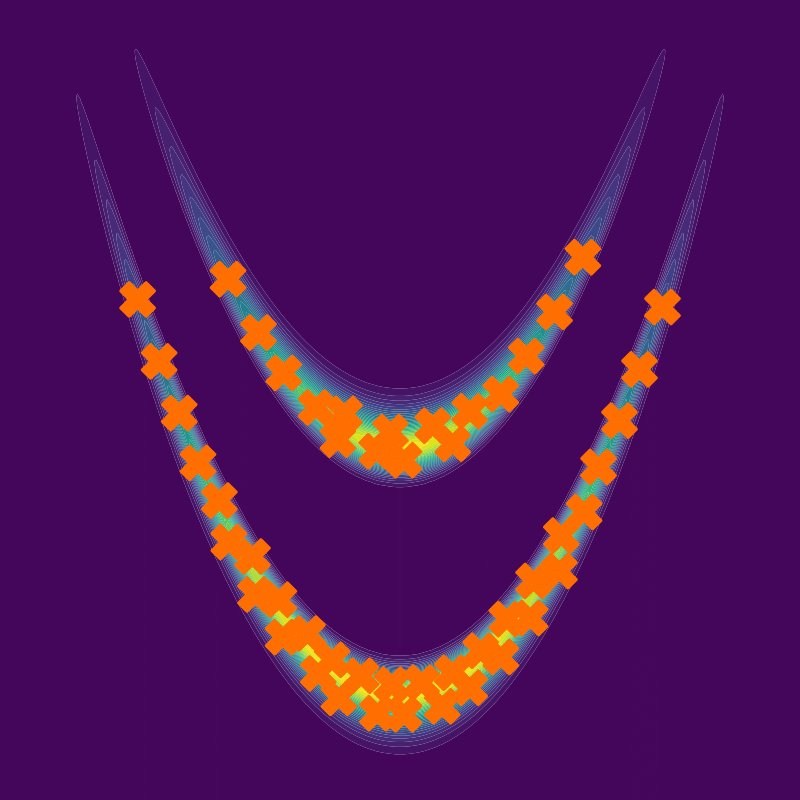} \\
    \raisebox{1.5em}{\rotatebox{90}{\small (c) Star}}\hspace{10pt}
    \raisebox{0.5em}{\rotatebox{90}{Iteration=30}} \hspace{10pt}
    \includegraphics[width = \wsgva\textwidth]{figure/2d_mixture/mixture_iter=0.png} &
    \includegraphics[width = \wsgva\textwidth]{figure/2d_mixture/rbf_iter=50.png} &
    \includegraphics[width = \wsgva\textwidth]{figure/2d_mixture/newton_iter=50.png} &
    \includegraphics[width = \wsgva\textwidth]{figure/2d_mixture/gaussian_iter=50.png} &
    \includegraphics[width = \wsgva\textwidth]{figure/2d_mixture/mixture_iter=50.png} \\
     \scriptsize ~~~~Initializations &  \scriptsize Vanilla SVGD &     \scriptsize SVN  &   \scriptsize Matrix-SVGD (average)  &   \scriptsize    Matrix-SVGD (mixture) \\
    \end{tabular}
    \caption{The particles obtained by various methods 
    at the 30/100/30-th iteration on three toy 2D distributions. 
  } 
    \label{fig:app_2d_toy}
    \vspace{12pt}
    \centering
    \setlength\tabcolsep{0.5pt} 
    \begin{tabular}{ccc}
     \hspace{2em} (a) Sine & (b) Double banana & \hspace{-7em}(c) Star  \\
    \raisebox{1em}{\rotatebox{90}{\small Log MMD}}
    \hspace{10pt}
     \includegraphics[height=0.18\textwidth]{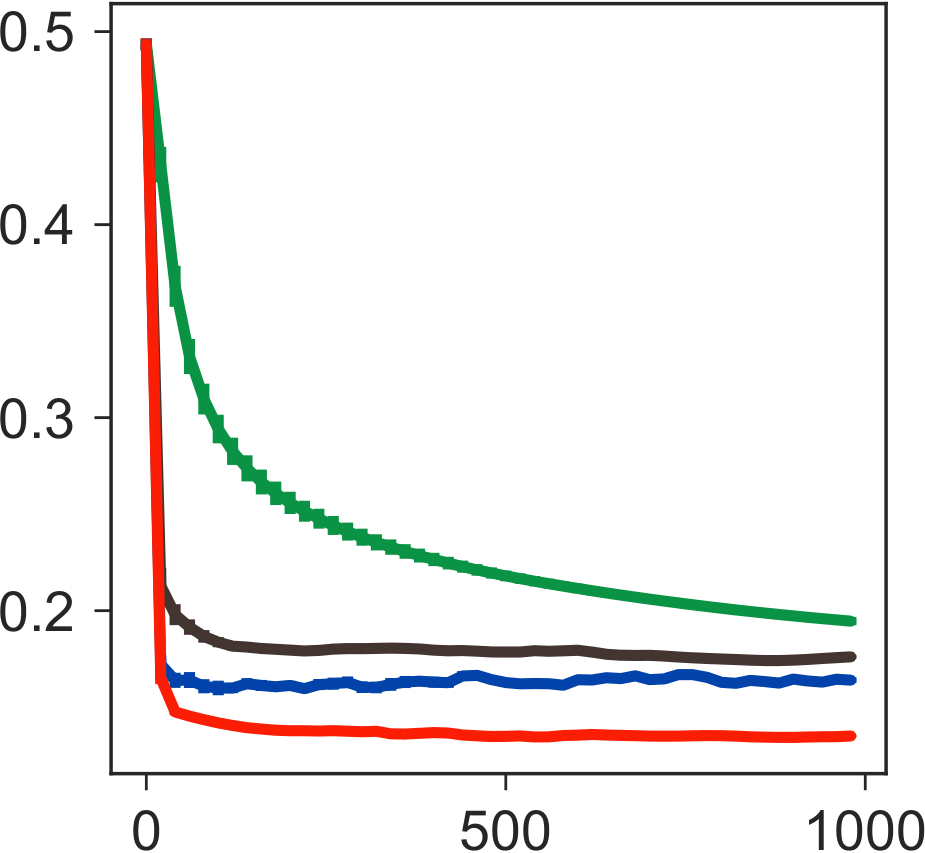} ~~ &
     \includegraphics[height=0.18\textwidth]{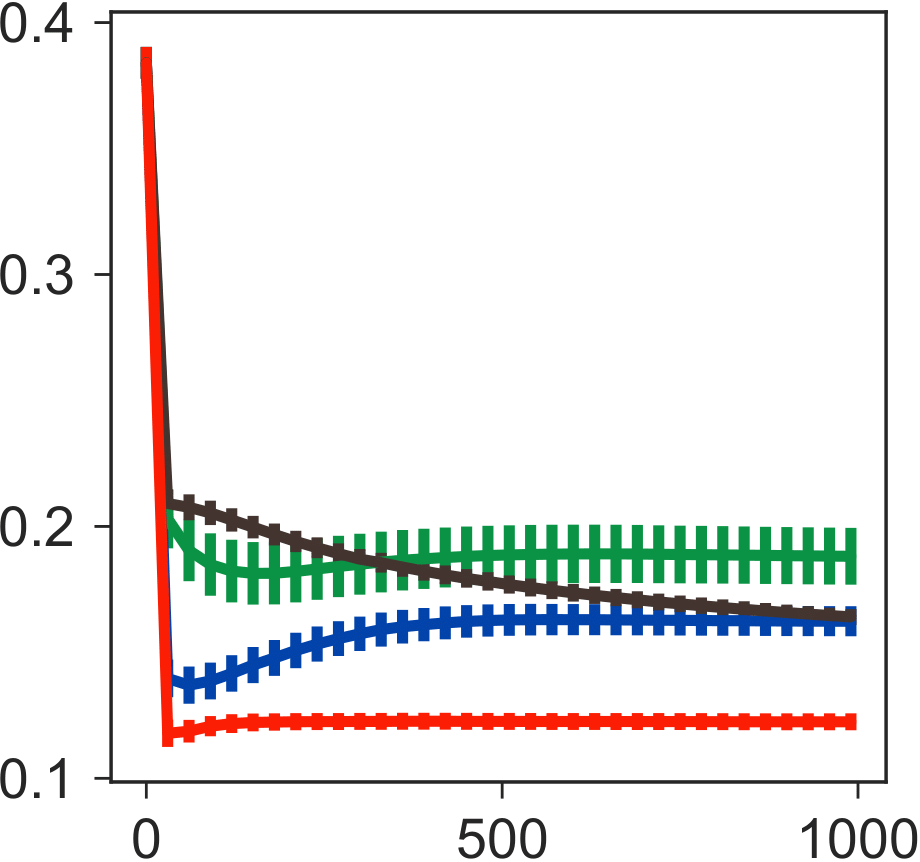} ~~ &
     \includegraphics[height=0.18\textwidth]{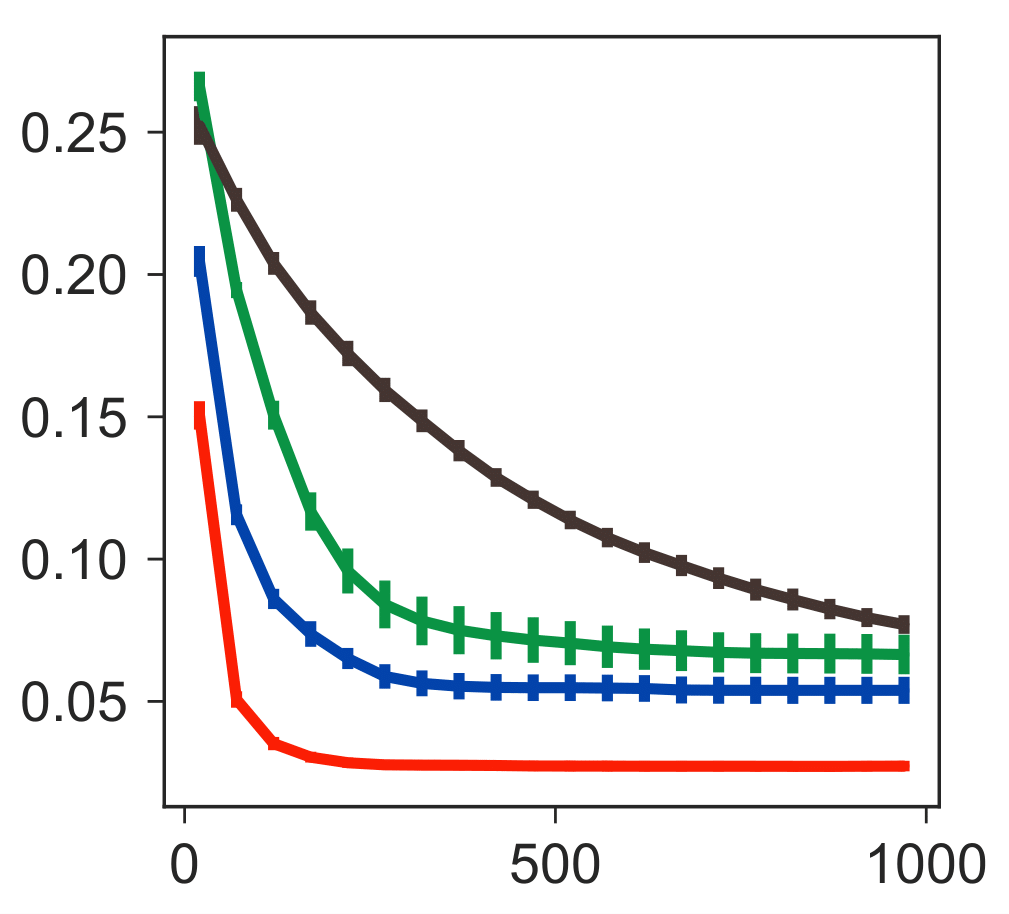} ~~ 
     \raisebox{3em}{\includegraphics[width = 0.2\textwidth]{figure/2d_mixture/star_legend.png}} \\
     \hspace{2em}\# of Iterations & \# of Iterations & \hspace{-7em}\# of Iterations  \\
    \end{tabular}
    \caption{The MMD vs. training iteration of different algorithms on the three toy distributions.}
    \label{fig:app_MMD_2d_toy}
\end{figure*}

\clearpage

\subsection{Sine}\label{sec:sine}

The density function of the ``Sine'' distribution is defined by 
$$p(x_1,x_2)\propto \exp\left(\frac{-(x_2 + \sin(\alpha x_1))^2}{2\sigma_1} - \frac{x_1^2 + x_2^2}{2\sigma_2}\right),$$
where we choose $\alpha = 1$, $\sigma_1 = 0.003$, $\sigma_2 = 1.$

\begin{figure*}[ht]
    \centering
    \setlength\tabcolsep{1pt} 
    \begin{tabular}{ccccc}
    ~~~~~~~~~~Iteration = 0 & Iteration = 5 & Iteration = 10 & Iteration = 30  &  Iteration = 500  \\
    \raisebox{.5em}{\rotatebox{90}{Vanilla}}
    \raisebox{0.6em}{\rotatebox{90}{SVGD}}
     \hspace{10pt}
    \includegraphics[width = \wsgva\textwidth]{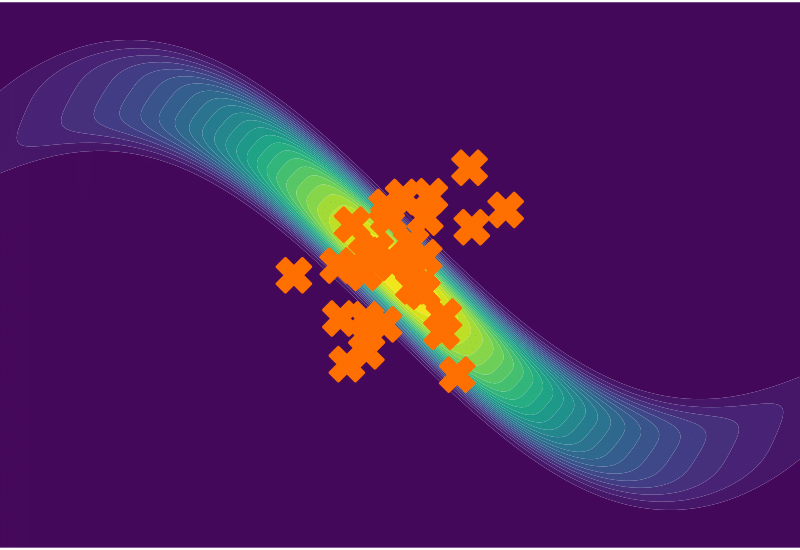} &
    \includegraphics[width = \wsgva\textwidth]{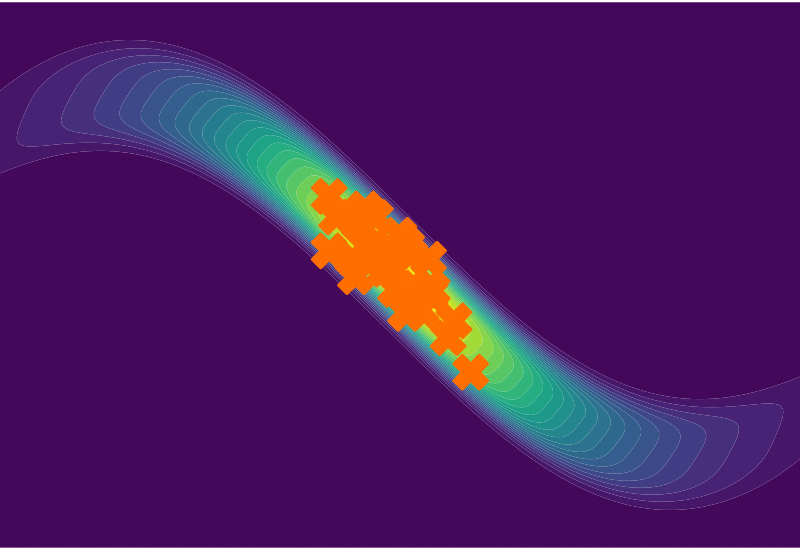} &
    \includegraphics[width = \wsgva\textwidth]{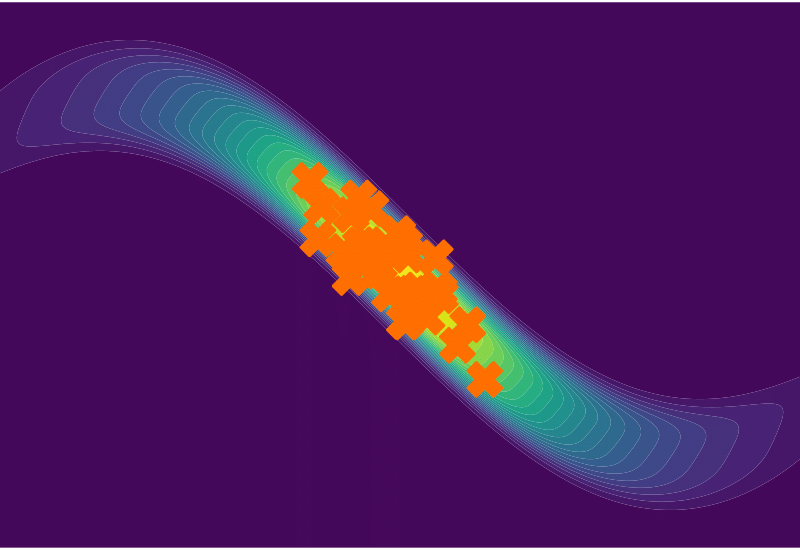} &
    \includegraphics[width = \wsgva\textwidth]{figure/2d_gaussian/rbf_iter=30.png} &
    \includegraphics[width = \wsgva\textwidth]{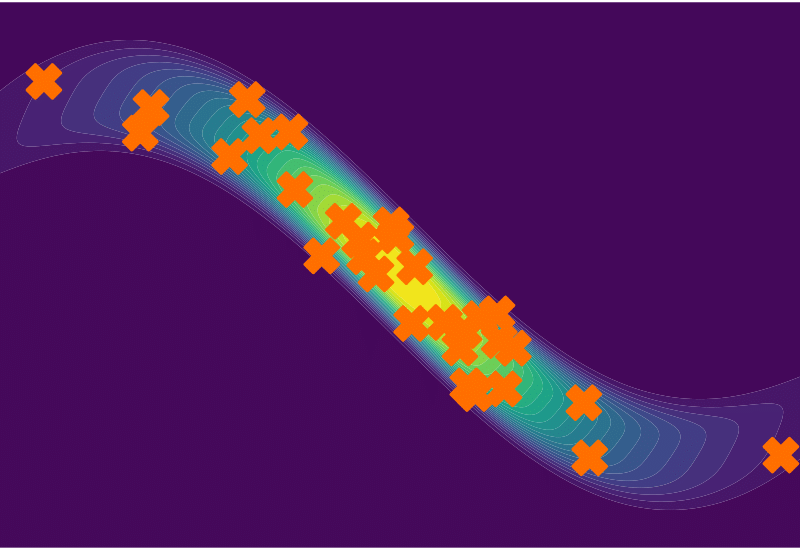} \\
     \raisebox{1em}{\rotatebox{90}{ SVN}} 
     \hspace{19pt}
    \includegraphics[width = \wsgva\textwidth]{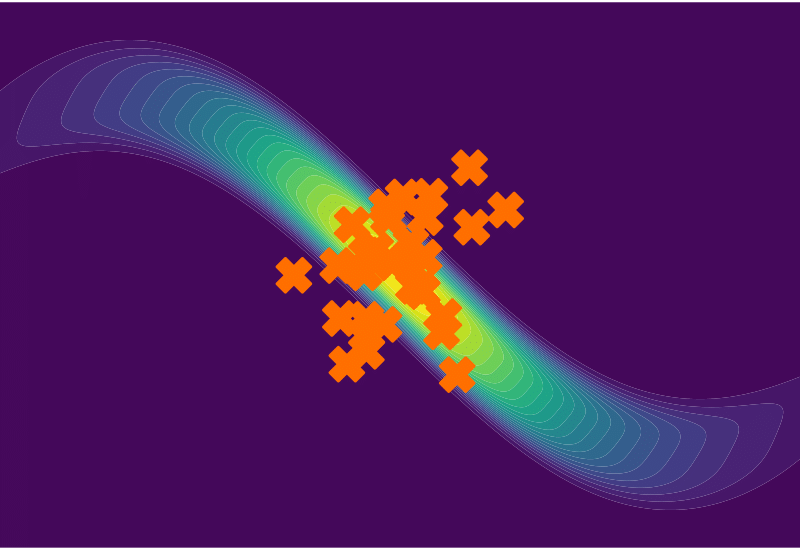} &
    \includegraphics[width = \wsgva\textwidth]{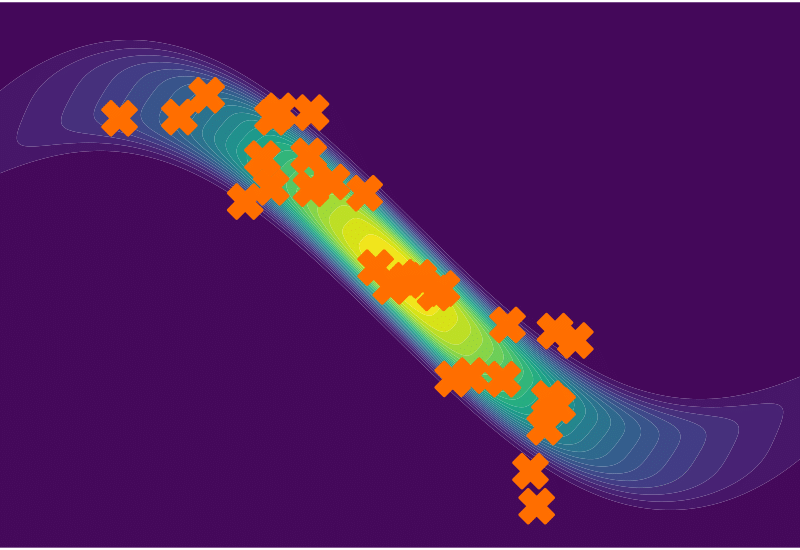} &
    \includegraphics[width = \wsgva\textwidth]{figure/2d_gaussian/newton_iter=30.png} &
    \includegraphics[width = \wsgva\textwidth]{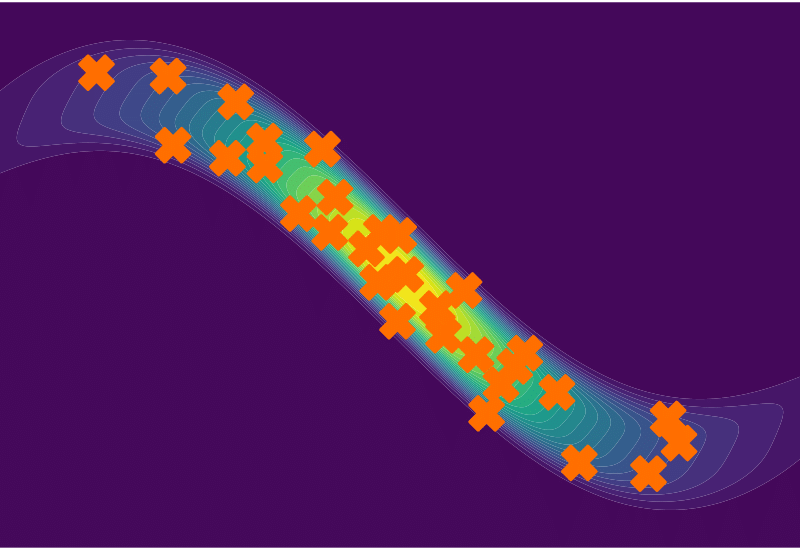} &

    \includegraphics[width = \wsgva\textwidth]{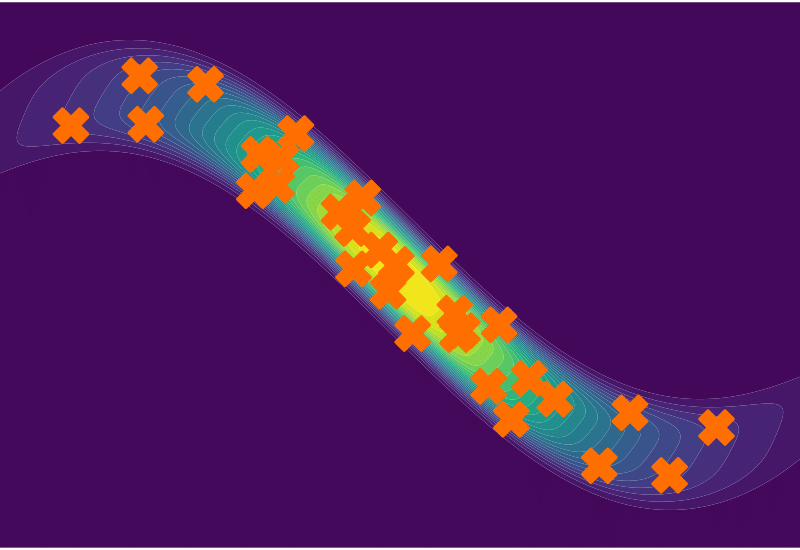} \\
    \raisebox{.0em}{\rotatebox{90}{Matrix SVGD}} 
    \raisebox{1.0em}{\rotatebox{90}{(average)}}
    \hspace{10pt}
    \includegraphics[width = \wsgva\textwidth]{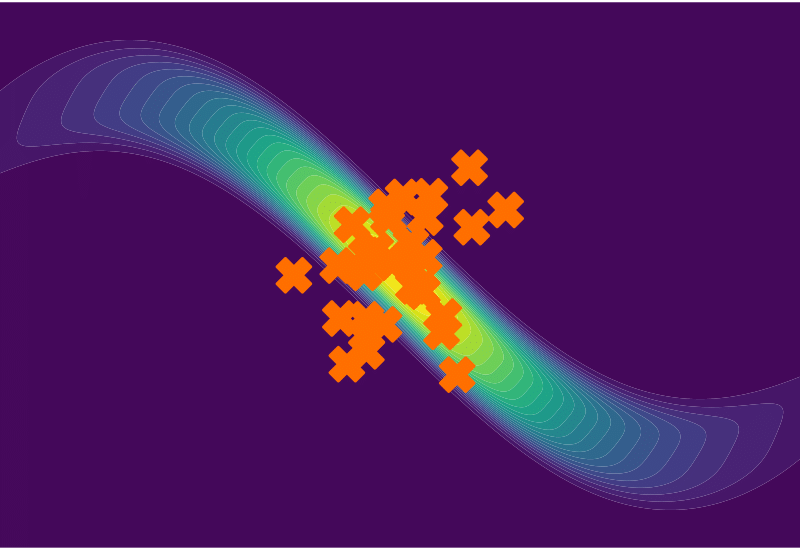} &
     \includegraphics[width = \wsgva\textwidth]{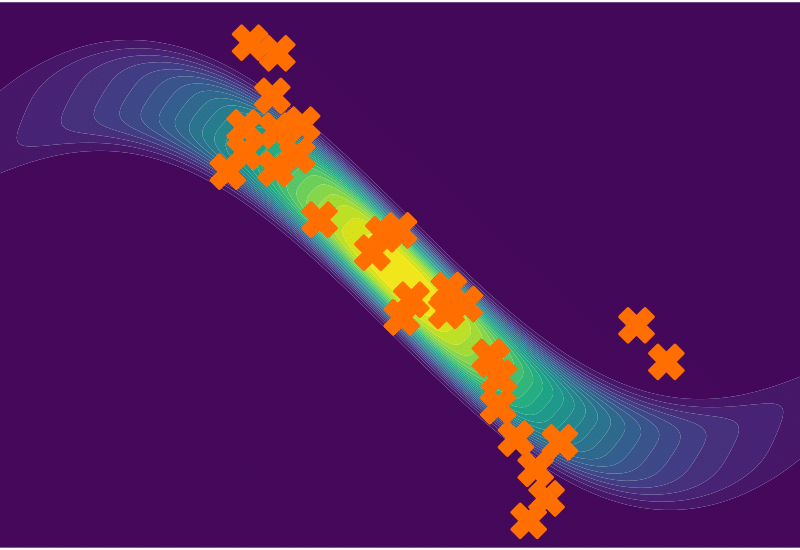} &
   \includegraphics[width = \wsgva\textwidth]{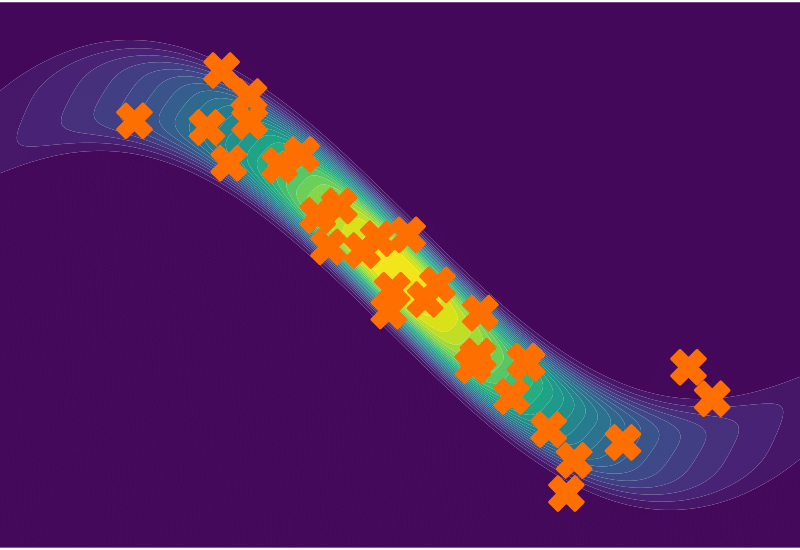} &
    \includegraphics[width = \wsgva\textwidth]{figure/2d_gaussian/gaussian_iter=30.png} &
    \includegraphics[width = \wsgva\textwidth]{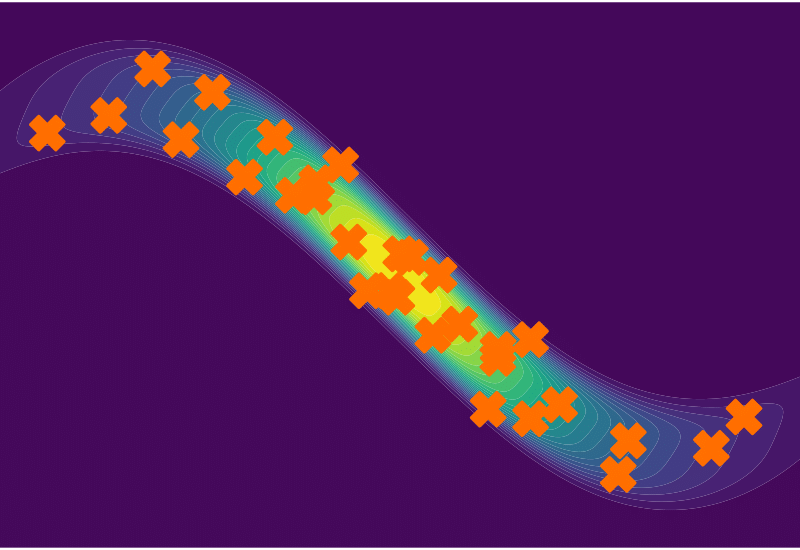} \\
     \raisebox{.0em}{\rotatebox{90}{Matrix SVGD}} 
     \raisebox{1.0em}{\rotatebox{90}{(mixture)}}\hspace{10pt}
    \includegraphics[width = \wsgva\textwidth]{figure/2d_gaussian/mixture_iter=0.png} &
     \includegraphics[width = \wsgva\textwidth]{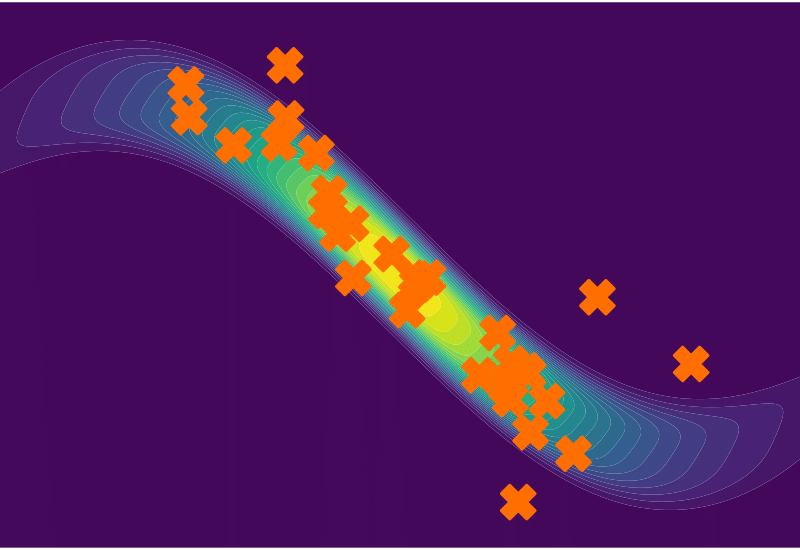} &
   \includegraphics[width = \wsgva\textwidth]{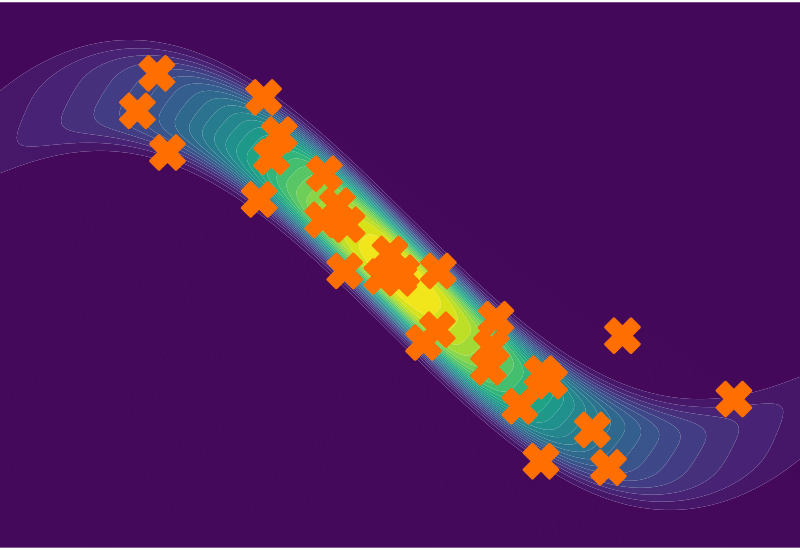} &
    \includegraphics[width = \wsgva\textwidth]{figure/2d_gaussian/mixture_iter=30.png} &
    \includegraphics[width = \wsgva\textwidth]{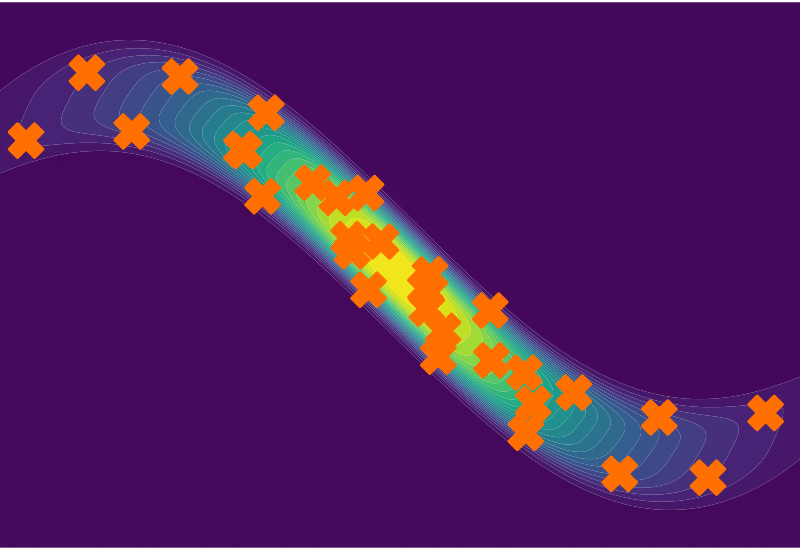} \\
    \end{tabular}
    \caption{The particles obtained by various methods 
   on the toy Sine distribution. }
    \label{fig:2d_iter_toy1}
\end{figure*}

\clearpage

\subsection{Double Banana}
\label{sec:double} 

We use the ``double banana'' distribution 
constructed in \citet{detommaso2018stein}, 
whose probability density function is 
$$p(\vv x) 
\propto \exp\left(-\frac{\|\vv x\|_2^2}{2\sigma_1} - \frac{(y-F(\vv x))^2}{2\sigma_2}\right),$$ 
where  $\vx=[x_1,x_2]\in \RR^2$ and 
$F(\vx) = \log((1-x_1)^2 + 100(x_2-x_1^2)^2)$ and $ y = \log (30)$, $\sigma_1 = 1.0, \sigma_2 = 0.09$. 

\begin{figure*}[ht]
    \centering
    \setlength\tabcolsep{1pt} 
    \begin{tabular}{ccccc}
    ~~~~~~~~~~Iteration = 0 & Iteration = 10 & Iteration = 30 & Iteration = 100  & Iteration =1000 \\
    \raisebox{1em}{\rotatebox{90}{Vanilla SVGD}}
     \hspace{19pt}
    \includegraphics[width = \wsgva\textwidth]{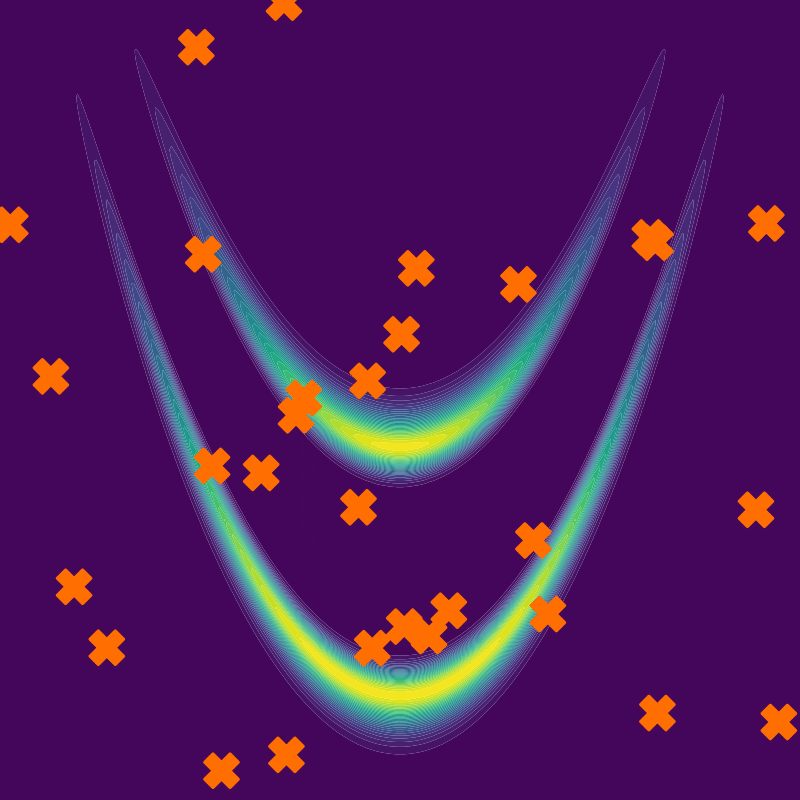} &
    \includegraphics[width = \wsgva\textwidth]{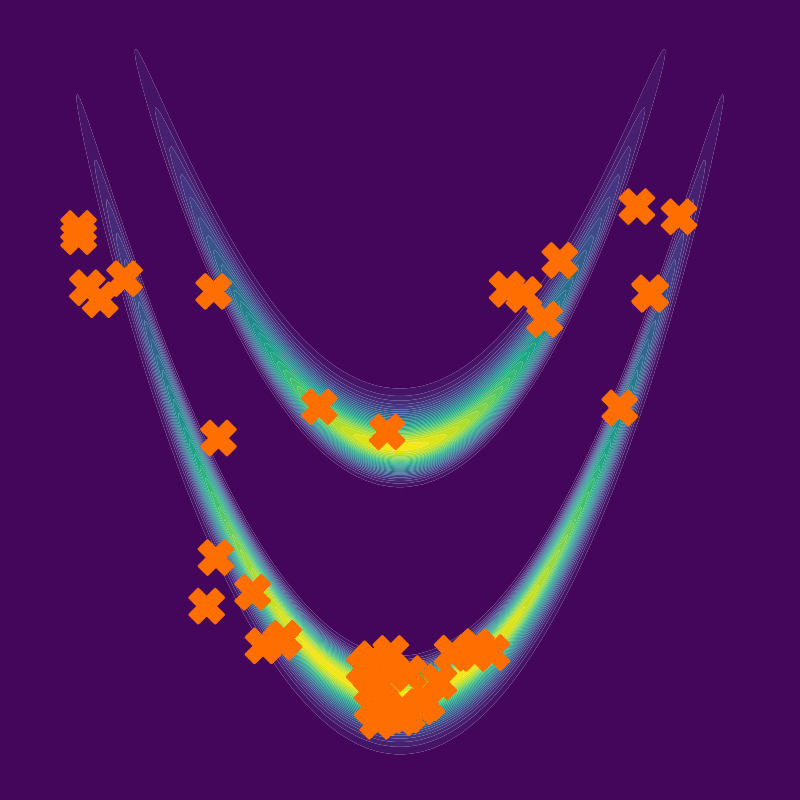} &
    \includegraphics[width = \wsgva\textwidth]{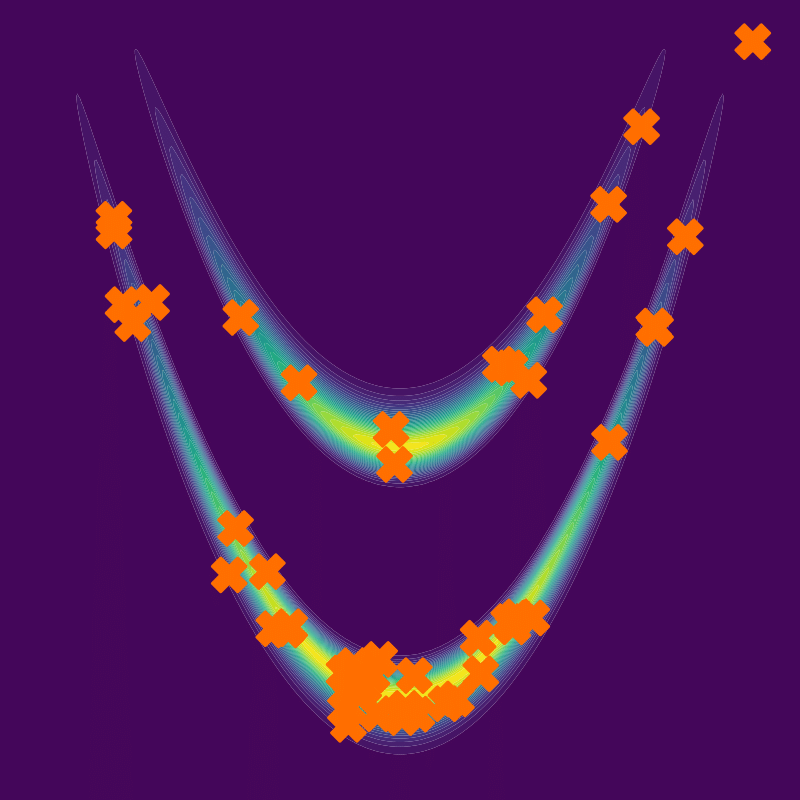} &
    \includegraphics[width = \wsgva\textwidth]{figure/banana/rbf_iter=100.png} &
    \includegraphics[width = \wsgva\textwidth]{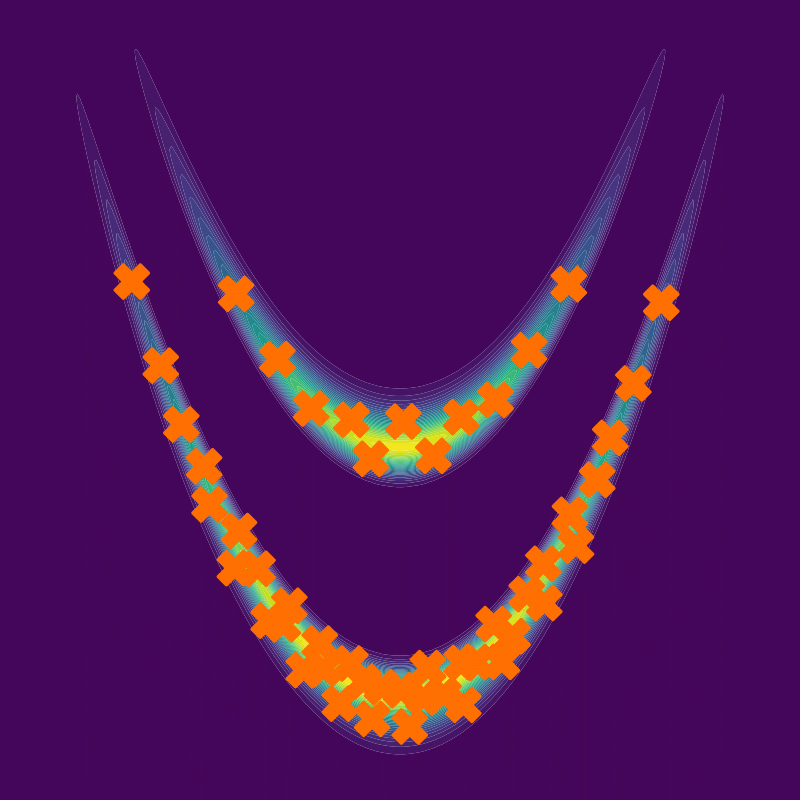} \\
     \raisebox{3em}{\rotatebox{90}{SVN}} 
     \hspace{19pt}
    \includegraphics[width = \wsgva\textwidth]{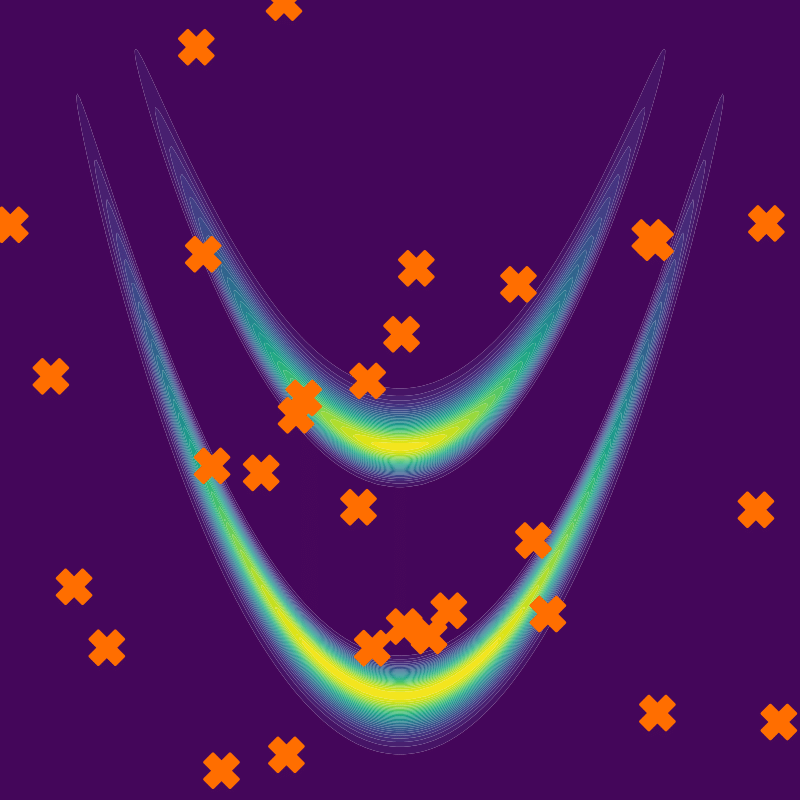} &  \includegraphics[width = \wsgva\textwidth]{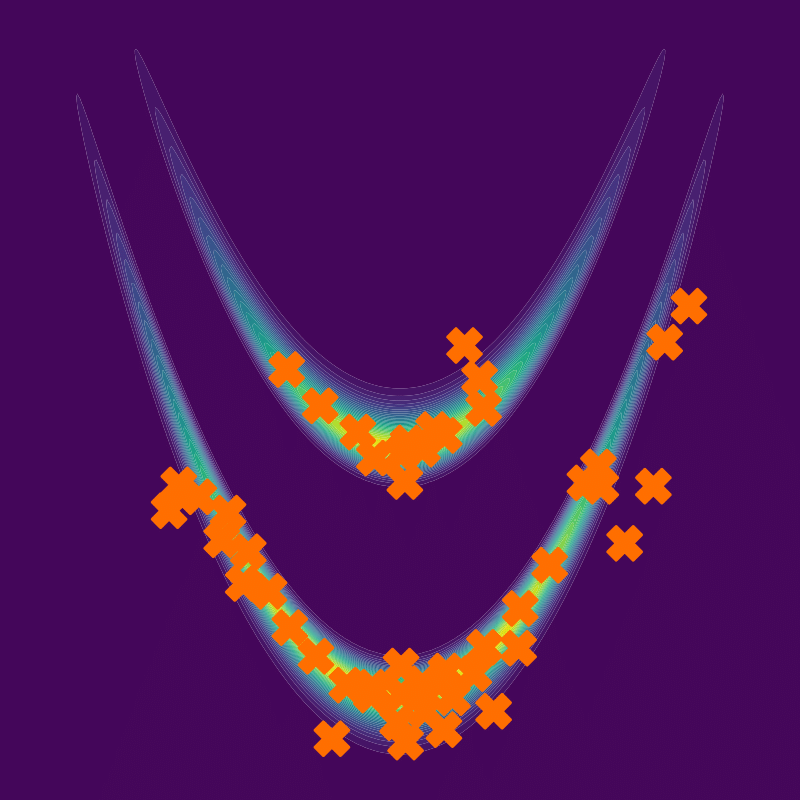} &
    \includegraphics[width = \wsgva\textwidth]{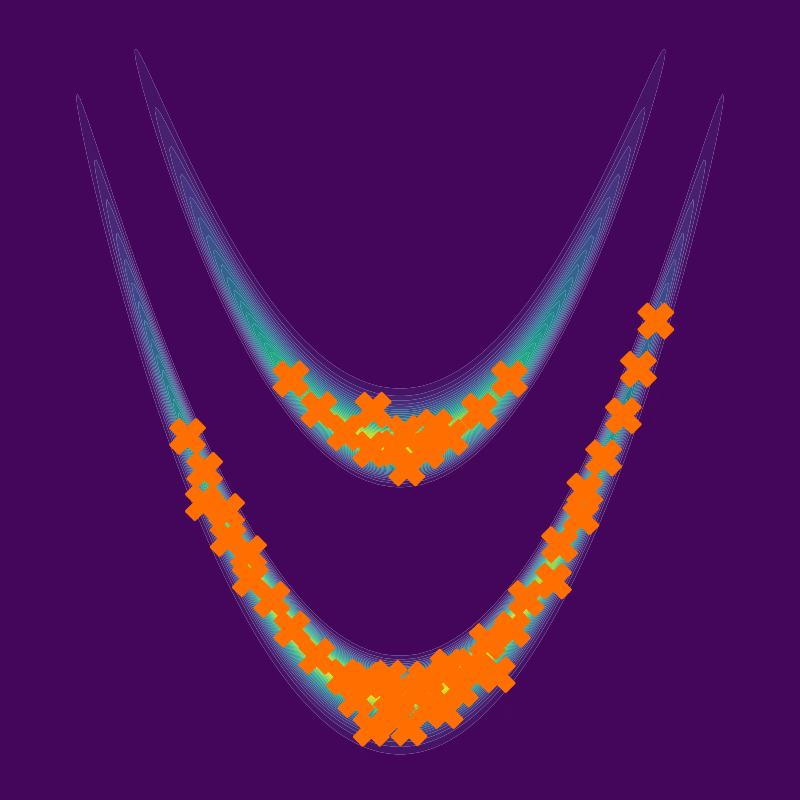} &
    \includegraphics[width = \wsgva\textwidth]{figure/banana/newton_iter=100.png} &
    \includegraphics[width = \wsgva\textwidth]{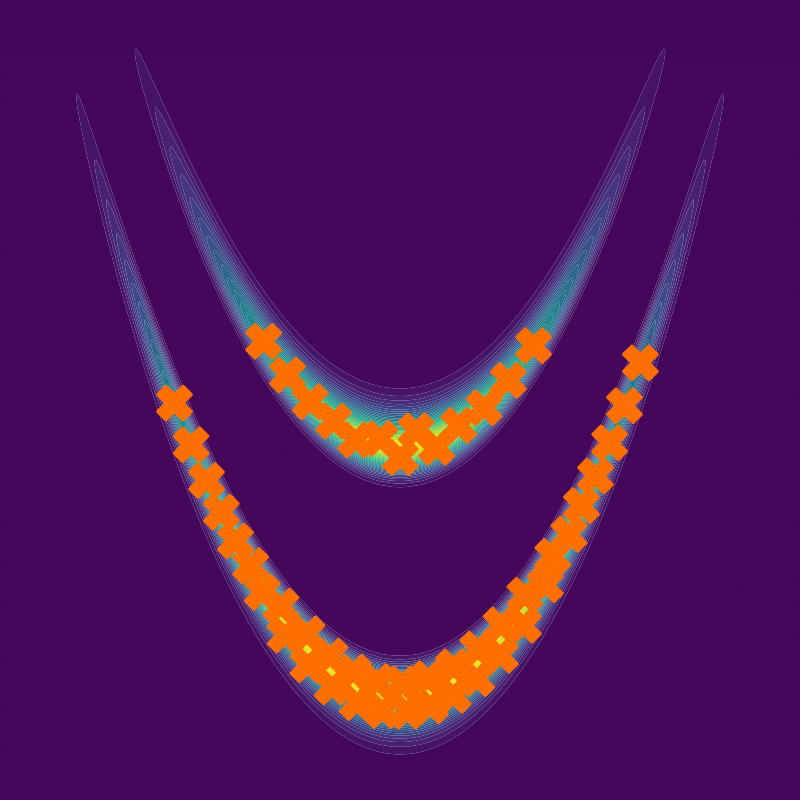} \\
    \raisebox{1.5em}{\rotatebox{90}{Matrix SVGD}} \raisebox{2.5em}{\rotatebox{90}{(average)}}
    \hspace{10pt}
    \includegraphics[width = \wsgva\textwidth]{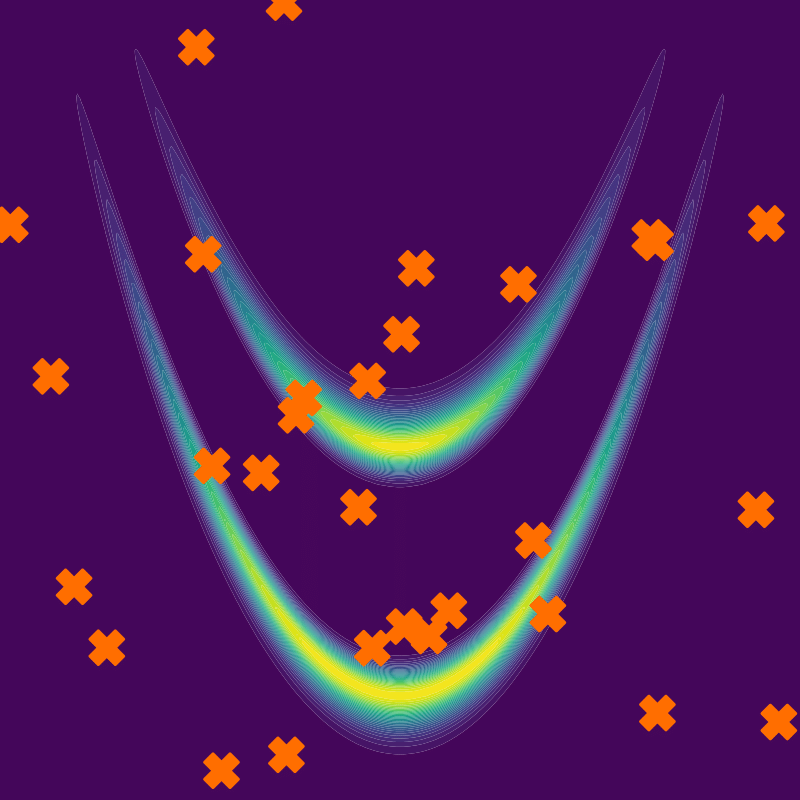} &
        \includegraphics[width = \wsgva\textwidth]{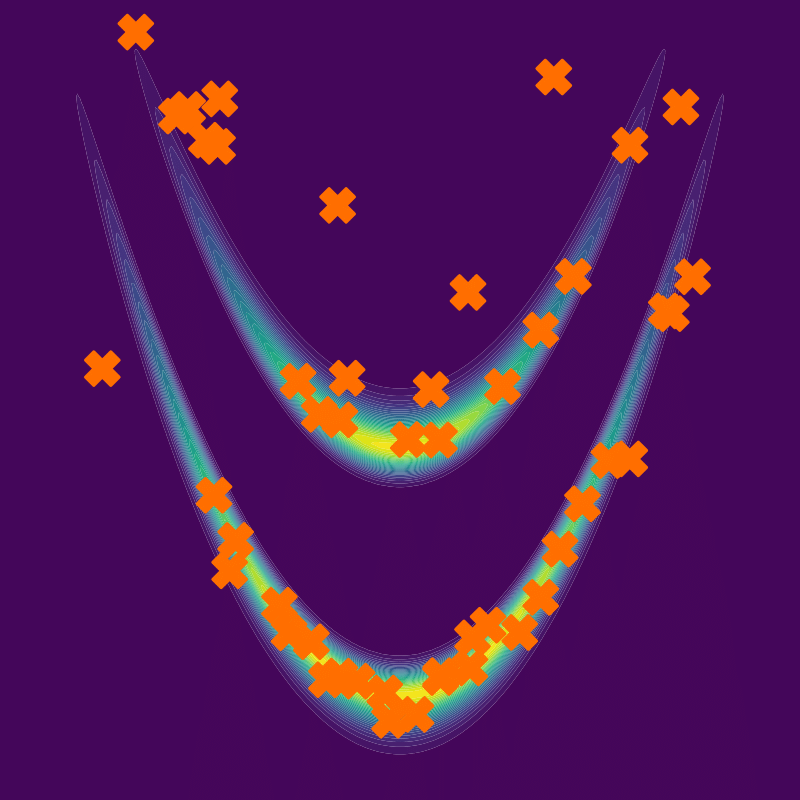} &
    \includegraphics[width = \wsgva\textwidth]{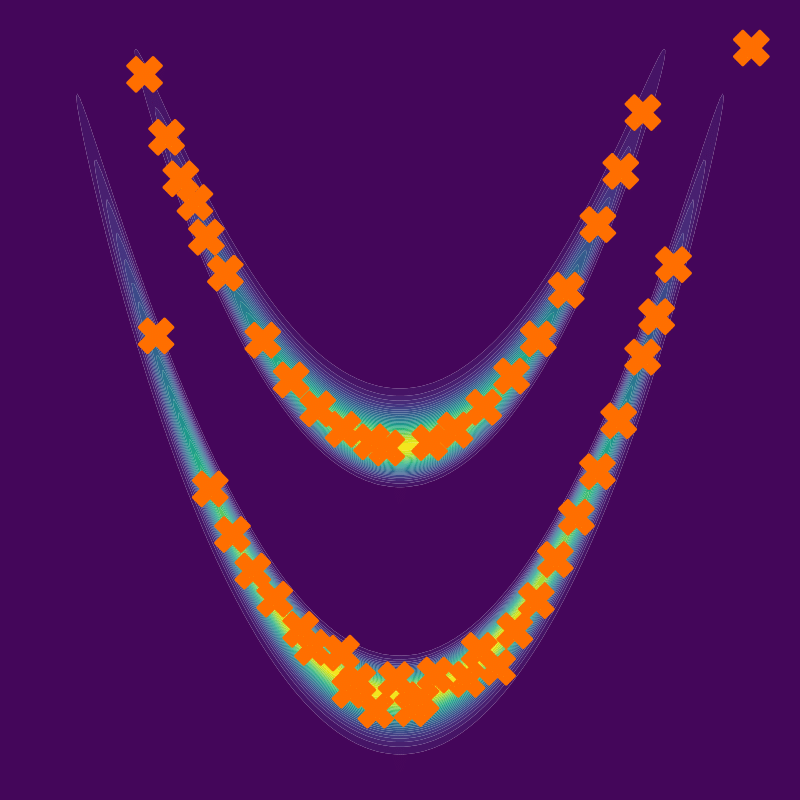} &
    \includegraphics[width = \wsgva\textwidth]{figure/banana/gaussian_iter=100.png} &
    \includegraphics[width = \wsgva\textwidth]{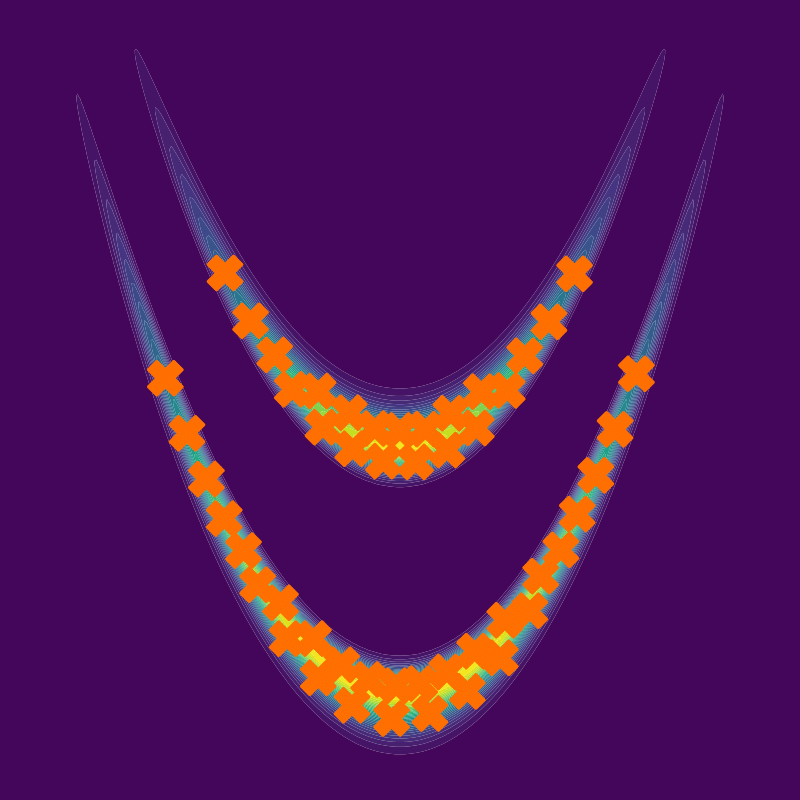} \\
     \raisebox{1.5em}{\rotatebox{90}{Matrix SVGD}} 
     \raisebox{2.5em}{\rotatebox{90}{(mixture)}}\hspace{10pt}
    \includegraphics[width = \wsgva\textwidth]{figure/banana/mixture_iter=0.png} &
        \includegraphics[width = \wsgva\textwidth]{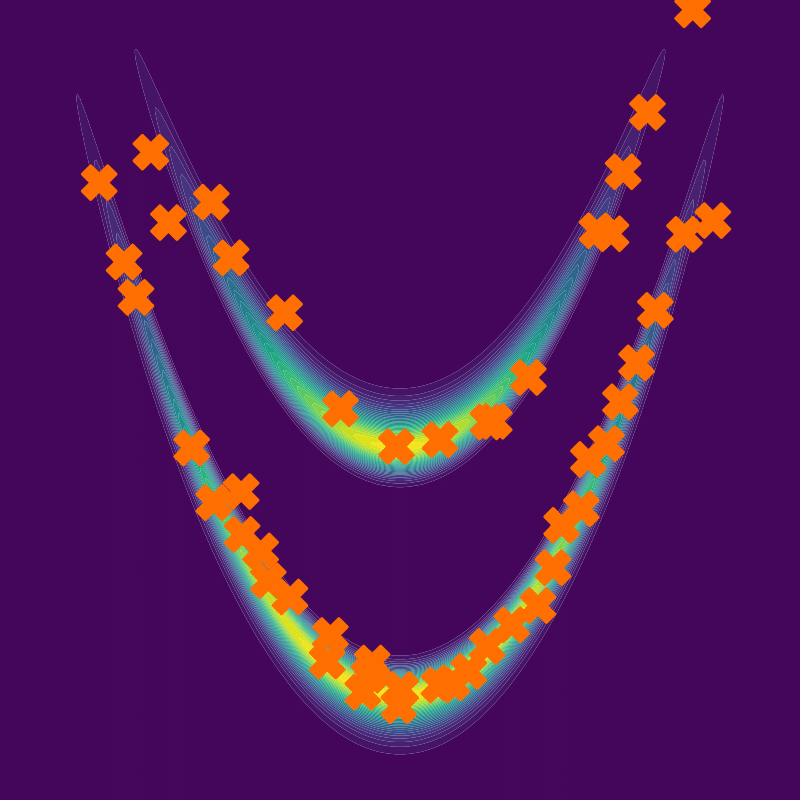} &
    \includegraphics[width = \wsgva\textwidth]{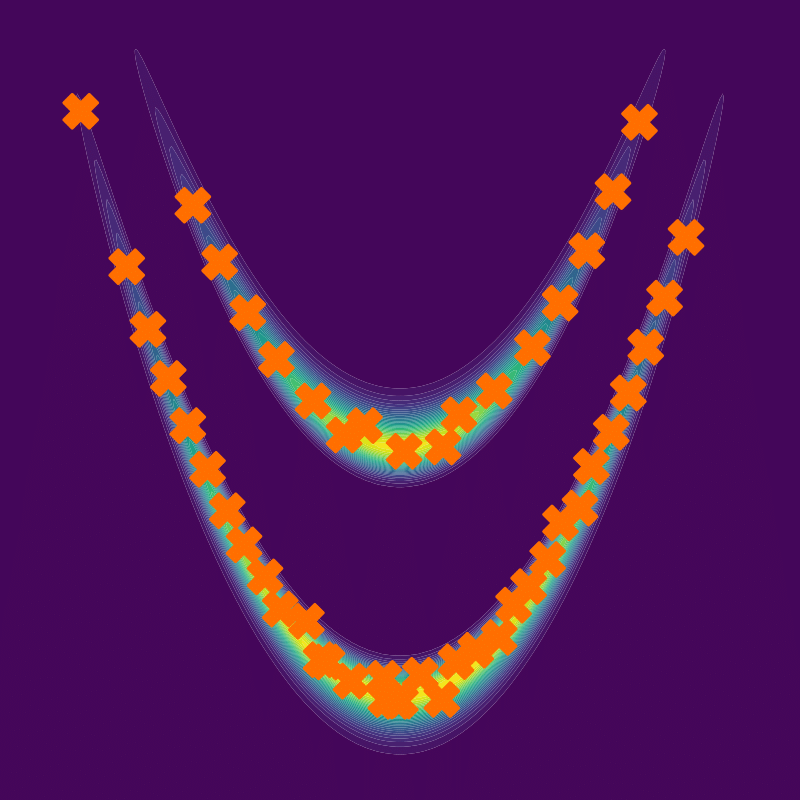} &
    \includegraphics[width = \wsgva\textwidth]{figure/banana/mixture_iter=100.png} &
    \includegraphics[width = \wsgva\textwidth]{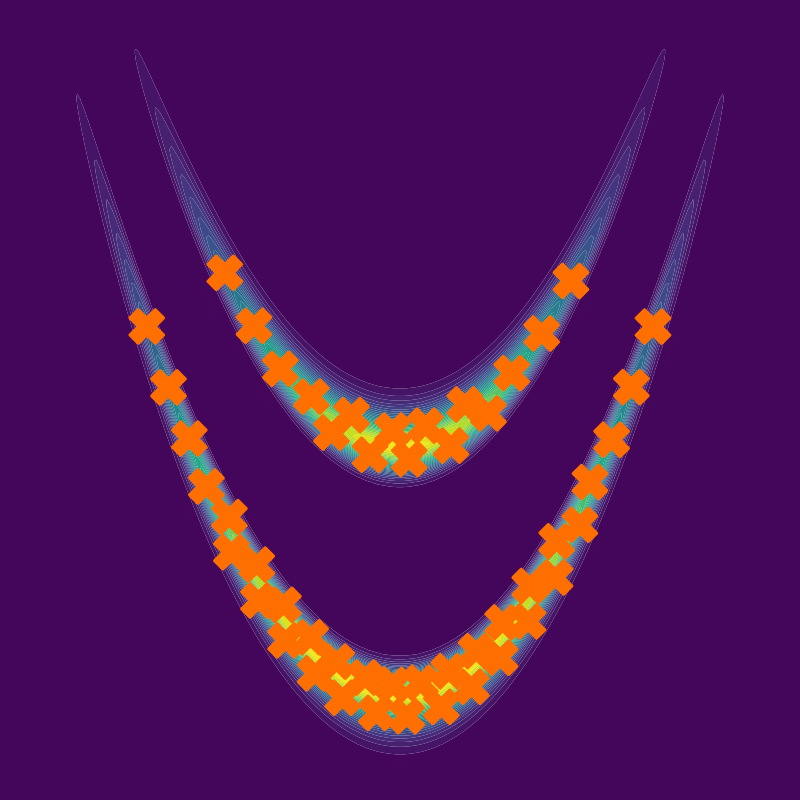} \\
    \end{tabular}
    \caption{The particles obtained by various methods 
   on the  double banana distribution.}
    \label{fig:2d_iter_toy2}
\end{figure*}

\clearpage

\subsection{Star}
\label{sec:star}
We construct the ``star'' distribution with a Gaussian mixture model, 
whose density function is 
$$p(\vx) =
\frac{1}{K}\sum_{i=1}^K \N(\vx; \vv\mu_i, \vv\Sigma_i),$$
with $\vx\in \RR^2$
 $\vv\mu_1 = [0; ~ 1.5]$, $\vv\Sigma_1 = \mathrm{diag}([1; ~ \frac{1}{100}])$, and 
the other means and covariance matrices are defined by rotating their previous mean and covariance matrix. To be precise, 
$$ \vv\mu_{i+1} = \vv U \vv\mu_{i}, ~~~~\vv\Sigma_{i+1} = \vv U \vv\Sigma_i \vv U^\top,~~~~
~~\vv U = \left[\begin{array}{cc}\cos(\theta) & \sin(\theta)\\ -\sin(\theta) & \cos(\theta) \end{array}\right],
$$ 
with angle $\theta = \frac{2\pi}{K}$.
We set the number of component $K$ to be 5.

\begin{figure*}[ht]
    \centering
    \setlength\tabcolsep{1pt} 
    \begin{tabular}{ccccc}
    ~~~~~~~~~~Iteration = 0 & Iteration = 30 & Iteration = 100 & Iteration = 500 & Iteration =1000 \\
    \raisebox{1em}{\rotatebox{90}{Vanilla SVGD}}
     \hspace{19pt}
    \includegraphics[width = \wsgva\textwidth]{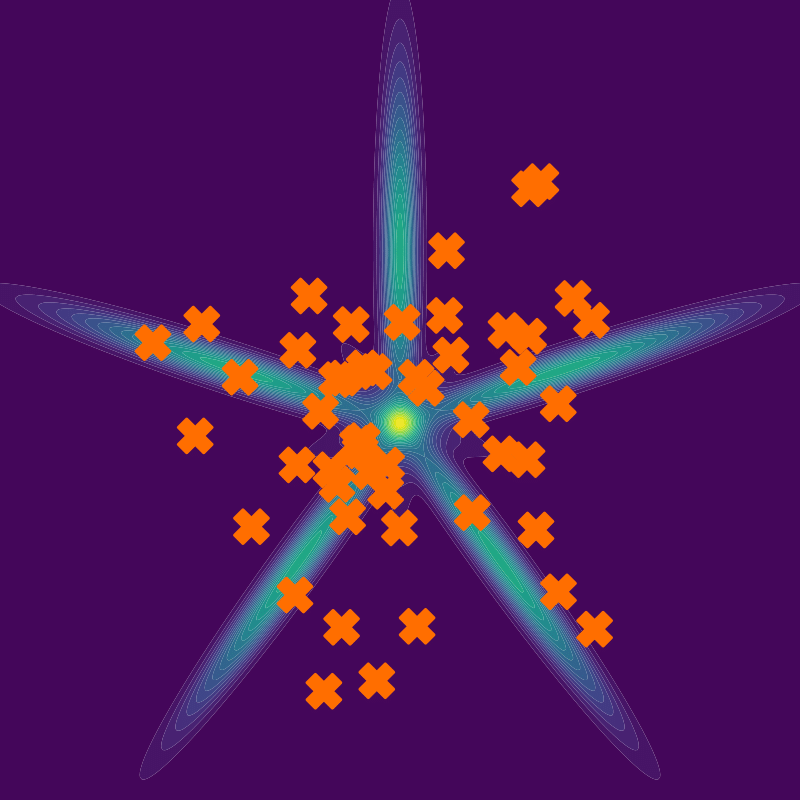} &
    \includegraphics[width = \wsgva\textwidth]{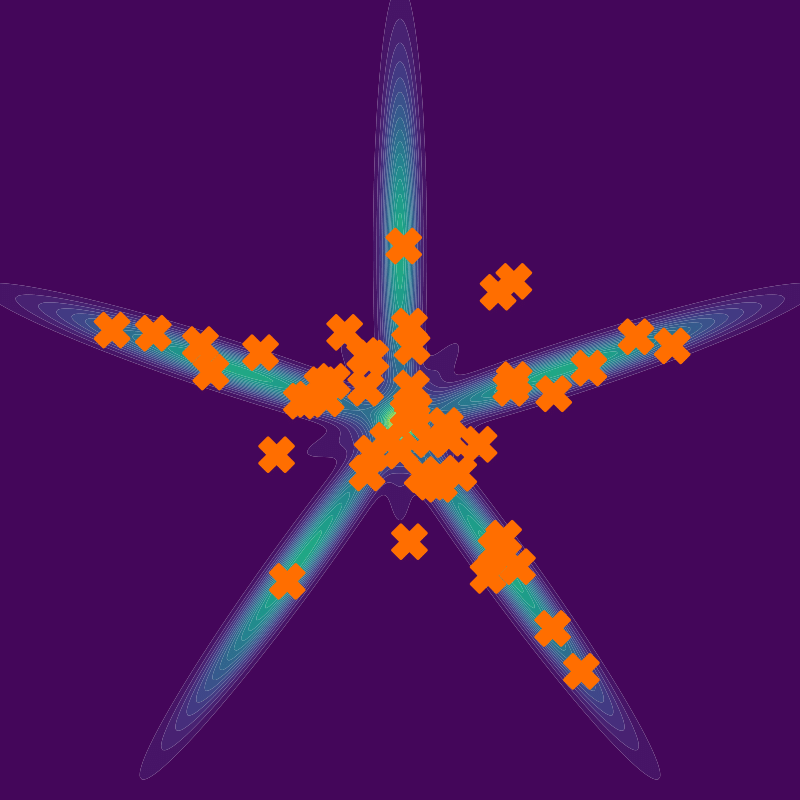} &
    \includegraphics[width = \wsgva\textwidth]{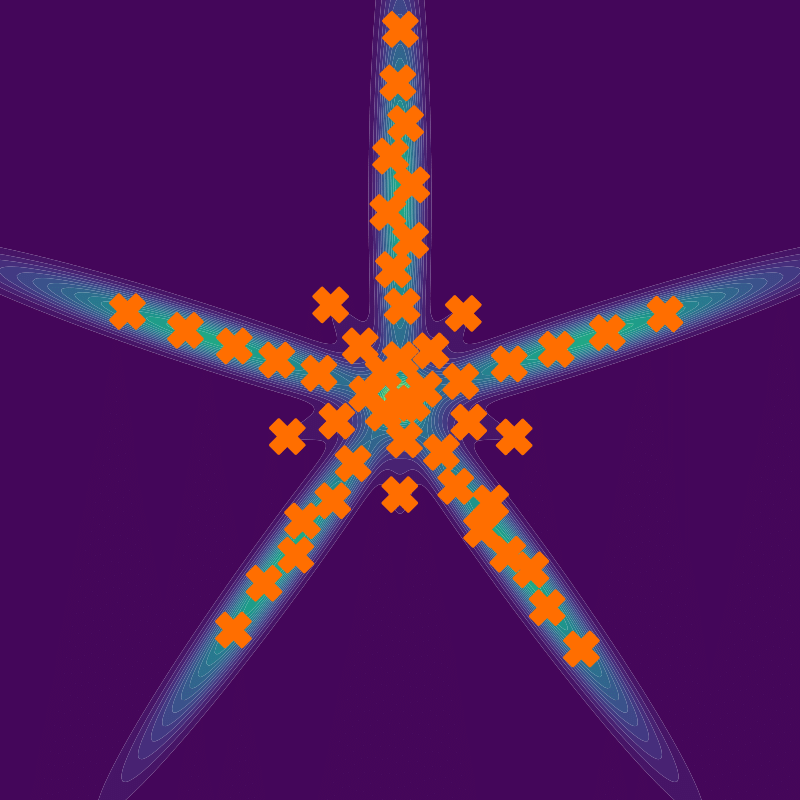} &
        \includegraphics[width = \wsgva\textwidth]{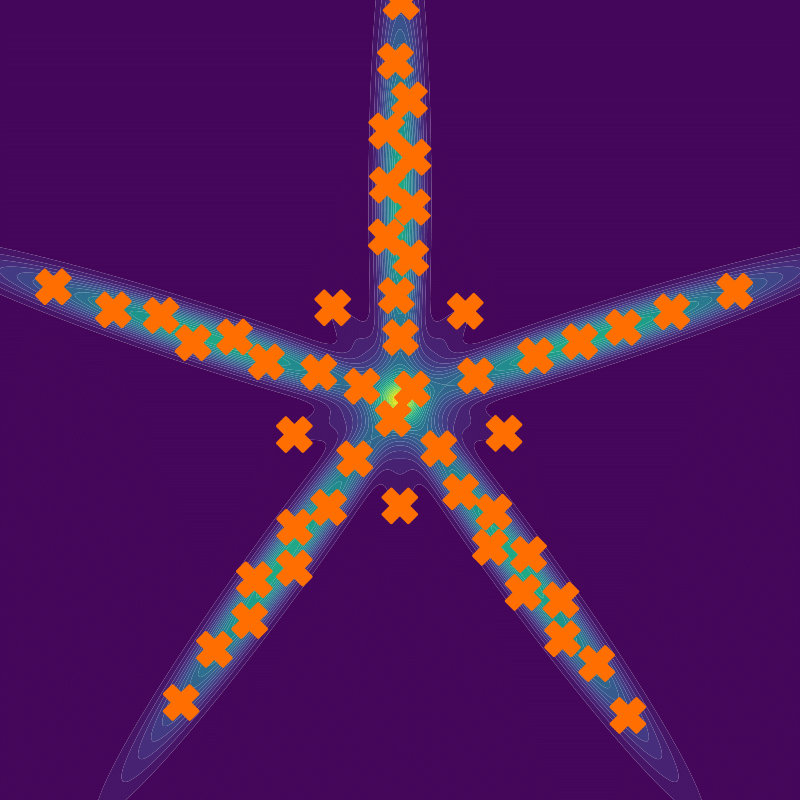} &
    \includegraphics[width = \wsgva\textwidth]{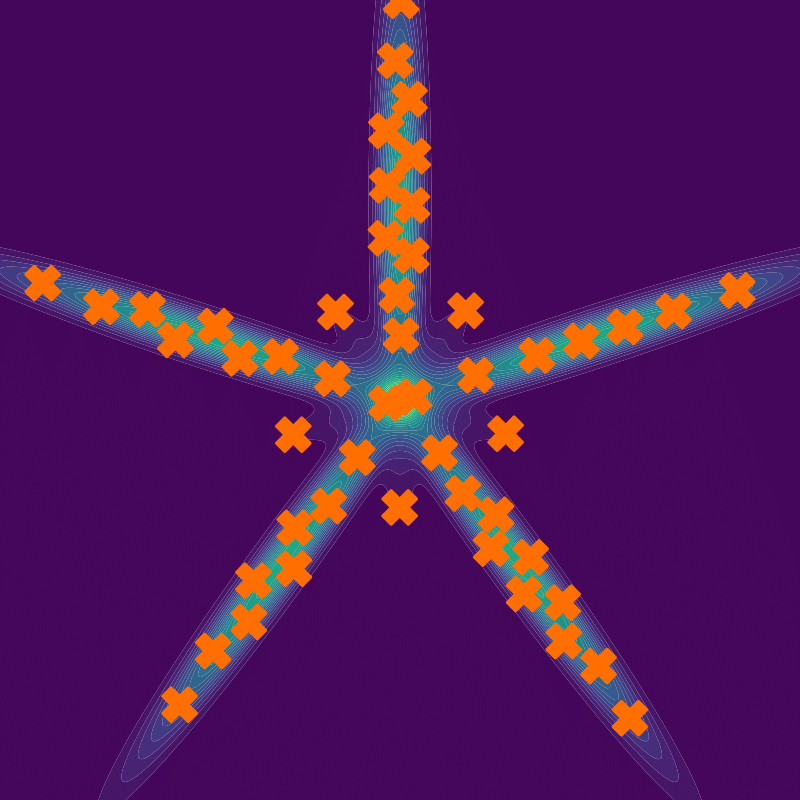} \\
     \raisebox{4em}{\rotatebox{90}{SVN}} 
     \hspace{19pt}
    \includegraphics[width = \wsgva\textwidth]{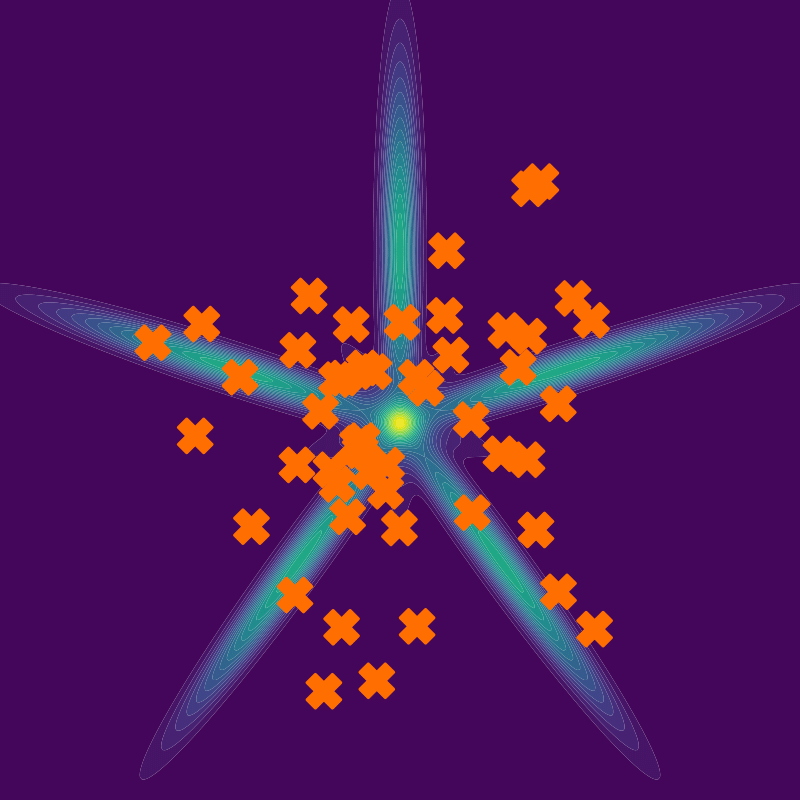} &
    \includegraphics[width = \wsgva\textwidth]{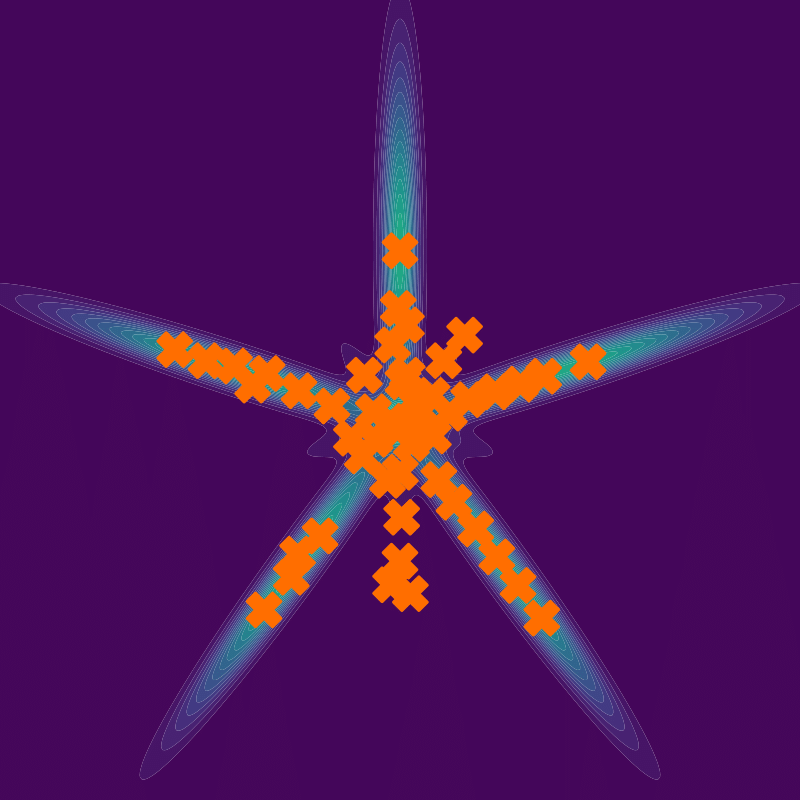} &
    \includegraphics[width = \wsgva\textwidth]{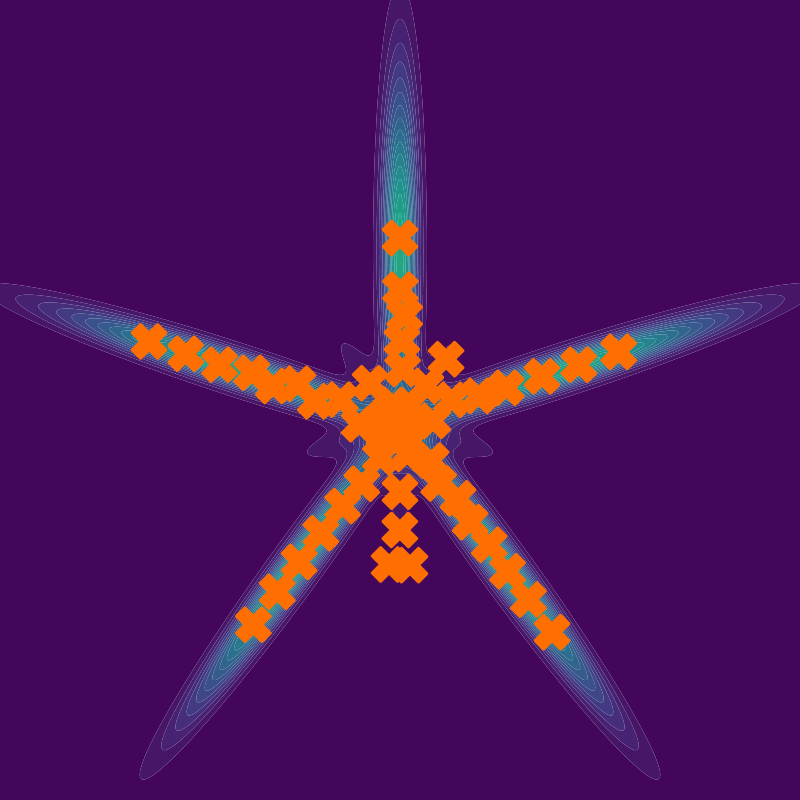} &
        \includegraphics[width = \wsgva\textwidth]{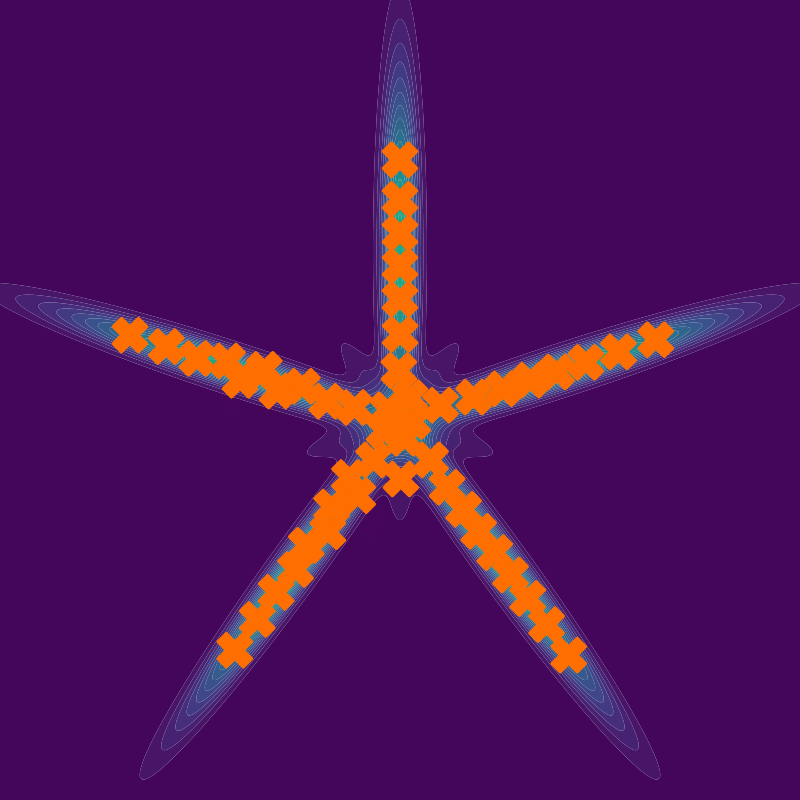} &
    \includegraphics[width = \wsgva\textwidth]{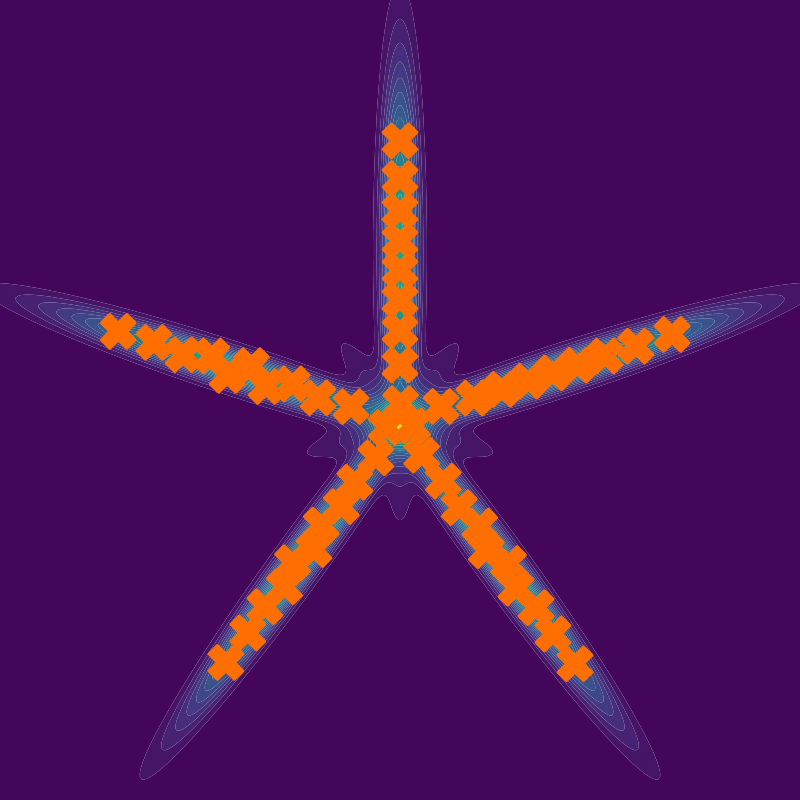} \\
    \raisebox{1.5em}{\rotatebox{90}{Matrix SVGD}} \raisebox{2.5em}{\rotatebox{90}{(average)}}
    \hspace{10pt}
    \includegraphics[width = \wsgva\textwidth]{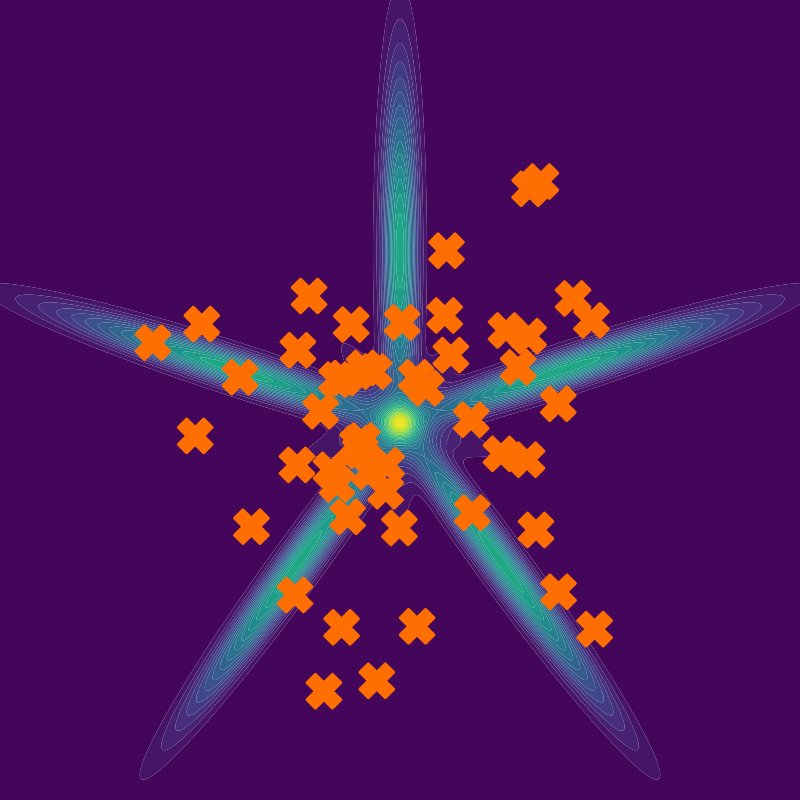} &
    \includegraphics[width = \wsgva\textwidth]{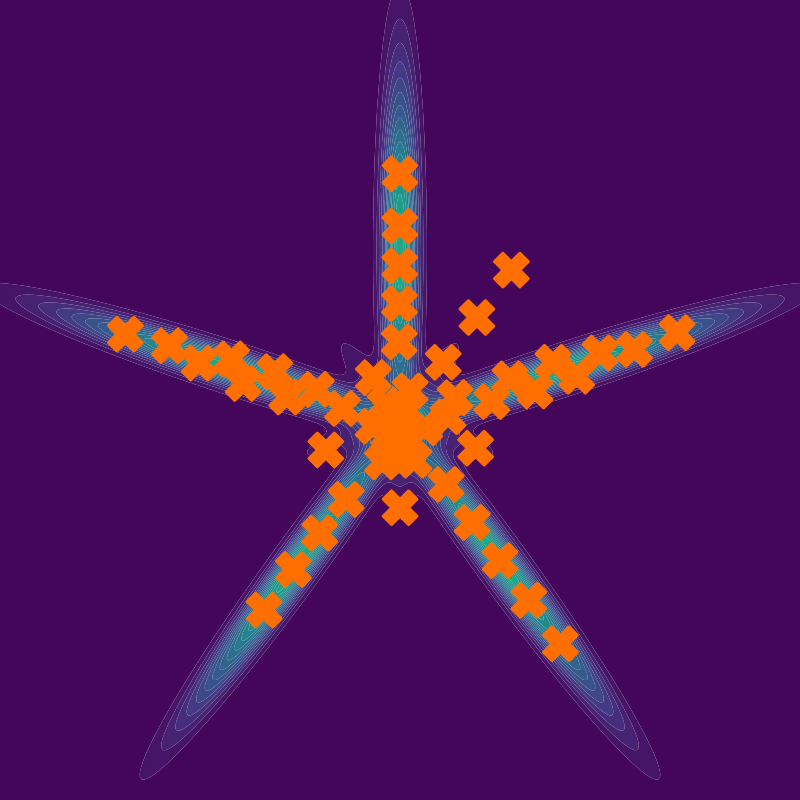} &
    \includegraphics[width = \wsgva\textwidth]{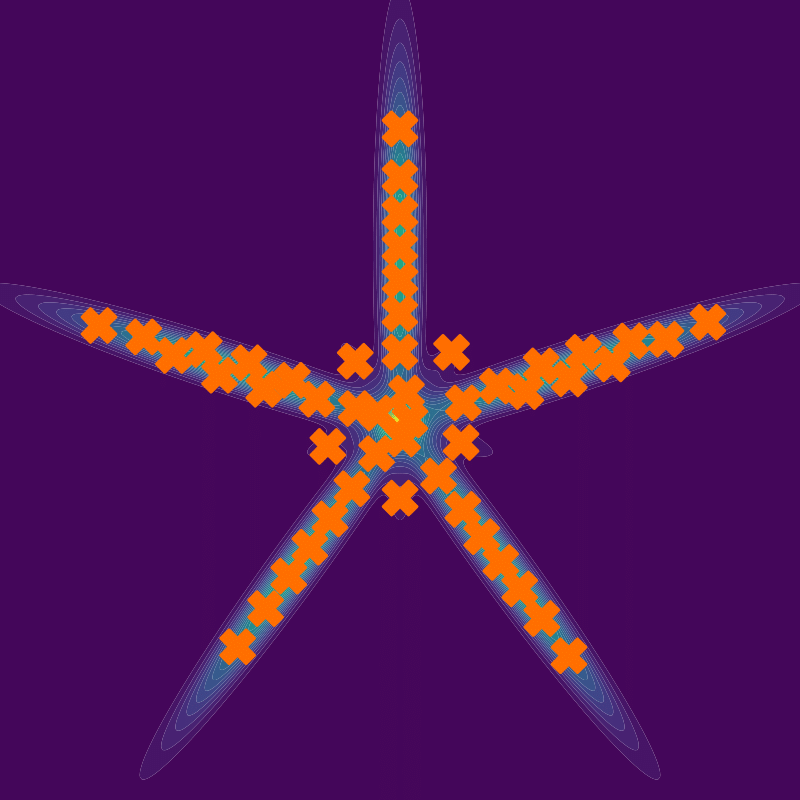} &
        \includegraphics[width = \wsgva\textwidth]{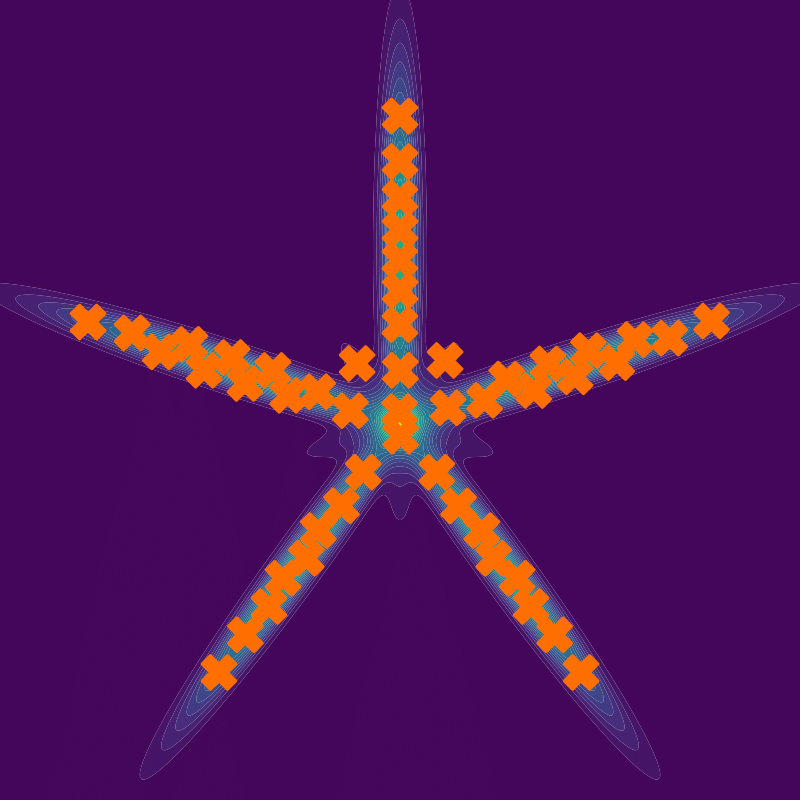} &
    \includegraphics[width = \wsgva\textwidth]{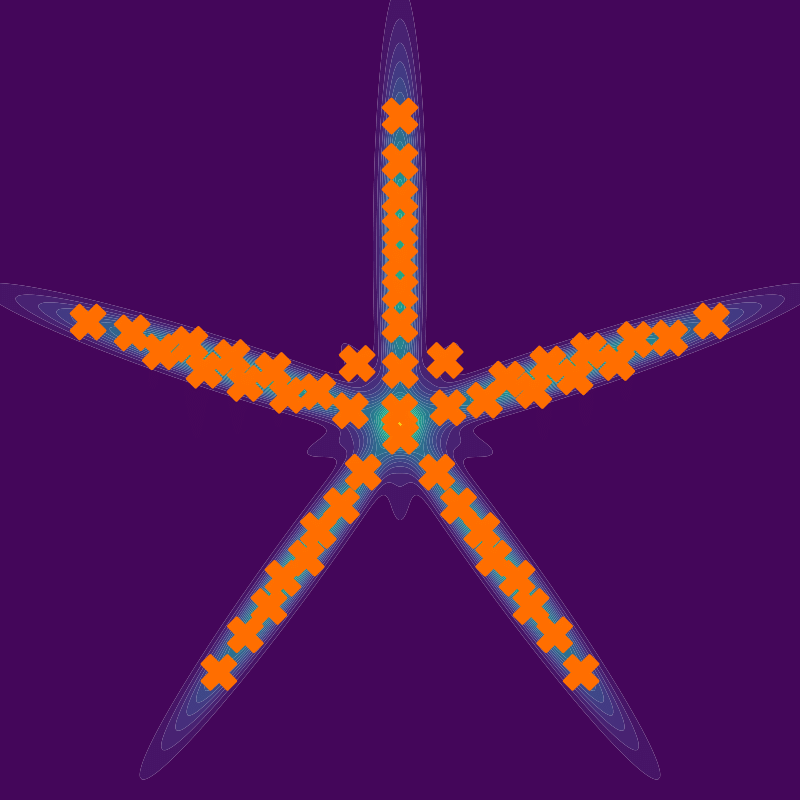} \\
     \raisebox{1.5em}{\rotatebox{90}{Matrix SVGD}} 
     \raisebox{2.5em}{\rotatebox{90}{(mixture)}}\hspace{10pt}
    \includegraphics[width = \wsgva\textwidth]{figure/2d_mixture/mixture_iter=0.png} &
    \includegraphics[width = \wsgva\textwidth]{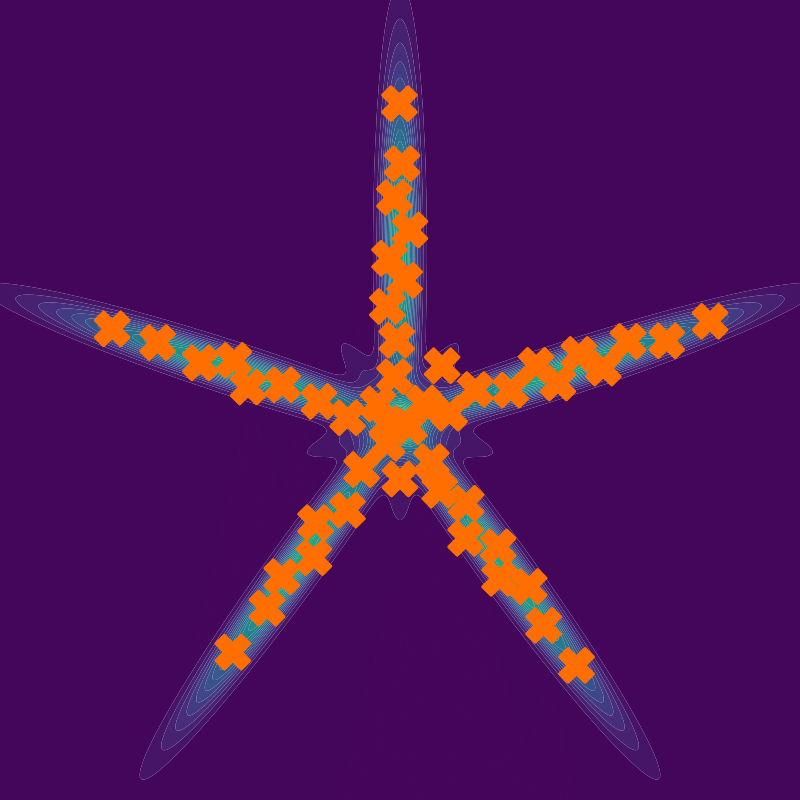} &
    \includegraphics[width = \wsgva\textwidth]{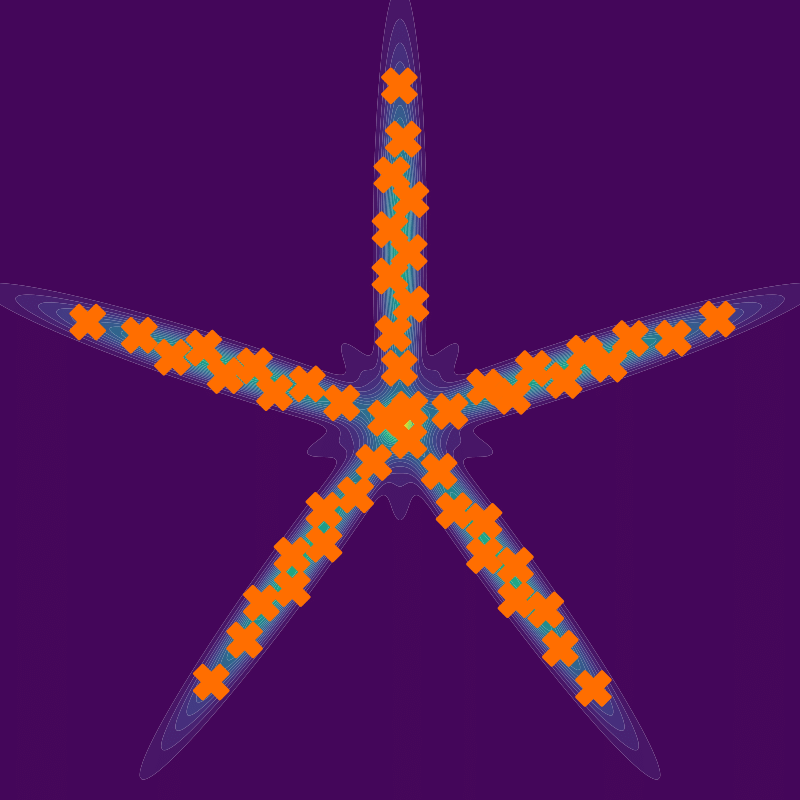} &
        \includegraphics[width = \wsgva\textwidth]{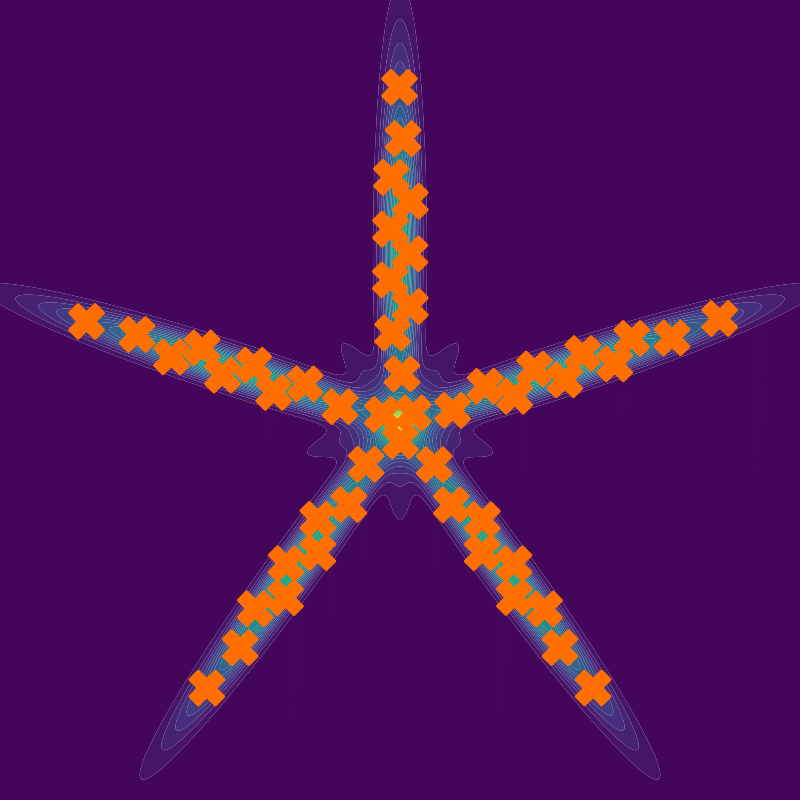} &
    \includegraphics[width = \wsgva\textwidth]{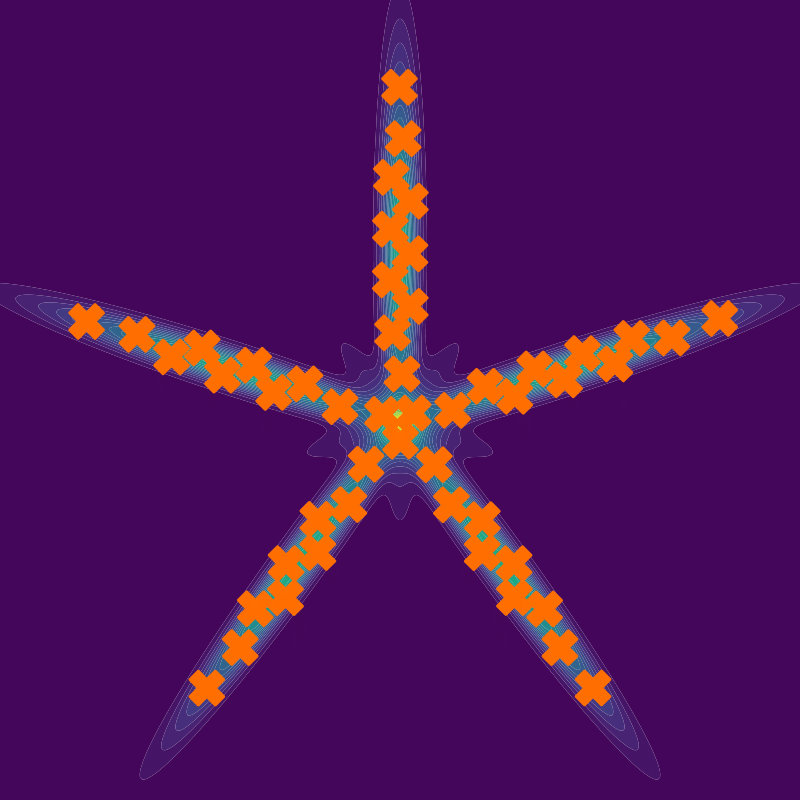} \\
    \end{tabular}
    \caption{The particles obtained by various methods 
   on the star-shaped distribution. }
    \label{fig:2d_iter_toy3}
\end{figure*}